FACHHOCHSCHULE LAUSITZ
University of Applied Sciences

Studiengang Informatik

# DIPLOMARBEIT

Umgebungserfassungssystem
für mobile Roboter

Dirk Hesselbach

(geb. am 28. Juli 1979)

1. Betreuer: Prof. Dr.-Ing. Ernst Reinhold

2. Betreuer: Prof. Dr. Wolfgang Laßner

# Aufgabenstellung

Zur Umgebungserfassung eines mobilen Roboters soll eine kostengünstige Alternative zu den mit hoher Messgenauigkeit arbeitenden, jedoch sehr teuren Laserscannern gefunden werden.

Das zu erstellende Modul soll der Kartographisierung der Umgebung autonom oder teilautonom agierender, mobiler Robotersysteme dienen und dabei eine Messungenauigkeit von circa einem Zentimeter aufweisen. Die Struktur, Farbe oder das Material der Objekte im Aktionsradius sowie die Umgebungshelligkeit und Ausleuchtung der selbigen sollen keinen Einfluss auf die Messergebnisse haben.

Um den Stromverbrauch des Moduls zu minimieren ist ein Energiemanagement zu integrieren. Dieses aktiviert die benötigten und deaktiviert die nicht benötigten Hardwarekomponenten.

## Eidesstattliche Erklärung

Hiermit versichere ich, dass ich die vorliegende Diplomarbeit ohne Hilfe Dritter und nur mit den angegebenen Quellen und Hilfsmitteln angefertigt habe. Diese Arbeit hat in gleicher oder ähnlicher Form noch keiner Prüfungsbehörde vorgelegen.

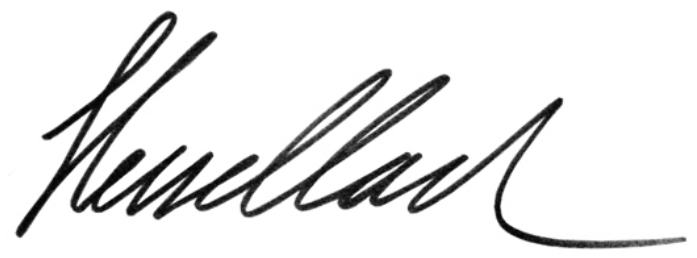

Senftenberg, November 2005 Dirk Hesselbach

## Danksagung

An dieser Stelle möchte ich mich bei einigen Menschen bedanken, welche mich bei der Erstellung dieser Diplomarbeit unterstützt haben und mir ihre wertvolle Zeit widmeten.

Mein erster Dank gilt Herrn Dipl.-Ing.(FH) Kai-Uwe Irrgang welcher mir beim Erstellen der Fräsdaten und beim Fräsen des 2. Prototypen geholfen hat. Des Weiteren bedanke ich mich herzlich bei Herr Dipl.-Ing.(FH) René Noack aus dem Fachbereich E-Technik, welcher die erste Version des 3. Modulprototypen im Leiterplattenlabor gefertigt hat.

Ein weiterer Dank gebührt Markus Urbanski und Kai-Uwe Steuer, welche mich bei der Endkorrektur dieser Arbeit unterstützten und sich in ihrer Freizeit durch dieses Schriftstück gearbeitet haben. Sie halfen mir noch ein paar Schusselfehler und Schreibfehler zu beseitigen.

Die meiste Unterstützung erhielt ich jedoch von Prof. Dr.-Ing. Ernst Reinhold welcher mehrmals die Rohfassung der Diplomarbeit las und mir sehr nützliche Hinweise und Verbesserungsvorschläge unterbreitete. Durch ihn konnte ich die Passagen über die Steuerungs- und Regelungselemente des Moduls optimal gestalten.

Ähnliches gilt auch für Prof. Dr. Wolfgang Laßner, welcher mich auf dem Gebiet des Softcomputings und der Mathematik beriet.

# Inhaltsverzeichnis

# Abbildungsverzeichnis

# Tabellenverzeichnis

# 1 Einleitung

## 1.1 Motivation

Interessiert man sich als Schüler oder Student für das Gebiet der Robotik, bieten einige Firmen heute gute Grundsysteme an. Sie bieten die Möglichkeit ein grundlegendes Verständnis für Sensoren, Aktoren und deren Zusammenwirken zu entwickeln.

Ein solches System ist beispielsweise der CCRP5 der Conrad Elektronik GmbH. Es ist *„[…] ein programmierbarer Kleincomputer der mit mit [!] zahlreichen Sensoren bestückt und auf einem Raupenfahrgestell montiert ist. CCRP5 ist, entsprechend programmiert, ein voll funktionsfähiger Kleinroboter [,] der auf Umweltreize ansprechen und reagieren kann."* [Conrad 2003]

Als Einstieg in die Robotik ist der CCRP5 geeignet, da man bereits ohne Erweiterungen alle grundlegenden Funktionen nutzen kann. Es sind ebenfalls einfache Sensoren vorhanden. Sollten diese nicht mehr ausreichen, kann das CCRP5-System durch eigene Schaltungsideen mit einer separat erhältlichen Experimentierplatine erweitert werden.

Möchte man sich jedoch näher mit der Materie beschäftigen, wird man sehr schnell an die Grenzen dieser Hardware stoßen. Sie bieten leider nur beschränkte Erweiterungsmöglichkeiten, sowohl auf dem Gebiet der Hard- als auch der Software. Der Energievorrat ist ebenfalls begrenzt. Jede Erweiterung des Systems verringert deshalb die Betriebszeit.

Statt ein Fertigsystem zu nutzen, besteht die Möglichkeit einen eigenen Roboter zu entwickeln. Er wird im Bereich der Energieversorgung und Erweiterbarkeit großzügiger dimensioniert, als es bei einem kommerziellen System der Fall ist. Dadurch stehen für Erweiterungen mehr Systemressourcen zur Verfügung.

Daraus entstand ein Konzept für ein System, welches kostengünstig nachgebaut, weiterentwickelt und erweitert werden kann. Ein Modul soll dabei mit Platinenfertigung und allen Bauteilen nicht mehr als 100€-150€ kosten. Die Module weisen standardisierte Schnittstellen auf, damit alle Module zueinander kompatibel sind. Aus all diesen Überlegungen entstand der „Experimentelle Mini Roboter", im Folgenden kurz EMR genannt.

Da für jeden autonom oder teilautonom agierenden Roboter, also auch dem EMR, die Orientierung im Raum für die Erfüllung seiner Aufgaben notwendig ist, sollte ein preiswerter, kleiner und leichter Umgebungsscanner entwickelt werden. Dieser ist Gegenstand dieser Diplomarbeit. Durch die Nutzung des $I^2C$-Busses soll das Umgebungserfassungssystem nicht nur im EMR sondern in jedem System mit diesem Busstandard einsetzbar sein.

## 1.2 Gliederung der Arbeit

Bevor im Kapitel 6 im Detail auf den Aufbau und die Funktionsweise des Scanners eingegangen wird, werden die nötigen Grundlagen beschrieben.

Im Kapitel 2 wird ein allgemeiner Überblick und Vergleich über die in der Robotik verwendeten Messsensoren gegeben. Dabei wird ihre Funktionsweise kurz erklärt und auf die Vor- und Nachteile der einzelnen Typen eingegangen. Durch den Überblick, soll verdeutlicht werden, aus welchen Gründen bestimmte Sensorklassen für das Umgebungserfassungssystem nicht in Betracht kommen und die Wahl des Sensors nachvollziehbar machen.

Danach (im Kapitel 3) wird auf die mechanischen Komponenten des Scanners eingegangen. Dazu zählen der Schrittmotor, das Untersetzungsgetriebe und die Sensorplattform.

Die folgenden Kapitel 4 und 5 beschreiben die modulexterne Software. Das Trainingsprogramm und die Grundlagen der Trainingsdatengewinnung stellt der Abschnitt 4 dar. Im anschließenden Kapitel (5) wird das Testprogramm für das Modul vorgestellt. Es dient einerseits zum Prüfen der Funktionsfähigkeit und stellt andererseits die im Kapitel 5.1 beschriebenen lokalen Karten dar.

Das Kapitel 5.1 vermittelt, wie die Daten des Umgebungsscanners verarbeitet und als Datensatz einer Karte der örtlichen Umgebung des Scannbereiches (lokale Karte) umgerechnet werden. Die Erstellung eines Plans der gesamten Umgebung (globale Karte) aus den lokalen Karten wird im Anschluss beschrieben. Abschnitt 5.2 ist ein Ausblick auf die Datenweiterverarbeitung der lokalen Karten im Roboter und als Anwendungsbeschreibung des Scanners zu sehen.

Nach den theoretischen Aspekten des Systems wird im Kapitel 6 die Schaltung und Konzeption der Modulhardware und die Erstellung und der Funktionsumfang der Modulsoftware dargestellt.

Im Anhang und auf dem beiliegendem Datenträger sind weiteres Material, Detailbeschreibungen, sowie eine multimedial aufbereitete Darstellung der Diplomarbeit mit Ergänzungen (auf CD) verfügbar.

# 2 Sensoren

Sensoren sind in der Robotik essentiell wichtig und unentbehrlich geworden. Erst sie bieten halb- oder vollautonomen Systemen die Möglichkeit, Informationen über ihre Umwelt oder ihren eigenen Systemzustand zu sammeln.

In der Praxis werden mehrere Sensorarten gleichzeitig genutzt. Durch den Einsatz verschiedener Typen ist es möglich, die Schwachstellen der genutzten Sensoren auszugleichen und die Vorteile verschiedener Sensorklassen zu nutzen. Damit werden lückenlose Informationen, trotz der Selektivität der Sensoreneigenschaften, gesammelt.

Die Orientierung in der Umgebung ist ein wichtiger Aspekt des Sensoreneinsatzes. Er soll es den Robotern ermöglichen auf unerwartete Änderungen in der Umwelt zu reagieren und sich in dieser zu orientieren. Sensoren sind in der Robotik dass, was für den Menschen seine Sinne sind. Der Einsatz von Sensoren ist unumgänglich, um die ausgeführten Aktionen und Reaktionen des Systems kontrollieren und steuern zu können oder Vermessungsaufgaben durchzuführen.

## 2.1 Sensorarten für die Entfernungsmessung

Im zu konzipierenden Umgebungserfassungssystem kommen Sensoren zur Entfernungsmessung zum Einsatz. Es existieren verschiedene Sensorklassen mit denen Abstandsmessungen möglich sind. Im Folgenden werden diese kurz vorgestellt und ihre Vor- und Nachteile beschrieben, welche Einfluss auf die Wahl des Sensors hatten.

### 2.1.1 Infrarotsensoren

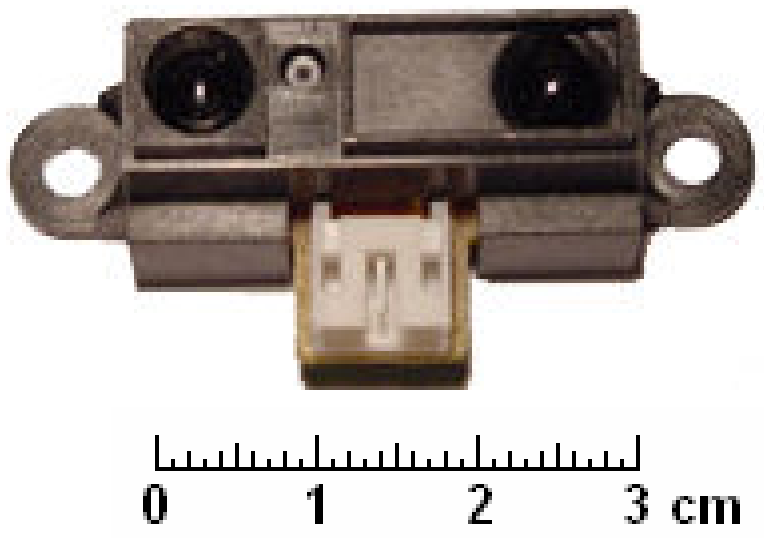

Abbildung 1 - Sharp Infrarotsensor GP2D12

Infrarotsensoren (im Folgenden als IR-Sensoren bezeichnet) werden im Bereich der Robotik zur Entfernungsmessung eingesetzt. Sie enthalten eine Sendediode (welche ein moduliertes Signal aussendet), einen „Position Sensitiv Detector“ (PSD) und eine Auswerteelektronik. Der Sensor arbeitet nach dem Triangulationsprinzip.

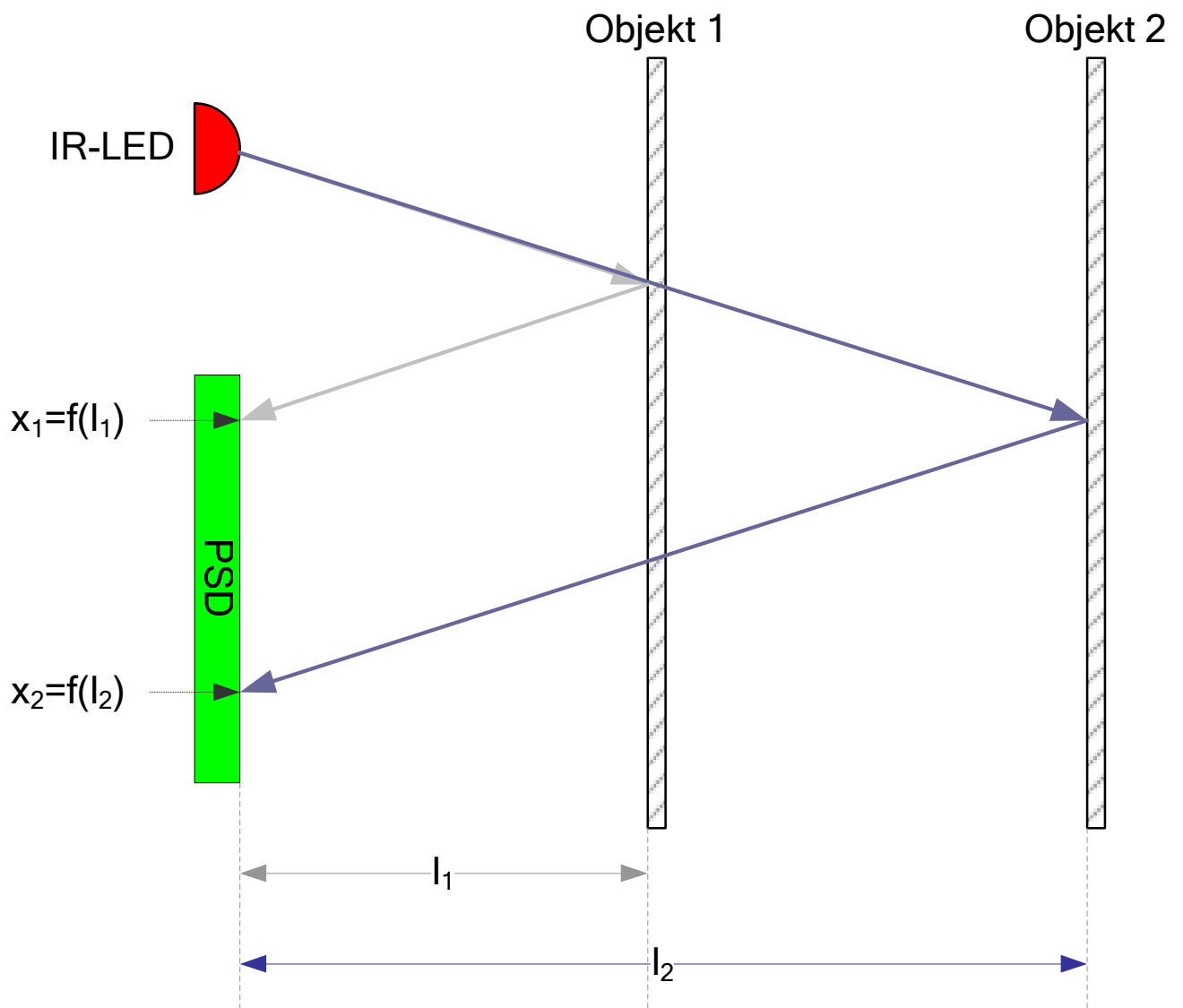

Abbildung 2 - Triangulationsprinzip

Der Auftreffpunkt des reflektierten Infrarotsignals auf dem PSD ist abhängig von der Objektentfernung. Das vom Infrarot-Entfernungsmesssensor ausgegebene Signal ist eine Funktion der Entfernung. Im Anhang befinden sich detaillierte Spannungs-Entfernungsdiagramme, welche diesen Sachverhalt mit Hilfe von Messdaten verdeutlichen.

In Abbildung 3 ist das Blockschaltbild des Sensors mit der Infrarot-LED und PSD dargestellt.

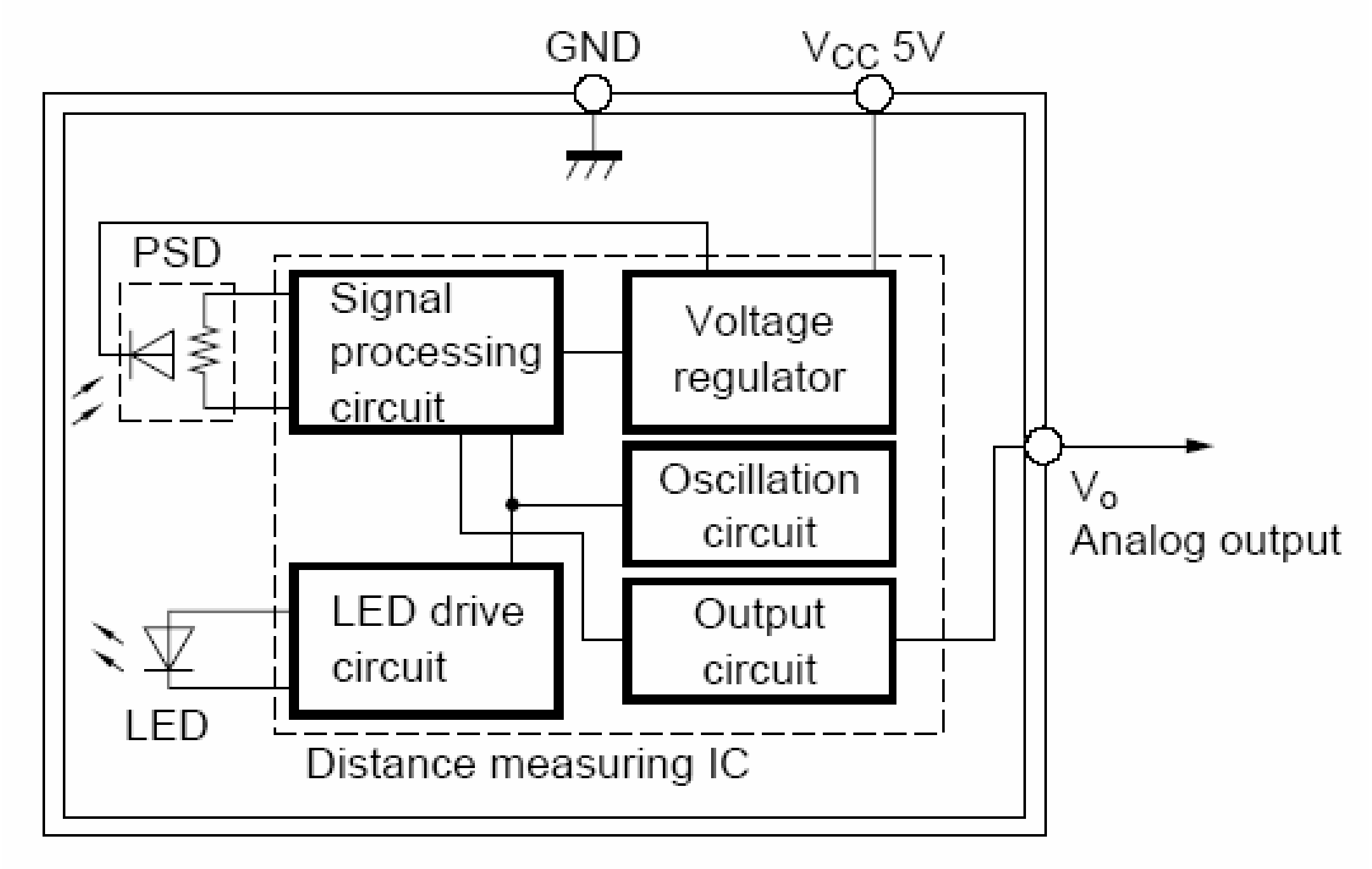

Abbildung 3 - Blockschaltbild des Sharp GP2D120 [Sharp 2000a]

Die Sharp Electronics (Europe) GmbH (im Folgenden kurz Sharp genannt) beispielsweise, bietet ein vielfältiges Angebot solcher Infrarot-Entfernungsmesssensoren. Die Toleranz wird vom Hersteller mit etwa 0,5cm bis 40cm und 2,0cm ab 40cm (für den GP2D12) Messentfernung angegeben.

Bei einer Umgebungsausleuchtung ab circa 7000Lux verringert sich die Genauigkeit auf circa 2,0cm bis 30cm und ab 30cm auf 5,0cm. Die Abnahme der Messgenauigkeit wurde nur bei direktem Lichteinfall von Sonnenlicht in den IR-Sensor erreicht. Bei einem Einsatz der Sensoren in normal beleuchteten Innenräumen wurden keine Messungenauigkeiten festgestellt.

Die Sensoren werden über ein  dreiadriges Kabel (Typen mit analogem Ausgang, mit der Belegung Masse, Betriebsspannung, Signalpin) oder einem vieradrigem Kabel (Typen mit digitalem Ausgang, mit der Belegung Masse, Betriebsspannung, Takt, Signalpin) angeschlossen. Die Abbildung 50 im Kapitel 6.5.1 stellt den Anschluss des IR-Sensors detailliert dar.

Am Signalpin liegt, je nach verwendetem, analogem Sensortyp, eine Spannung zwischen 0V und 5V an. Sie ist jedoch nicht linear proportional zur gemessenen Entfernung, wie beispielsweise den Datenblättern zu entnehmen ist.

Sharp stellt auch eine Serie von IR-Sensoren mit digitalem Ausgang her, welche je nach eingestellter Entfernung ein Hi oder Low (1bit) am Signalpin ausgeben. Diese Typen werden als Grenzwertschalter genutzt und kommen beispielsweise in automatischen Händetrocknern zum Einsatz. Diese digitale Version wird in der Robotik kaum verwandt, da mit ihnen eine genaue Entfernungsmessung zum Objekt nicht möglich ist.

Der Sensortyp wird ebenfalls als 8bit Version angeboten. Sie geben die Entfernung seriell mit Hilfe eines Taktsignals aus. Vom Aufbau und Betriebsverhalten sind sie den Typen mit analogem Ausgang gleichzusetzen. Der einzige Unterschied besteht darin, dass bei den digitalen IR-Sensoren der Analog/Digital-Wandler bereits integriert ist. Diese können an Microcontrollern ohne AD-Wandler angeschlossen und genutzt werden.

Ein großer Vorteil, sowohl der analogen, als auch der digitalen Typen, ist die bereits im Sensor integrierte Auswertungselektronik. Sie moduliert das ausgesendete Infrarot-Signal und filtert es beim Empfang, um den Einfluss von Fremdlicht zu eliminieren. Die mit dem PSD gemessene Entfernung wird von der Elektronik danach als analoge Spannung oder digitales Signal ausgegeben.

Durch den geringen Abstrahlwinkel des Infrarotstrahls und der Unabhängigkeit von den physikalischen Umgebungsgrößen, eignen sich IR-Sensoren für Vermessungsaufgaben. Objekte können daher nur in einem Winkelbereich von circa 1,5° erkannt werden. Zur Hinderniserkennung während der Bewegung des Roboters sollten Sensoren mit größerem Erfassungswinkel, beispielsweise Ultraschallsensoren, gewählt werden.

### 2.1.2 Ultraschallsensoren

Der Vorteil von Ultraschallsensoren ist deren hohe Messreichweite und großer Objekterfassungswinkel. Zum Teil, je nach eingesetztem Modul oder Schaltung, können Objekte in bis zu 10m Entfernung geortet werden. Die Genauigkeit ist von der eingesetzten Auswertungselektronik abhängig und liegt in der Regel bei unter 1cm.

Wie exakt der Abstand zum Objekt gemessen wird, hängt von den physikalischen Umgebungsgrößen ab. Dazu zählen Temperatur, Luftfeuchtigkeit und Druck. Sie beeinflussen die Signallaufzeit und müssen in der Berechnung berücksichtigt werden, wodurch sich jedoch der Schaltungsaufwand erhöht.

Wie auch bei den Infrarotsensoren gibt es bei den Ultraschallsensoren Firmen, welche komplette Module inklusiver der Auswertungselektronik anbieten.

Ein in der Robotik oft eingesetztes Modul wird von der Firma Polaroid hergestellt. Er arbeitet mit einem Sender, der nach dem Abstrahlen des Impulses als Empfänger genutzt wird.

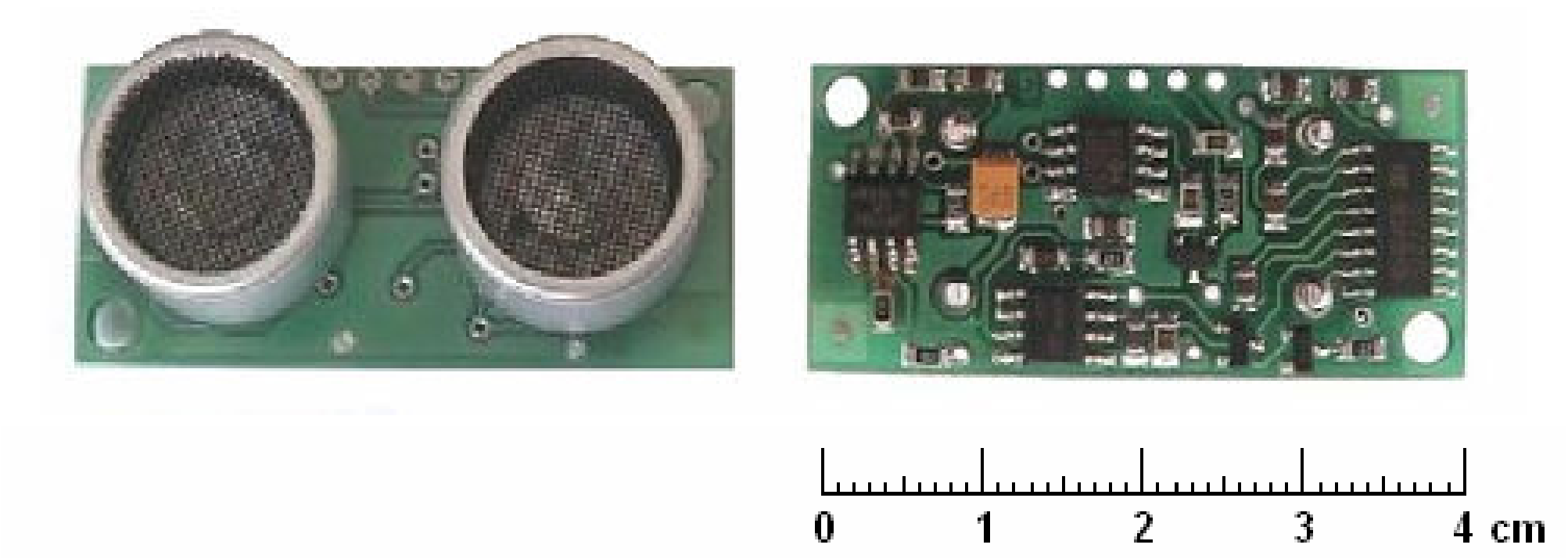

Abbildung 4 - Ultraschallmodul SRF04 der Firma Devantech

Die Firma Devantech hingegen bietet Ultraschallmodule mit getrennten Sender- und Empfängerkapseln an (siehe Abbildung 4).

Es ist jedoch ohne Aufwand möglich Schaltungen im Internet zu finden. Sie sind preisgünstiger im Nachbau, als die erhältlichen Fertigmodule. Eine Erweiterung oder Modifikation für den entsprechenden Anwendungszweck ist, im Gegensatz zu den Fertigmodulen, problemlos möglich.

Die im Handel erhältlichen Ultraschallmodule geben nur den Entfernungswert des Objektes aus, welches dem Sensor am nächsten liegt. Einige Schaltungen aus dem Internet können alle Objekte im Empfangsbereich erkennen.

Mit mehreren Empfängern und einem Sender ist hier sogar die Positionsbestimmung eines Objektes mit Hilfe des Sonarprinzips (Sound Navigation And Ranging) möglich.

Dabei wird ein Schallimpulse (Ping [in Abbildung 5 grün dargestellt]) ausgesendet und die Zeit bis zum Eintreffen seines Echos gemessen. Aus der Laufzeit errechnet sich die Entfernung [in Abbildung 5 Entfernung 1 und 2] bis zum Objekt. Der Abstand zum Objekt wird mit jedem Empfänger gemessen. Diese Entfernungen stellen die Seiten eines Dreiecks dar, wodurch die Lage des Hindernisses bestimmt wird.

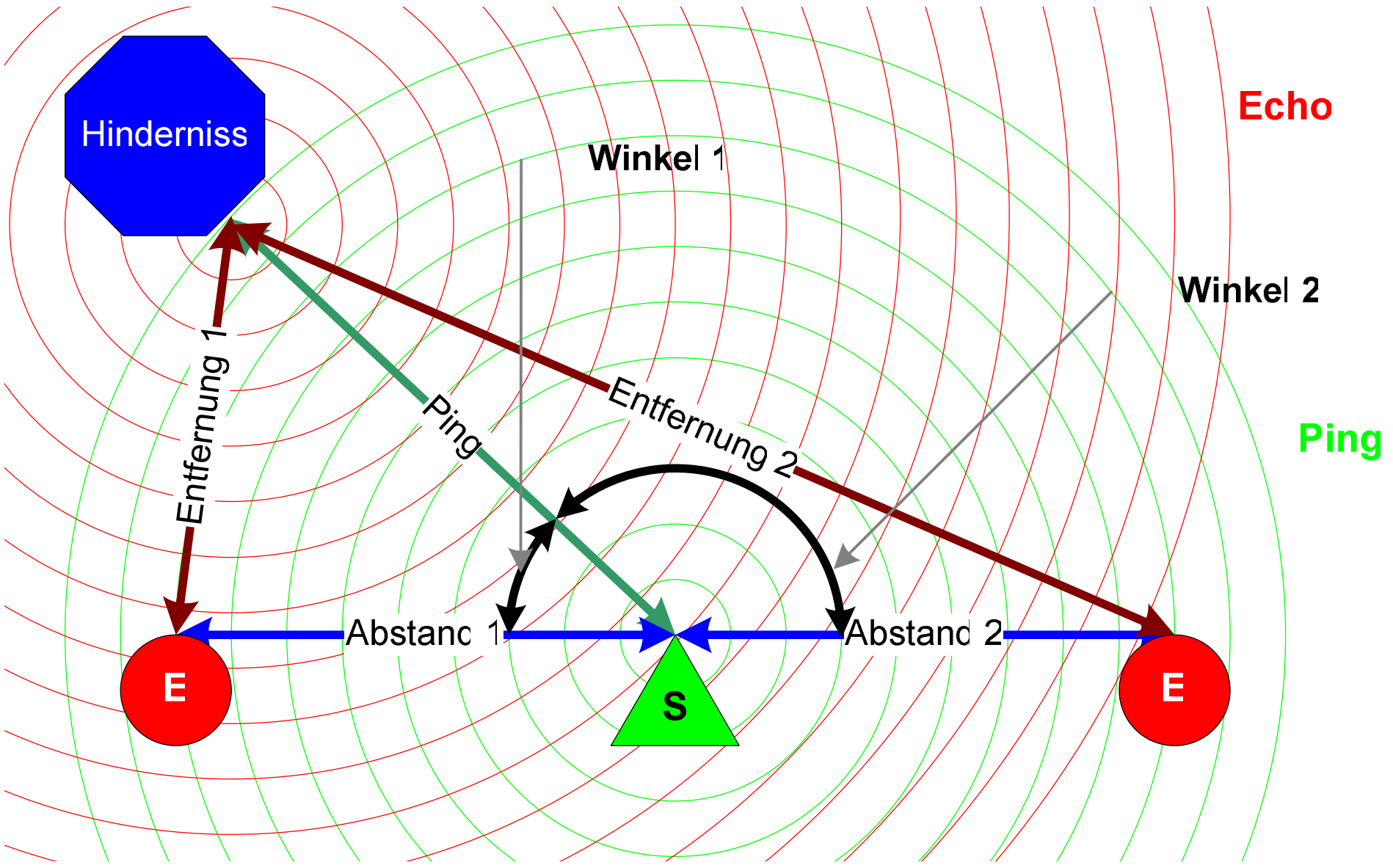

Abbildung 5 - Positionsbestimmung eines Objektes per Doppelsonar

Es ist ein hoher Schaltungsaufwand und ein zweiter Empfänger erforderlich [in Abbildung 5 mit „E“ gekennzeichnet], um neben der Entfernung auch die Position des Objektes zu erfassen.

Die Berechnung der Entfernung und der Winkel relativ zum Sender erfolgt mit Hilfe trigonometrischer Funktionen (Winkelfunktionen). Vor der eigentlichen Berechnung ist jedoch eine aufwendige Datenaufbereitung notwendig. Darauf soll an dieser Stelle jedoch nicht im Detail eingegangen werden.

Das reflektierte Signal ist sehr stark von den Eigenschaften der Objekte im Messbereich, den Reflexionswinkeln und den Signalechos aus der Umgebung abhängig.

Die Störeffekte herauszufiltern erfordert einen hohen technischen Schaltungsaufwand. Um weitere Informationen zu erhalten, sollte entsprechende Fachliteratur zu Rate gezogen werden.

Ultraschallsensoren eignen sich zum Erkennen von Hindernissen, auch bei großen Messentfernungen und in einem großen Winkelbereich. Für die exakte Vermessung der Umgebung sind sie durch den großen Abstrahlwinkel des Signalimpulses ungeeignet.

### 2.1.3 Laserscanner

Laserscanner kommen beispielsweise bei der Erstellung von 3D-Abbildungen durch positionsgenaue Realisierung von Flächenscans unter vorgegebenen Winkeln, der Digitalisierung von Innenräumen für Bau- und Fabrikplanung, der räumlichen Lageerfassung von beweglichen Gütern im Logistikbereich, der Überwachung sicherheitsrelevanter Bereiche an bewegten Systemen, der Kontrolle der Fahrwege bei fahrerlosen Transportsystemen zur Kollisionsvermeidung und der Umgebungserfassung bei mobilen Robotern zur Selbstorientierung und Navigation zum Einsatz.

Es existieren auf dem Markt verschiedene Produkte. Dazu zählen zum Beispiel Citygrid der Firma Geodata Ziviltechnikergesellschaft mbH und PowerCube von der Sick AG.

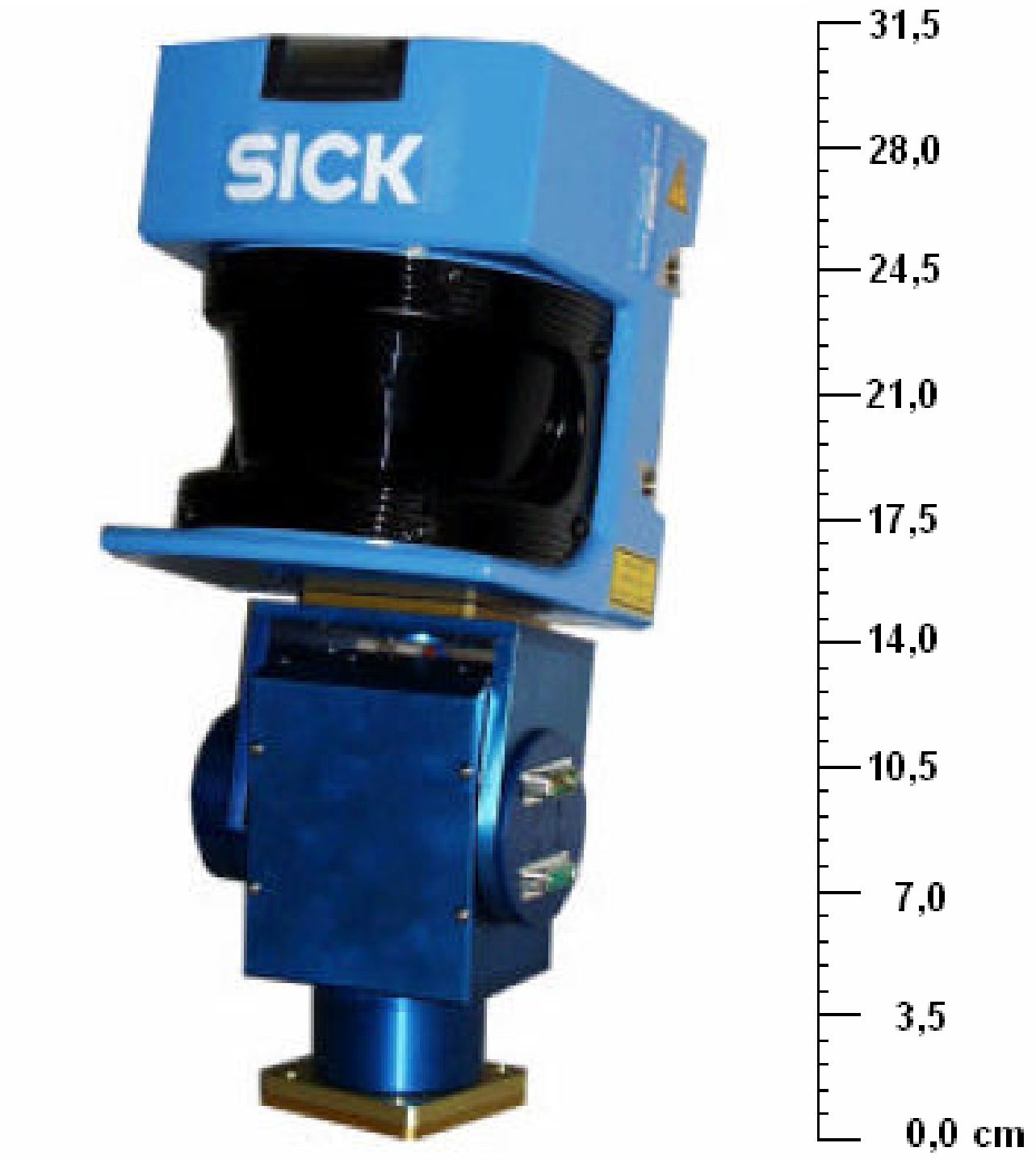

Abbildung 6 - Laserscanner PowerCube der Sick AG

Die Reichweite der Laserscanner liegt, je nach Hersteller, bei mehreren hundert Metern. Je nach eingesetzter Auswertungshardware können Auflösungen im Millimeterbereich erreicht werden.

Für den Einsatz von Laserscannern im kommerziellen Bereich sind die Kosten solcher Systeme zurzeit noch zu hoch. Sie liegen bei 10.000€ bis 50.000€. Dabei entfällt ein Großteil der Kosten auf die optischen Komponenten (Spiegel, Prismen, Linsen,…) und die Auswertungsschaltung.

Zur Umgebungserfassung sind Laserscanner sowohl in der Genauigkeit, als auch in der Vermessungsgeschwindigkeit den anderen Sensoren überlegen. Gegen den Einsatz in kleinen bis mittleren Robotern sprechen der hohe Preis und das Gewicht. Ihre Einsatzfelder sind zurzeit in der Messtechnik oder automotiven Robotik zu suchen.

### 2.1.4 Kameramodule

Im Gegensatz zu Laser-, Ultraschall- und Infrarotsensoren liefern Kameramodule ein Bild der Umgebung und keine Entfernungsdaten der Objekte. Dieses müssen mit Hilfe von Bilderkennungsalgorithmen gewonnen werden.

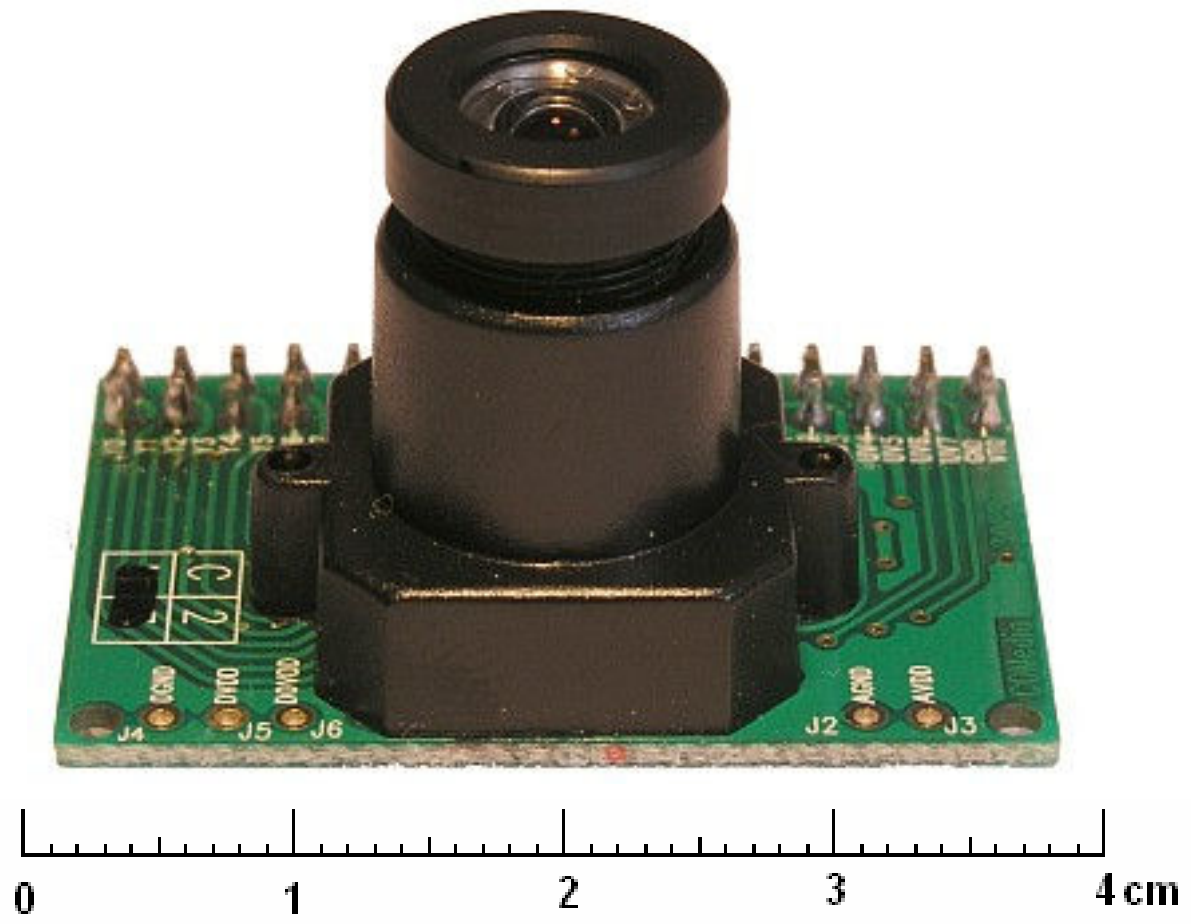

Abbildung 7 - Kameramodul

Eine gebräuchliche Methode die Abstände vom Roboter zum Hindernis zu berechnen, ist die Nutzung zweier Kameras. Dieses Verfahren wird als Stereoskopie bezeichnet.

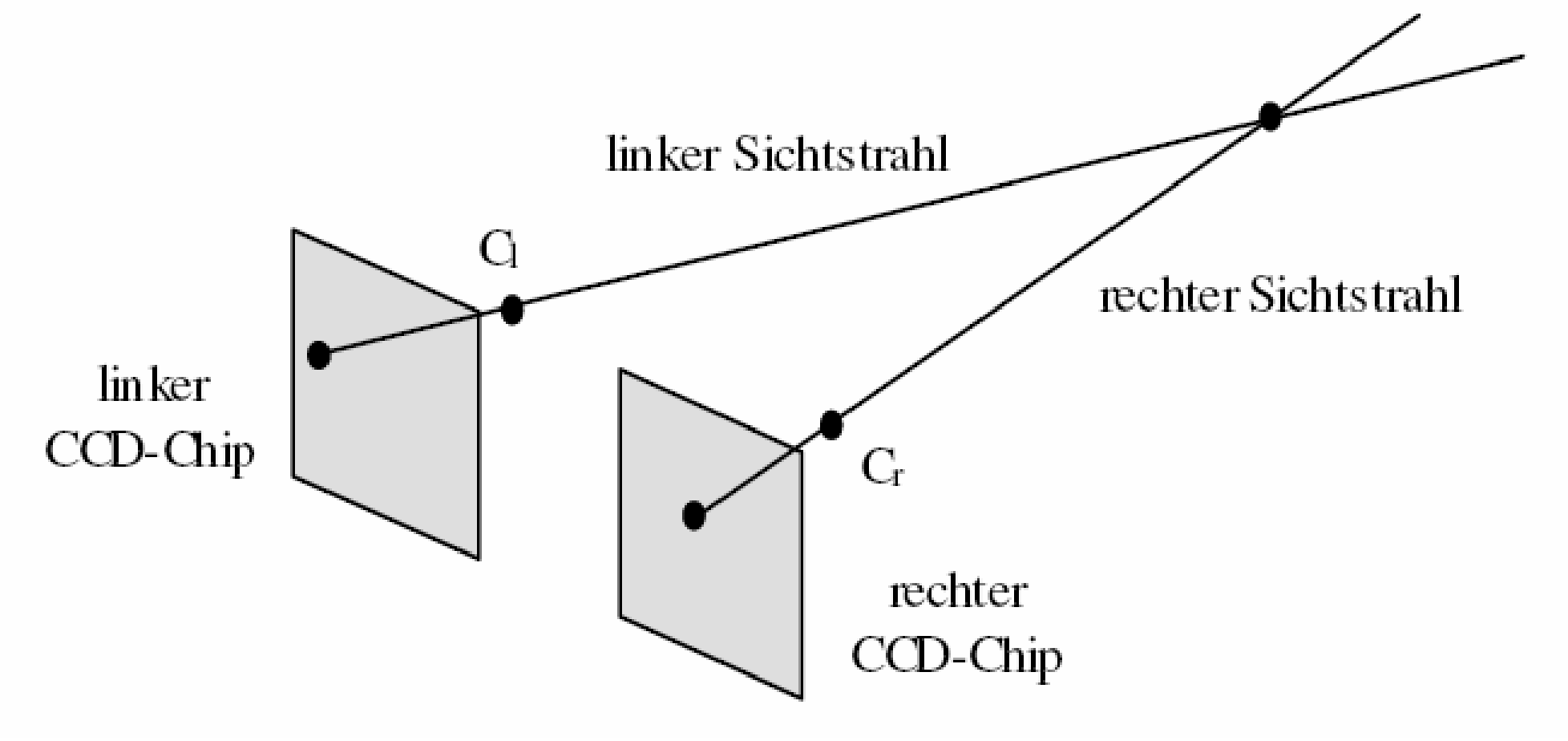

Abbildung 8 - Bildpunkte bei der Stereoskopie

Dabei wird ein Objektpunkt auf verschiedene x/y-Koordinaten der beiden Kamera-CCD-Chips projiziert (siehe Abbildung 9 - Bild der beiden Kameras bei der Stereoskopie).

Die Lage der optischen Zentren sind aus der Kamerakalibrierung bekannt, ebenso ist die Punktkorrespondenz (Bildpunkte beider Kameras, welche denselben Objektpunkt repräsentieren) mit Hilfe der Disparitätsschätzung ermittelt worden. Sie dient dazu, Punkte im linken und rechten Bild zu finden, welche den gleichen Bildinhalt beschreiben. Die Strahlen werden von dem CCD-Chip der Kameras durch das Opti-sche Zentrum ($C_L$, $C_R$) hindurch verlängert und zum Schnitt gebracht. Das Ergebnis hierbei sind reale dreidimensionale Koordinaten, die den Abstand von einem Hindernis zum Mittelpunkt des Sensors in Zentimetern angeben.

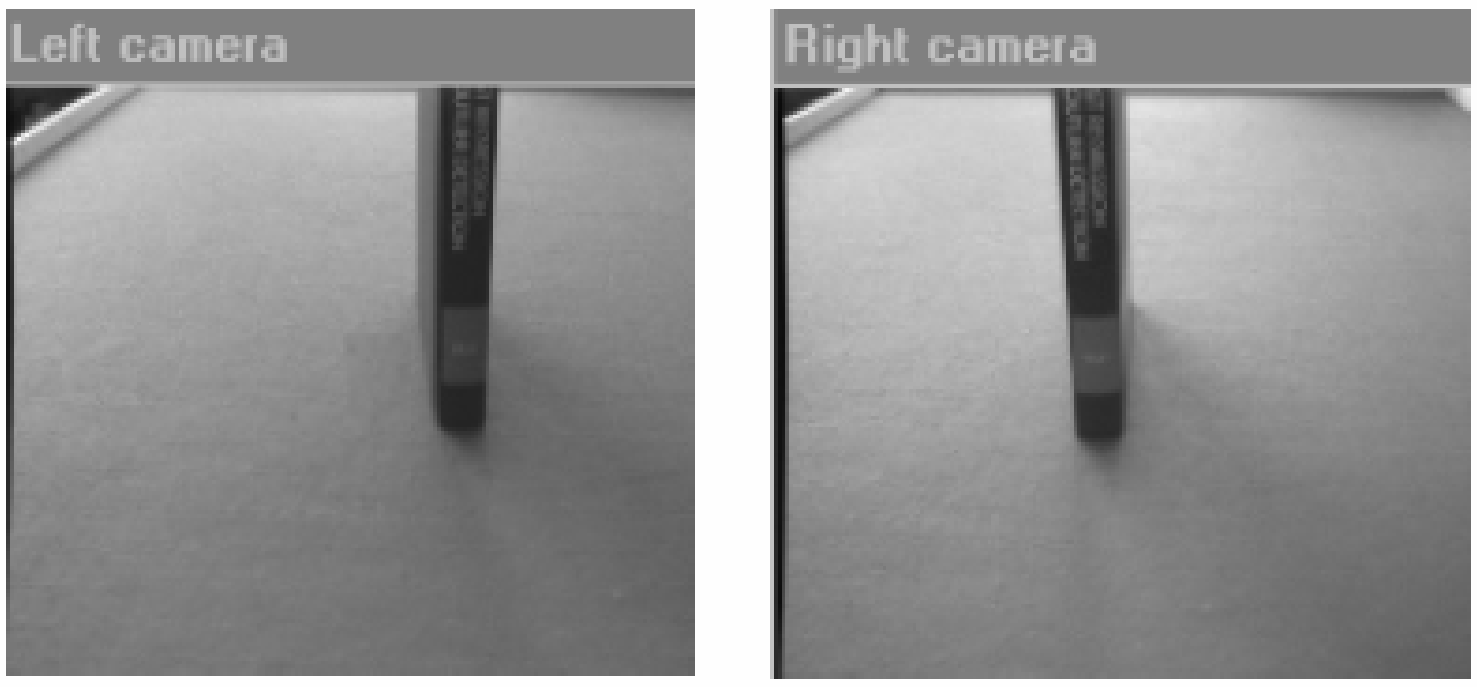

Abbildung 9 - Bild der beiden Kameras bei der Stereoskopie

Die Datengewinnung mit dieser Prozedur ist allerdings sehr rechenintensiv, wodurch zur Bildauswertung leistungsstarke Rechentechnik benötigt wird. Eine detaillierte Beschreibung des Verfahrens kann in den entsprechenden Quellen nachgelesen werden.

### 2.1.5 Sensoren im Vergleich

Jede der vorgestellten Sensorklassen ist prinzipiell zur Umgebungserfassung geeignet. In Tabelle 1 sind deren Eigenschaften im Überblick dargestellt.

| Sensortyp | Kosten | Aufwand für Datenauswertung | Gewicht | Liefert Daten über |
|---|---|---|---|---|
| Infrarot | gering (10€ bis 30€) | gering (Microcontroller) | gering (20g bis 50g) | Distanz (bis 150cm) |
| Ultraschall | gering (40€ bis 60€) | gering/mittel (Microcontroller) | gering (20g bis 100g) | Distanz (bis max. 10m) |
| Kamera | mittel (20€ bis 150€) | mittel/hoch (Embedded PC, PC) | gering (50g bis 250g) | Farbe, Helligkeit, Distanz (mit hohem Aufwand) |
| Laserscanner | hoch (10.000€ bis 50.000€) | hoch (leistungsstarker PC) | hoch (15kg bis 35kg) | Distanz Reflexionseigenschaften (mehrere hundert Meter) |

Tabelle 1 - Sensortypenvergleich

Um die Entscheidung für einen Sensortyp zu treffen, müssen Eigenschaften wie Preis, Gewicht, Messreichweite, Stromverbrauch, Aufwand für die Datenverarbeitung und Umwelteinflüsse auf den Sensor berücksichtigt werden.

Da das Umgebungserfassungsmodul für kleine bis mittlere mobile Roboter konzipiert wurde, sind der Stromverbrauch und das Gewicht, sowie die physikalischen Abmaße des Sensors die wichtigsten Kriterien. Durch die genannten Merkmale, den hohen Preis und dem enormen Aufwand zur Datenauswertung, war ein Laserscanner für das Umgebungserfassungsmodul ungeeignet.

Kameramodule sind für ungünstige Lichtverhältnisse und Dunkelheit nicht geeignet. Eine optimale Ausleuchtung der Umgebung wäre mit einem hohen Energieverbrauch verbunden und die Entfernungsberechnung durch Stereoskopie ist nur durch den Einsatz komplexer Schaltungen und teurer IC's möglich. Durch die extrem aufwendige Datenverarbeitung der erfassten Bilder kamen Kameramodule nicht in Betracht.

Zur Auswahl standen somit nur noch Ultraschall- und Infrarotsensoren. Beide Sensorklassen weisen ein ähnliches Preisniveau, Reichweite und Gewicht auf. Die ausschlaggebenden Kriterien waren damit die Eigenschaften von Infrarot- und Ultraschallsensoren.

Ultraschall hat den Nachteil, dass Reflexionen an allen Umgebungsobjekten entstehen und somit Echos und Interferenzen auftreten. Er kann ebenfalls absorbiert werden. Die Oberflächenbeschaffenheit der Objekte beeinflusst das Signal also sehr stark.

Infrarotstrahlung wird von den Eigenschaften der Oberflächen oder dem Material der Objekte kaum beeinflusst. Ein weiterer Vorteil ist der geringe Abstrahlwinkel. Während sich Ultraschall in alle Richtungen gleichmäßig ausbreitet, kann Infrarotlicht punktförmig abgestrahlt werden.

Die Wahl für den Messsensor im Umgebungsscanner fiel damit auf ein Infrarotmodul. Durch die bereits im IR-Sensor integrierte Auswerteelektronik und die Signalmodulierung der Infrarotstrahlung kam ein Fertigmodul der Firma Sharp zum Einsatz.

## 2.2 Infrarotsensormodultypen im Detail

Die Infrarotentfernungsmessmodule von Sharp lassen sich, wie im Kapitel 2.1.1 angedeutet, in zwei Klassen unterteilen. Sharp bietet Module mit digitalem oder analogem Ausgang an. Eine genaue Übersicht befindet sich im Kapitel 0.

Eine analoge Version des Infrarot-Entfernungsmesssensors ist der GP2D120. Die Funktionsweise entspricht der des GP2D02. Allerdings erfolgt die Ausgabe der Entfernung nicht als 8bit-Wert, sondern als analoge Spannung zwischen 0V und 5V (siehe Abbildung 10).

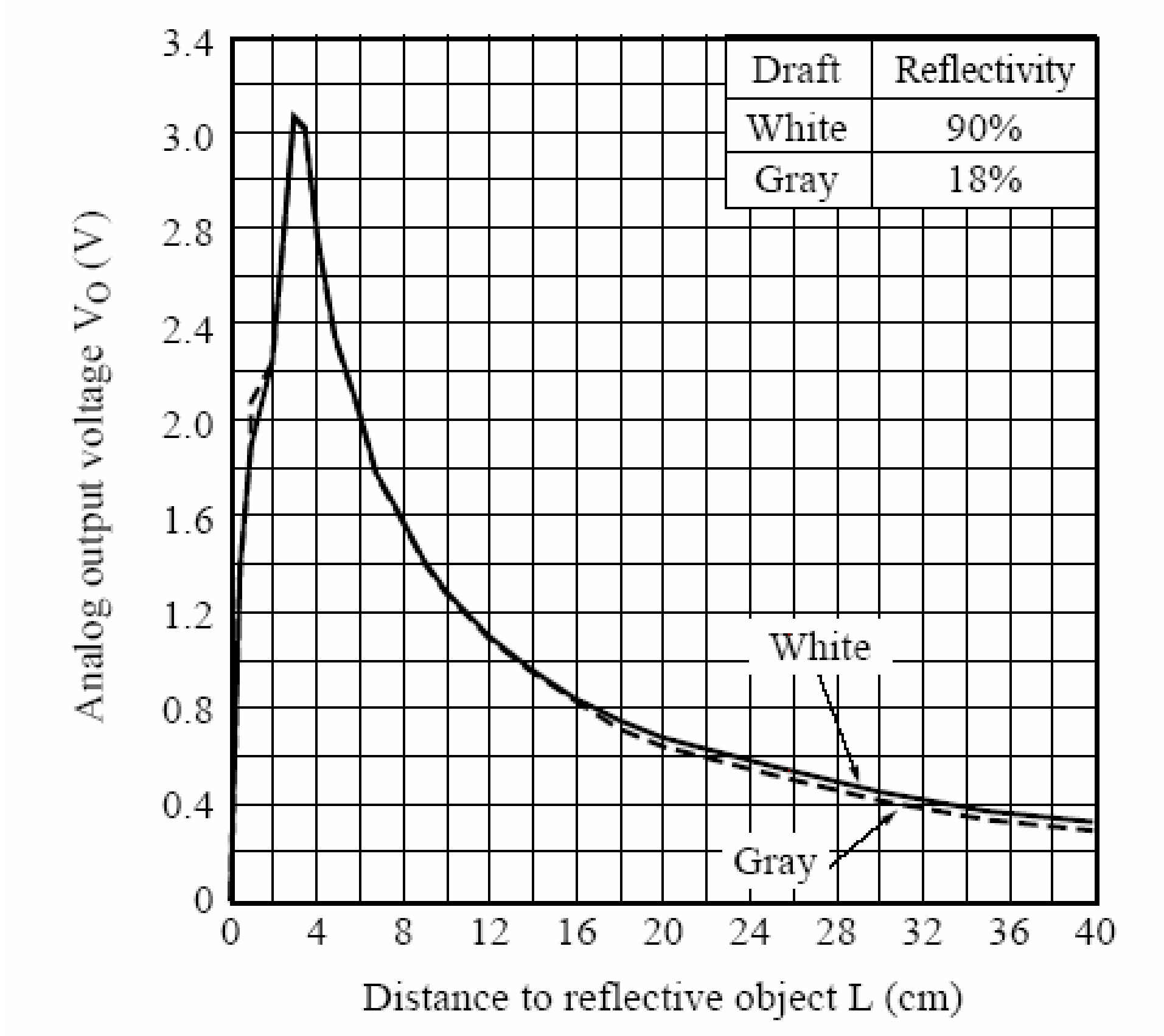

Abbildung 10 - Spannungs-Entfernungs-Diagramm des GP2D120 [Sharp 2000a]

Die Sensoren mit digitalem Ausgang sind in zwei Ausführungen zu unterteilen. Einige Typen sind mit einem seriellen Ausgang mit einer Genauigkeit von 8Bit ausgestattet. Sie können an Microcontrollern ohne AD-Wandler eingesetzt werden. Allerdings sind ist der ausgegebene Wert, wie bei allen Typen, nicht direkt proportional zur gemessenen Entfernung. Die Messgenauigkeit ist im Nahbereich höher, als bei weiter entfernten Objekten.

Der in Abbildung 11 dargestellte GP2D02 beispielsweise gibt die Entfernungsdaten seriell mit einer Auflösung von 8Bit aus. Diese werden mit einem Clock-Signal am „Control-Signal-Input“ über den $V_{out}$-Ausgang (Signalausgang) beispielsweise von einem Microcontroller abgefragt.

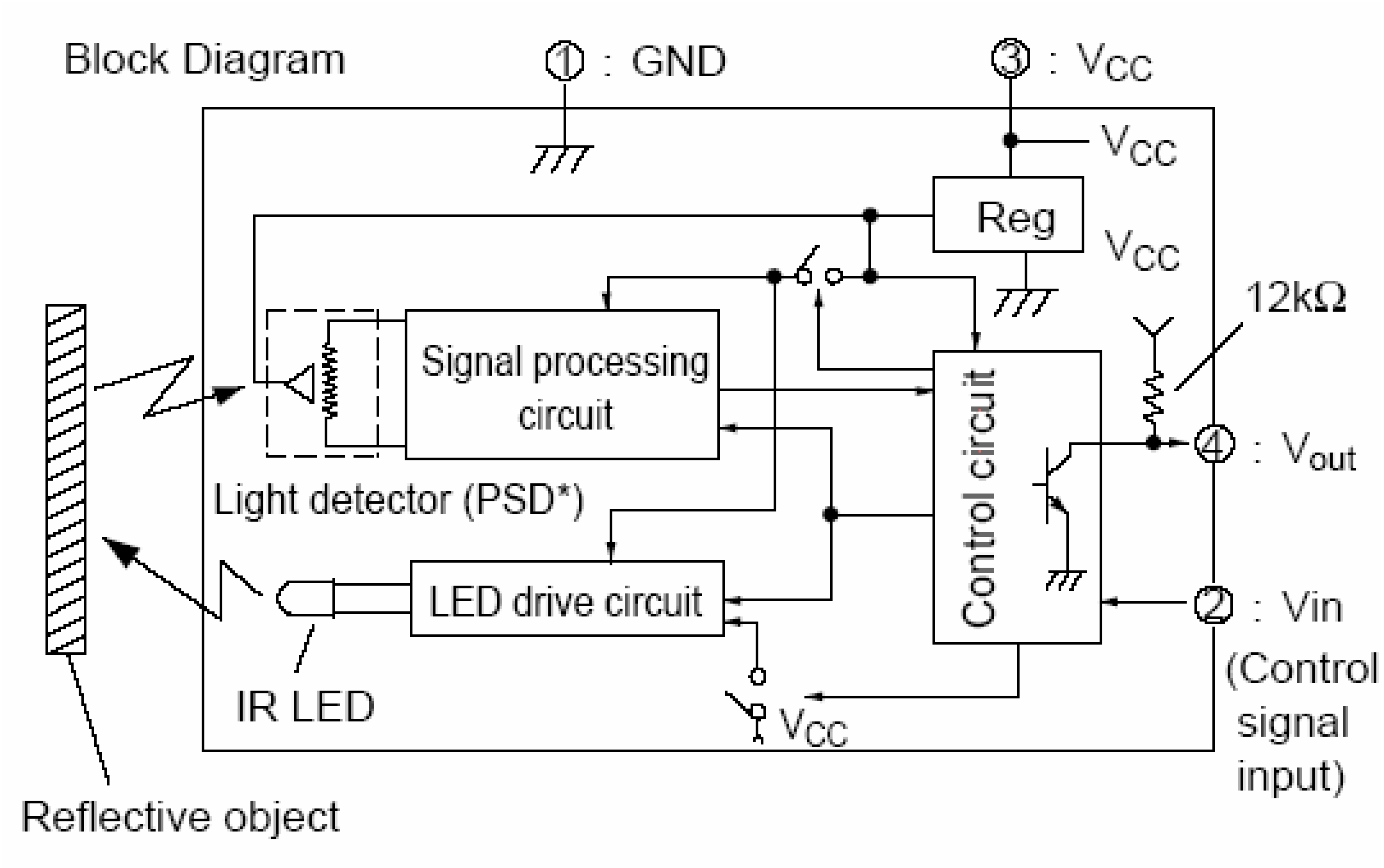

Abbildung 11 - Blockdiagramm des Sharp GP2D02 [Sharp 2000b]

Die Datenabfrage für den GP2D02 und Sensoren ähnlichen Typs stellt sich im zeitlichen Ablauf wie in Abbildung 12 veranschaulicht dar.

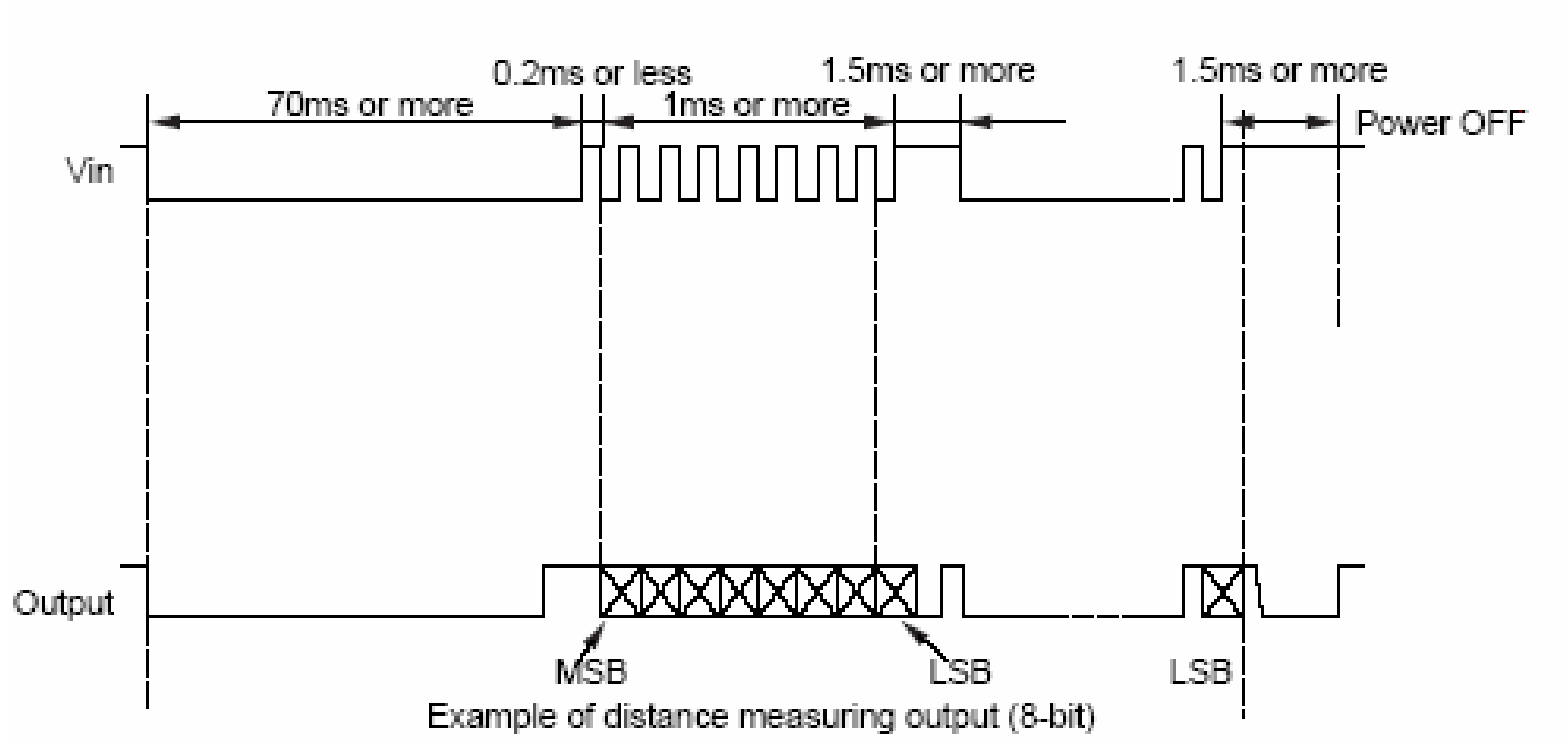

Abbildung 12 - Timing Chart

Zur Umwandlung des ADC-Wertes sowohl für Typen mit analogem, wie auch digitalem Ausgang, kann dieselbe Umwandlungsprozedur genutzt werden.

Die folgende Tabelle stellt die IR-Sensortypen von Sharp im Überblick mit der möglichen Messdistanz und der Art des Ausgangs dar.

| Sensortyp | Messdistanz | Ausgang | Auflösung |
|---|---|---|---|
| GP2D02 | 10-80cm | digital | 8bit |
| GP2D03 | 0-7cm | analog | 0-5V |
| GP2D05 | 10-80cm | digital | 1bit |
| GP2D12 | 10-80cm | analog | 1,75-2,25V |
| GP2D120 | 4-30cm | analog | 0,4-2,8V |
| GP2D15 | 10-80cm | digital | 1bit |
| GP2D150 | 3-30cm | digital | 1bit |
| GP2Y0A02YK | 20-150cm | analog | 0,4-2,75V |
| GP2Y0D02YK | 20-150cm | digital | 1bit |
| GP2Y0D340K | 10-60cm | digital | 1bit |
| GP2YA21YK | 10-80cm | analog | 0,4-2,6V |

Tabelle 2 - Infrarot-Sensor-Typenvergleich

Die 1-Bit-Typen sind für den Einsatz im Umgebungserfassungssystem ungeeignet.

Es bleiben die digitalen Typen mit einem 8-Bit-Ausgang oder jene mit analogem Ausgang. Da die Analog-Digital-Wandler im Microcontroller integriert sind, wird ein Infrarot-Sensor-Modul mit analoger Ausgabe genutzt.

Da sich Roboter in Räumen mit Abmessungen im Meter-Bereich bewegen, sollte ein Sensor mit möglichst großer Reichweite genutzt werden. Eine millimetergenaue Messung im Nahbereich ist dagegen unerheblich.

Die Entscheidung fällt nach dem Typenvergleich deshalb auf den GP2Y0D02YK. Messreihen, welche im Anhang abgebildet sind, ergaben sogar bis 200cm eine stabile Entfernungs-Spannungs-Messkurve.

## 2.3 IR-Sensoren und ihre Anordnung am Roboter

Das Umgebungserfassungsmodul verarbeitet die Messdaten der fest installierten und beweglich am Roboter angebrachten Infrarotsensoren (siehe Abbildung 13) mit der gleichen Prozedur. Bei dem beweglich angebrachten Sensor wird neben der Spannungs-Entfernungs-Umrechnungsprozedur noch die Schrittmotorsteuerprozedur aufgerufen. Die Prozeduren sind im Kapitel 6.6 beschrieben.

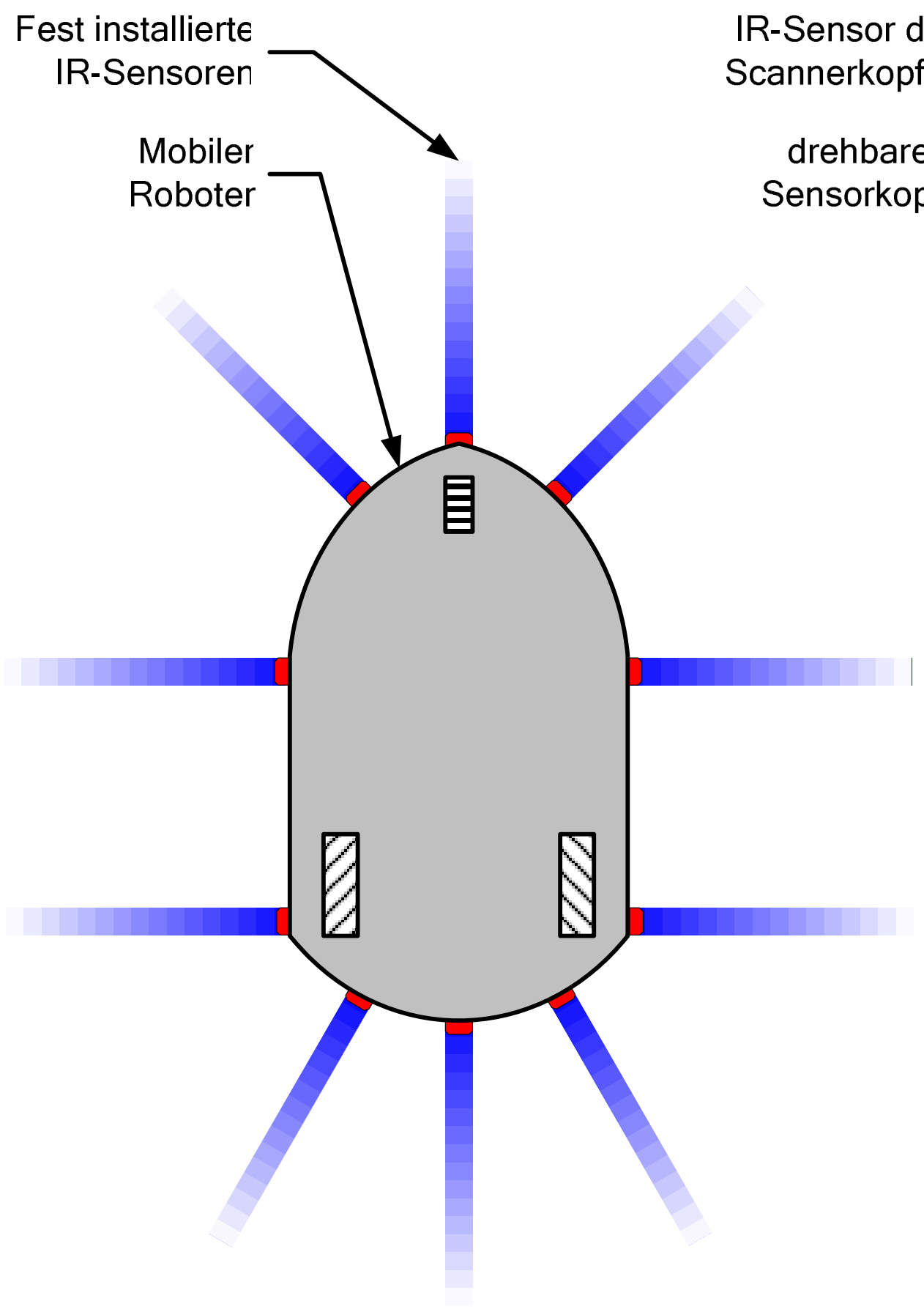

Abbildung 13 - IR-Sensoren und Sensorkopf am mobilen Roboter

Die fest am mobilen System installierten IR-Sensoren dienen zur Kollisionsvermeidung während der Bewegung. Deren Daten können ebenfalls zur Ausrichtung des Roboters, z.B. parallel zu einer Wand, eingesetzt werden. Das Umgebungserfassungssystem liefert dem Hostrechner des Roboters dafür die notwendigen Daten. Die Spannungswerte der am mobilen System befestigten IR-Sensoren werden durch das Modul zwar in die dazugehörige Entfernung umgerechnet, jedoch nicht interpretiert oder weiterverarbeitet.

Neben den festen IR-Sensoren ist ein beweglicher IR-Sensor notwendig (Abbildung 13 rechte Teilgrafik). Dieser wird auf eine bewegliche Plattform (Sensorkopf) montiert. Der Sensorkopf ist das eigentliche Kernstück des Umgebungserfassungssystems. Er dient zur Erfassung der Umgebung des Roboters in einem bestimmten Radius um den Sensorkopf. Die Größe des Umkreises hängt von der Messreichweite des eingesetzten IR-Sensors ab.

# 3 Aufbau von Sensorkopf und Scannermechanik

Zur Montage und zum Betrieb des Sensorkopfes ist ein gewisses Maß an Mechanik notwendig. Dadurch kann zum einen die Schrittgenauigkeit des Schrittmotors erhöht werden, zum anderen vermindert man die auf den Antrieb wirkenden Kräfte bei sonst gleich bleibender Leistung. Die Folge ist ein geringerer Stromverbrauch und geringfügigere Erwärmung des Motors.

## 3.1 Schematischer Aufbau

Das Umgebungserfassungssystem besteht neben der Modulplatine aus dem Sensorkopf. Dieser ist aus dem Schrittmotor, einem Getriebe und einer drehbaren Sensorplattform aufgebaut.

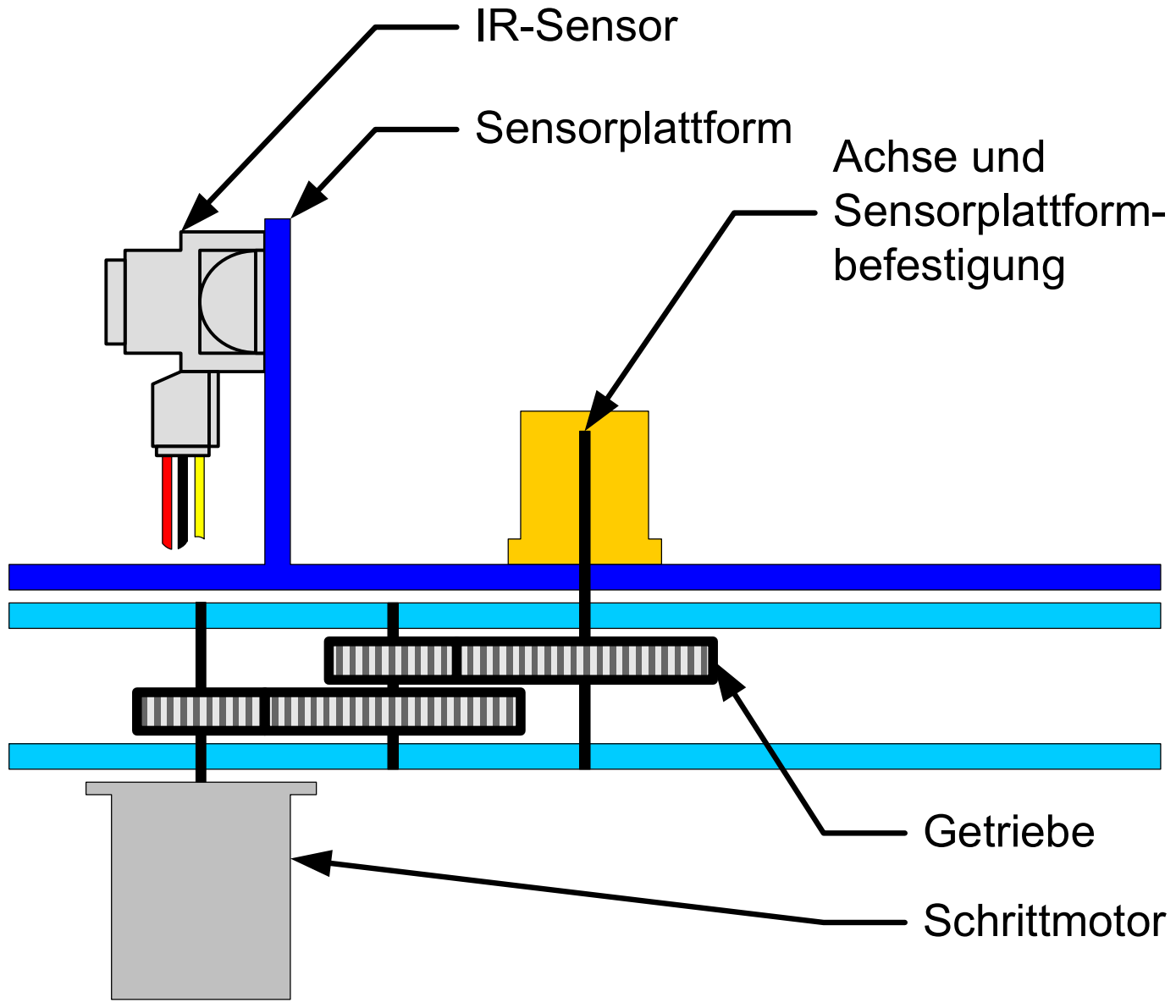

Abbildung 14 - Schematischer Aufbau des Sensorkopfes

Auf der Plattform ist der IR-Messsensor angebracht. Durch den Drehwinkel der Plattform und die Messdaten des IR-Sensors kann die Position der Objekte bestimmt werden. Das Verfahren wird in Kapitel 5.1 mit den entsprechenden mathematischen Grundlagen dargestellt.

## 3.2 Getriebe

Das Getriebe ist das Bindeglied zwischen dem Schrittmotor und dem Sensorkopf. Es dient der Auflösungserhöhung des Schrittmotors und zur Verringerung der mechanischen Kräfte auf diesen.

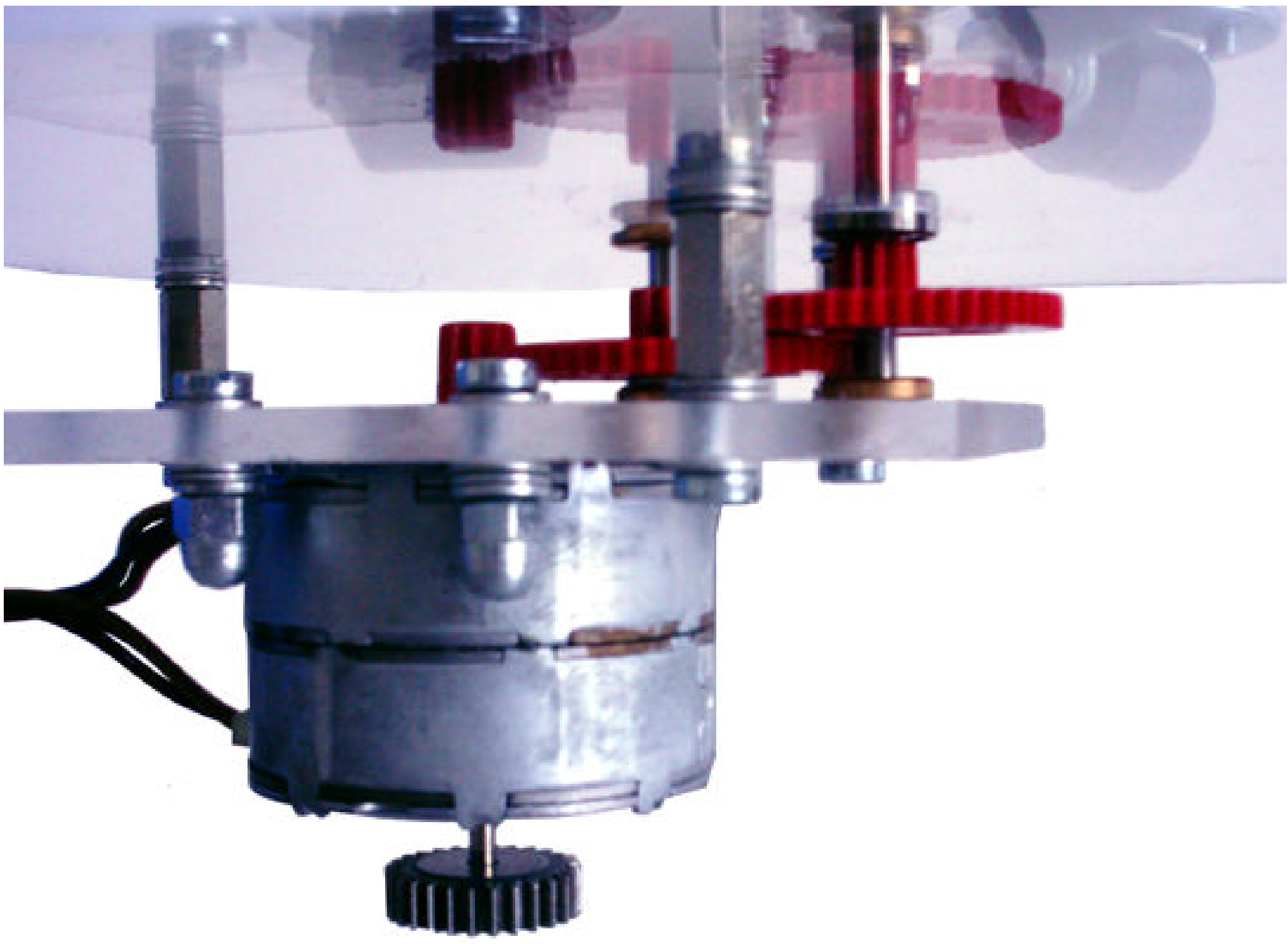

Abbildung 15 - Schrittmotor und Getriebe des Sensorkopfprototypen

### 3.2.1 Berechnung des Getriebeübersetzung

Um die real vom Sensorkopf auszuführenden Bewegungen in die nötigen Schrittmotorbewegungen zurückrechnen zu können, lassen sich folgende Formeln verwenden:

Drehzahl:

$$n_{aus} = \frac{z_{ein}}{z_{aus}} * n_{ein}$$

Die Drehzahl $n_{aus}$ berechnet sich aus den Verhältnis der Zähnezahl $z_{ein}$ des eingehenden und der Zähnezahl $z_{aus}$ des ausgehenden Zahnrades multipliziert mit der eingehenden Drehzahl $n_{ein}$.

Übersetzung:

$$i = \frac{\omega_{ein}}{\omega_{aus}} = \frac{z_{aus}}{z_{ein}} = \frac{n_{ein}}{n_{aus}}$$

Zur Berechnung der Unter- bzw. Übersetzung stehen drei Verhältnisse zur Verfügung.

Als Erstes das Teilungsverhältnis zwischen aus- und eingehender Zähnezahl ($z_{aus} / z_{ein}$), zum Zweiten dem zwischen ein- und ausgehender Drehzahl ($n_{ein} / n_{aus}$) und schließlich das zwischen ein- und ausgehendem Schrittwinkel ($\omega_{ein} / \omega_{aus}$).

### 3.2.2 Berechnung des Ausrichtwinkels des Sensorkopfes

Im Kapitel 3.2.1 wurden die mechanischen Größen für ein Zahnradpaar betrachtet. Zur Steuerung des Sensors muss das komplette Stirnradgetriebe berechnet werden. Der Schrittwinkel des Sensorkopfes $\alpha_{Sensor/Schritt}$ ergibt sich aus der Summe der Unter- bzw. Übersetzungen multipliziert mit dem Schrittwinkel des Schrittmotors $\alpha_{Schrittmotor/Schritt}$.

$$\alpha_{Sensor/Schritt} = \left( \prod_{k=1}^{m-1} \frac{n_k}{n_{k+1}} \right) * \alpha_{Schrittmotor/Schritt}$$

Dabei ist m die Anzahl der Achsen zwischen Motor und Sensorkopf inklusive des Motorritzels. Für $n_k$ wird die Zahnzahl des auf der Achse ausgehenden und für $n_{k+1}$ die Anzahl der Zähne des eingehenden Zahnrades der nächsten Achse eingesetzt.

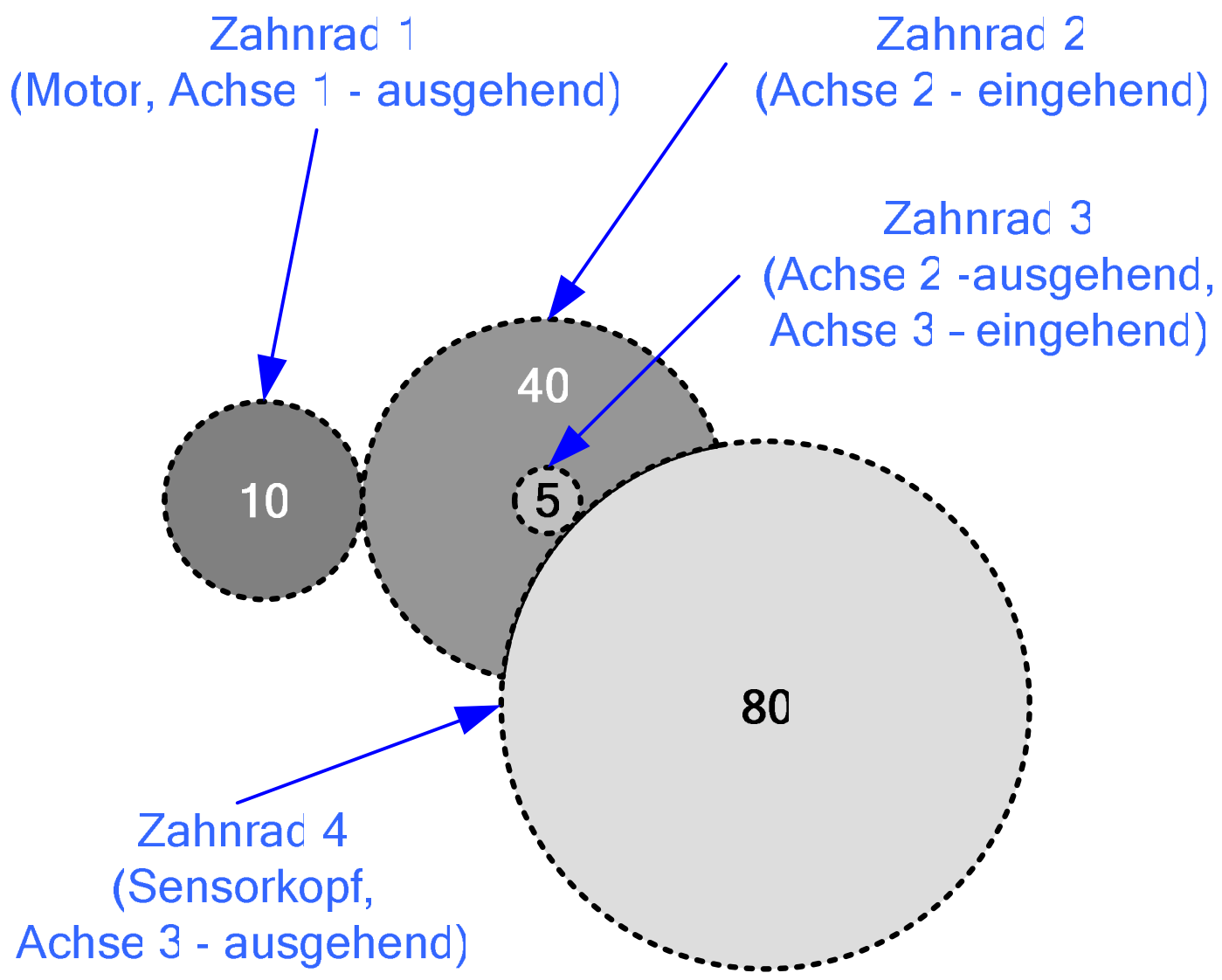

Abbildung 16 - Beispielgetriebe zur Berechnungsdemonstration

Für das Beispielgetriebe in Abbildung 16 und einem Schrittmotor mit einer Schrittauflösung von 3,6° ergibt sich:

$$\alpha_{Sensor/Schritt} = \frac{10}{40} * \frac{5}{80} * 3{,}6° = \underline{\underline{0{,}05625°}}$$

Dreht sich der Schrittmotor um 3,6°, bewegt sich der an Getriebeausgang angeschlossene Sensorkopf um 0,05625°. Das Getriebe hat somit eine Untersetzung von 1:64. Die notwendigen Angaben des eingesetzten Schrittmotors finden sich auf dem Motortypenschild oder können, wie im Kapitel 3.3.3 beschrieben, berechnet werden.

Um den Sensorkopfes exakt auszurichten, muss die Anzahl der Schritte des Motors für den gewünschten Drehwinkel errechnet werden.

Dabei gilt: $\alpha_{Sensor} = \alpha_{Sensor / Schritt} * n_{Schritte}$

Kombiniert man beide Formeln und formt sie nach der Schrittanzahl $n_{Schritte}$ um, ergibt sich:

$$n_{Schritte} = \frac{\alpha_{Sensor}}{\left( \prod_{k=1}^{m-1} \frac{n_k}{n_{k+1}} \right) * \alpha_{Schrittmotor / Schritt}}$$

Der Ausrichtwinkel des Sensorkopfes kann somit als Funktion der Schrittanzahl ausgedrückt werden. Die Schrittanzahl ($n_{Schritte}$) ist Element der ganzen Zahlen, wodurch eine gleitkommalose Implementierung im Microcontroller möglich wird. Damit wird Rechenzeit (des Microcontrollers) gespart, die mit Subprozessen ausgefüllt werden kann (z.B. Befehlsauswertung).

## 3.3 Schrittmotorsteuerung

*„Die Schrittmotoren werden als hochpräzise Stellelemente in der Elektrotechnik und der Feinmechanik benötigt. Man findet sie [ …] überall dort, wo man ganz exakte Positionierungen ausführen muss […]“* [Back 2003]. Beispielsweise sind sie in PC-Komponenten, wie Diskettenlaufwerken, Scannern oder Plotter zu finden. Auch in industrieller Umgebung kommen sie zum Einsatz. So sind CNC[1]-Maschinen oder Fertigungsroboter ohne sie undenkbar.

Genau durch diese Vorzüge eignen sich Schrittmotoren hervorragend zum Einbau in einem Umgebungsscanner. Die exakte Einstellung des Drehwinkels und somit die genaue Positionierbarkeit des Sensors sind für die Berechnung der Umgebung[2] unerlässlich

.

### 3.3.1 Grundlegender Aufbau und Funktion

Der im Umgebungsscanner verwendete Hybrid[3]-Schrittmotor ist eine Kombination aus Reluktanz- und Permanentmagnet-Schrittmotor. Er ist der heute am meisten eingesetzte Schrittmotor, da er hohe mechanische Leistungen bei kleinen Schrittwinkeln und kleiner Bauform vereint

Bei der Ansteuerung von Schrittmotoren wird zwischen uni- und bipolaren Typen unterschieden. In diesem Kapitel wird nur die Steuerung von bipolaren Schrittmotoren beschrieben, da unipolare Typen kaum noch zum Einsatz kommen.

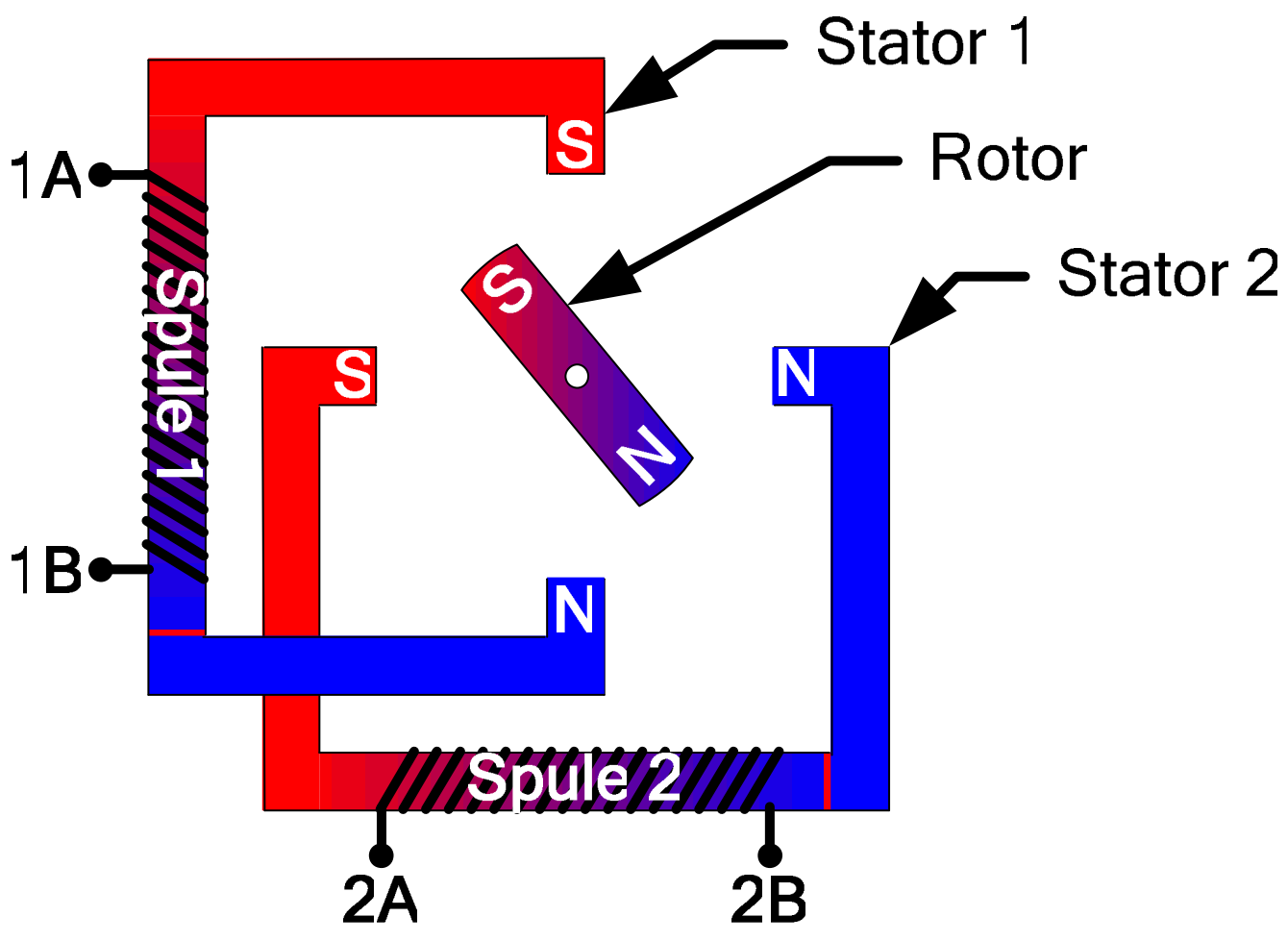

Abbildung 17 - Schematischer Aufbau eines bipolaren Schrittmotors

Der Grundaufbau eines bipolaren 4-Strang-Schrittmotors ist schematisch in Abbildung 17 dargestellt.

[1] Abkürzung für "Computerized Numerically Control" eine CNC-Maschine ist eine Werkzeugmaschine, die durch einen Computer gesteuert wird.

[2] Siehe Kapitel: „
Lokale Karten“

[3]Element, welches zwei unterschiedliche Funktionen oder Eigenschaften vereint.

Er besteht aus zwei Statorspulen und einem Rotor. Da jede der beiden Statorspulen je zwei Anschlüsse besitzt, sind vier Stränge (Kabel) aus dem Motor geführt. Daher die Bezeichnung bipolarer 4-Strang-Schrittmotor.

Abbildung 18 und 19 zeigt den in Abbildung 17 schematisch beschriebenen Aufbau, wie er in dieser Praxis realisiert wird.

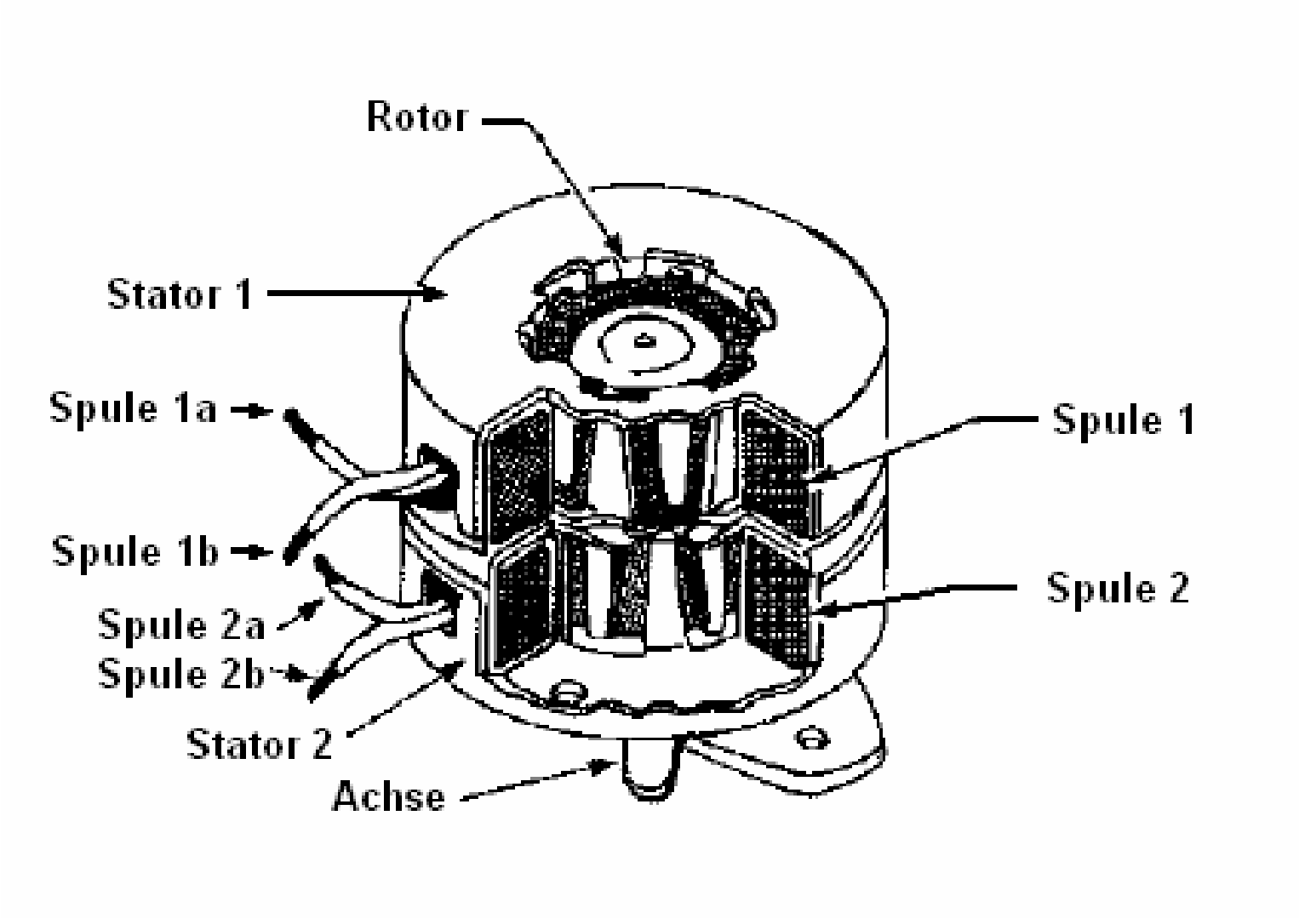

Abbildung 18 - Aufbau eines bipolaren Schrittmotors

Unipolare Schrittmotoren besitzen, im Gegensatz zu den bipolaren Schrittmotoren, eine Mittelanzapfung der Spulen. Diese Ausführung soll hier nicht betrachtet werden.

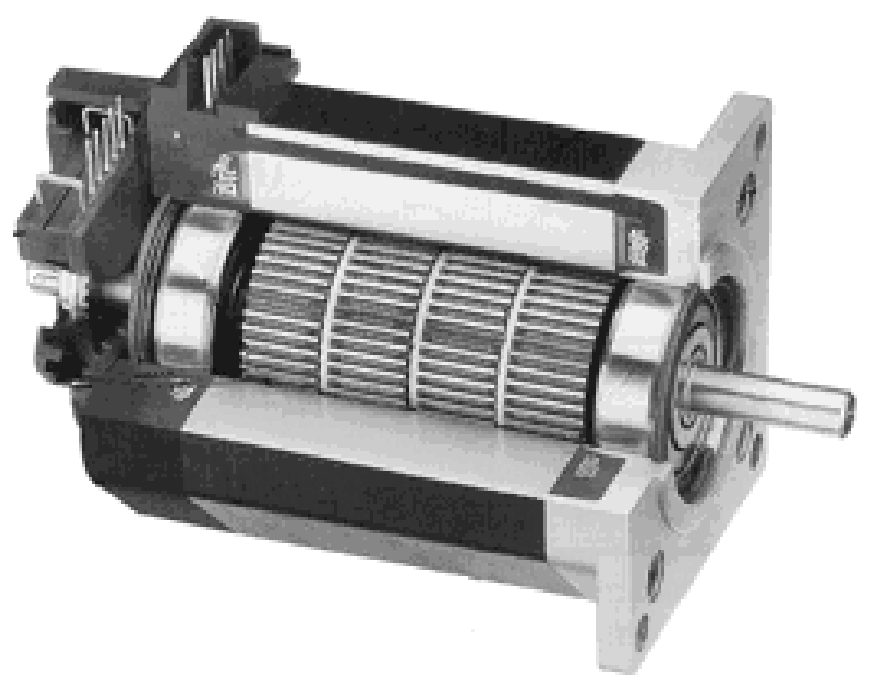

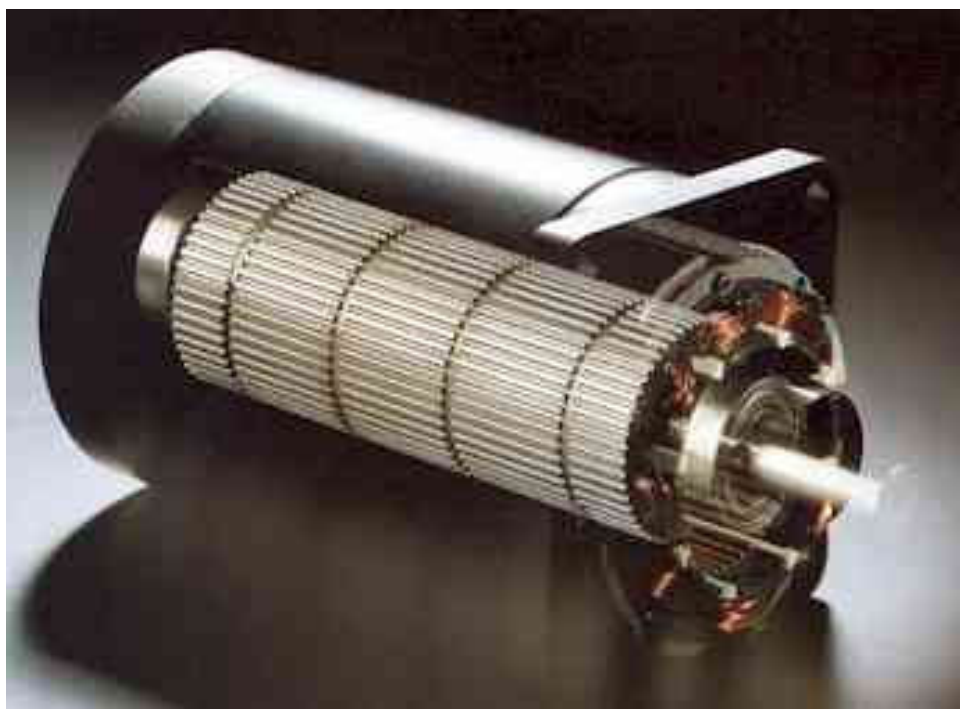

Abbildung 19 - Bipolarer 4-Strang-Schrittmotor

Die Motorspulen lassen sich je nach angelegter positiver oder negativer Spannung als Nord- oder Südpol polarisieren. Abhängig von der angelegten Spannung an den Statorspulen ändert sich die Stärke und Polung des elektromagnetischen Feldes, welches auf den Rotor wirkt.

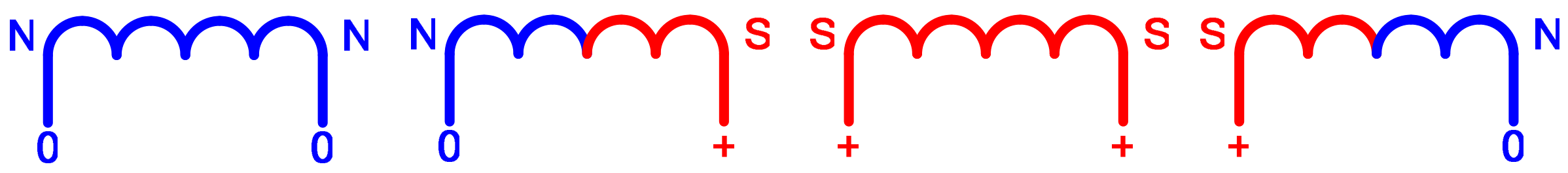

Abbildung 20 - Mögliche Polarisierung der Motorspulen eines bipolaren Schrittmotors

In welcher Reihenfolge die Ansteuerung der Spulen erfolgen muss wird in folgender Tabelle dargestellt:

| | Schritt 1 | Schritt 2 | Schritt 3 | Schritt 4 | |
|---|---|---|---|---|---|

Polung der Schrittmotorspulen

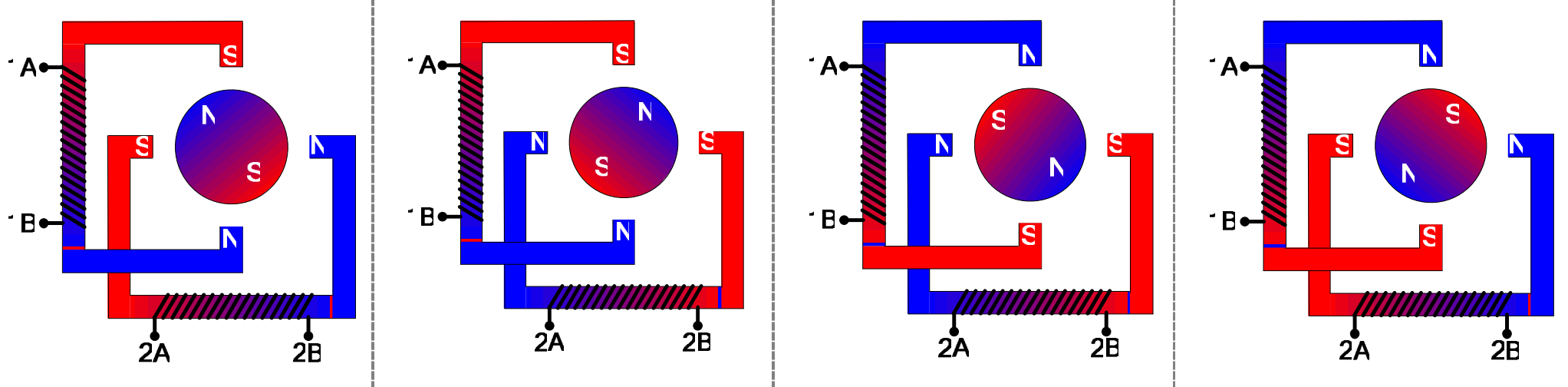

Logisches Ansteuerungsmuster

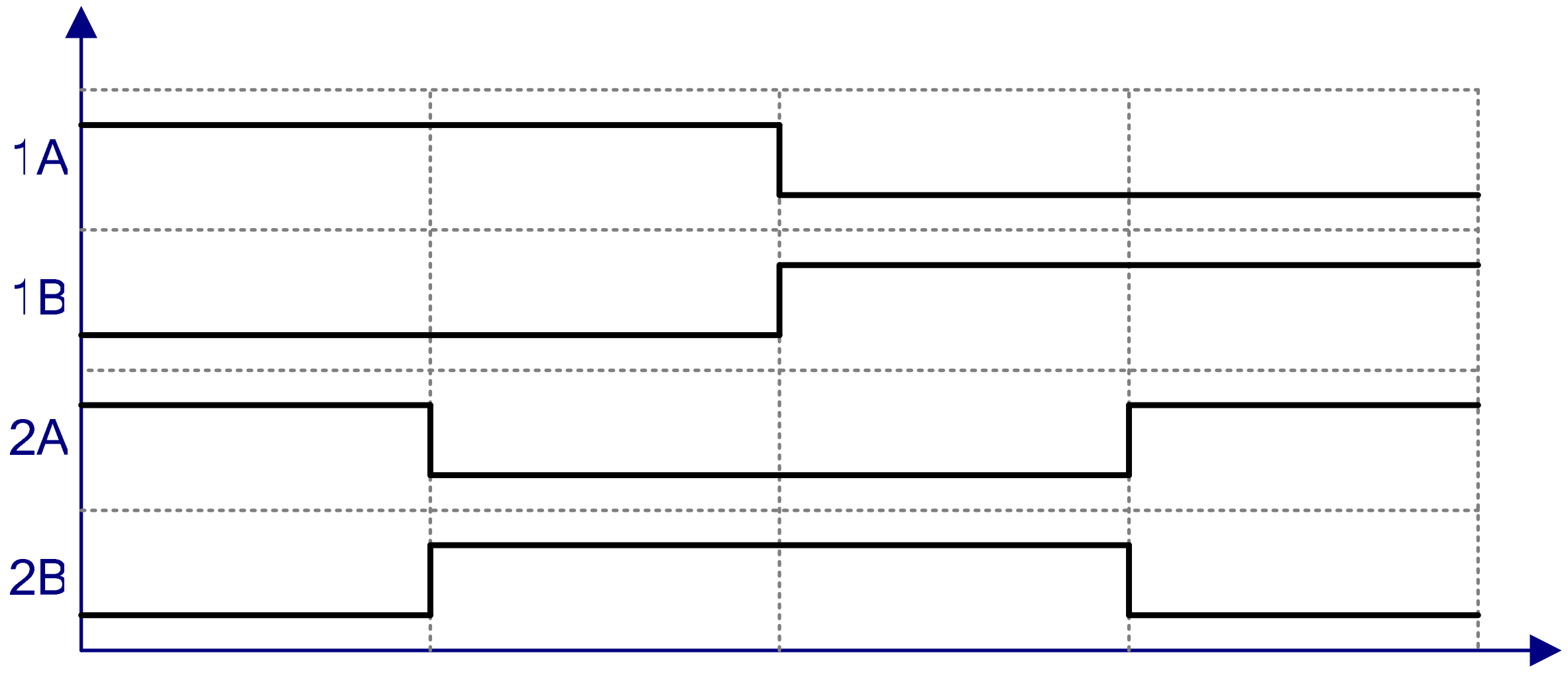

Verlauf des Spulenstroms

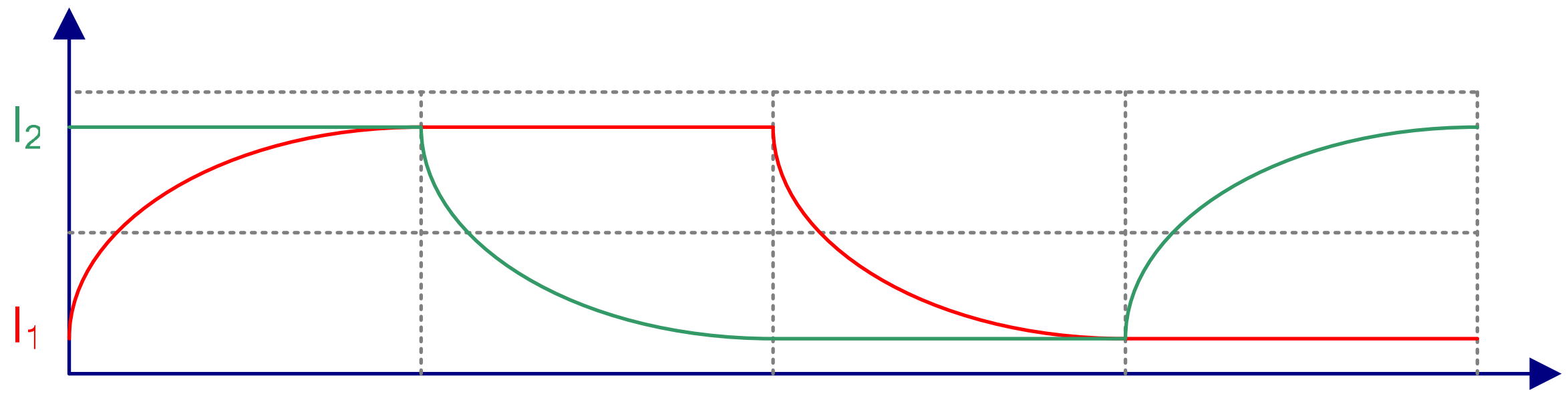

Tabelle 3 - Bipolarer Schrittmotor im Vollschrittbetrieb

Die Ansteuersequenz für die Phasen ist unabhängig vom verwendeten Typ (Reluktanz-, Permanentmagnet- oder Hybrid-Schrittmotor) immer identisch. Der Softwareanteil der Schrittmotorsteuerung wird im Kapitel 6.6.4 und die Schaltungsumsetzung im Abschnitt 6.5.3 detailliert beschrieben.

Im Folgenden ist die motorinterne Umsetzung der Sequenz für einen Hybrid-Typen schematisch dargestellt.

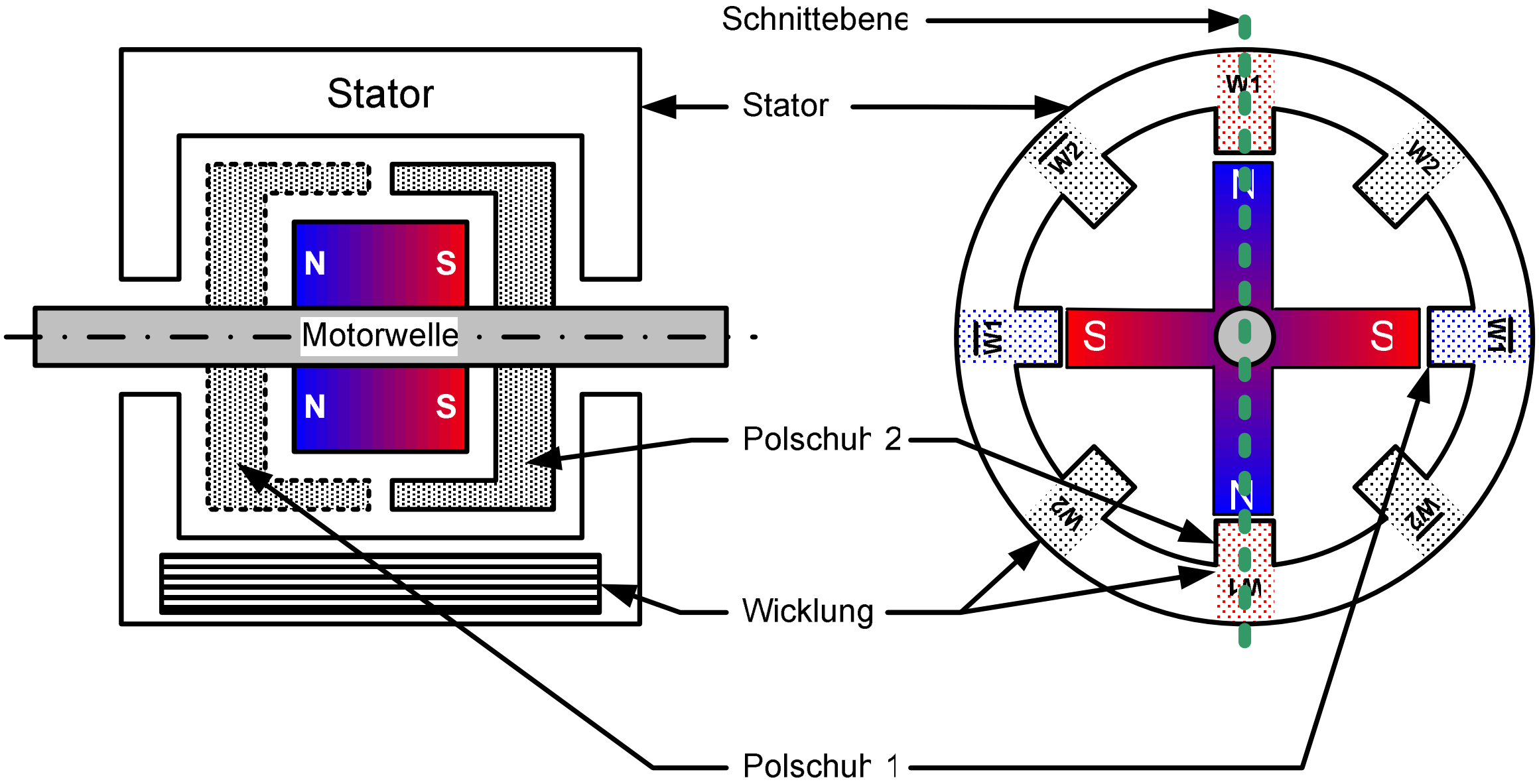

Abbildung 21 - Möglicher Aufbau eines Hybrid-Schrittmotors (schematisch vereinfacht)

Der Rotor bestehend aus zwei weichmagnetischen, gezahnten Polschuhen mit dazwischenliegenden Dauermagneten. Er wird in einer Sandwich-Bauweise realisiert. Die Polschuhe sind gegeneinander um eine ½ Zahnbreite versetzt angeordnet.

Wie der Rotor ist der Stator ebenfalls gezahnt. Um diesen liegen die Motorwicklungen. Diese sind die Ansteuerspulen des Schrittmotors (in Abbildung 21 und 22 W1, W2).

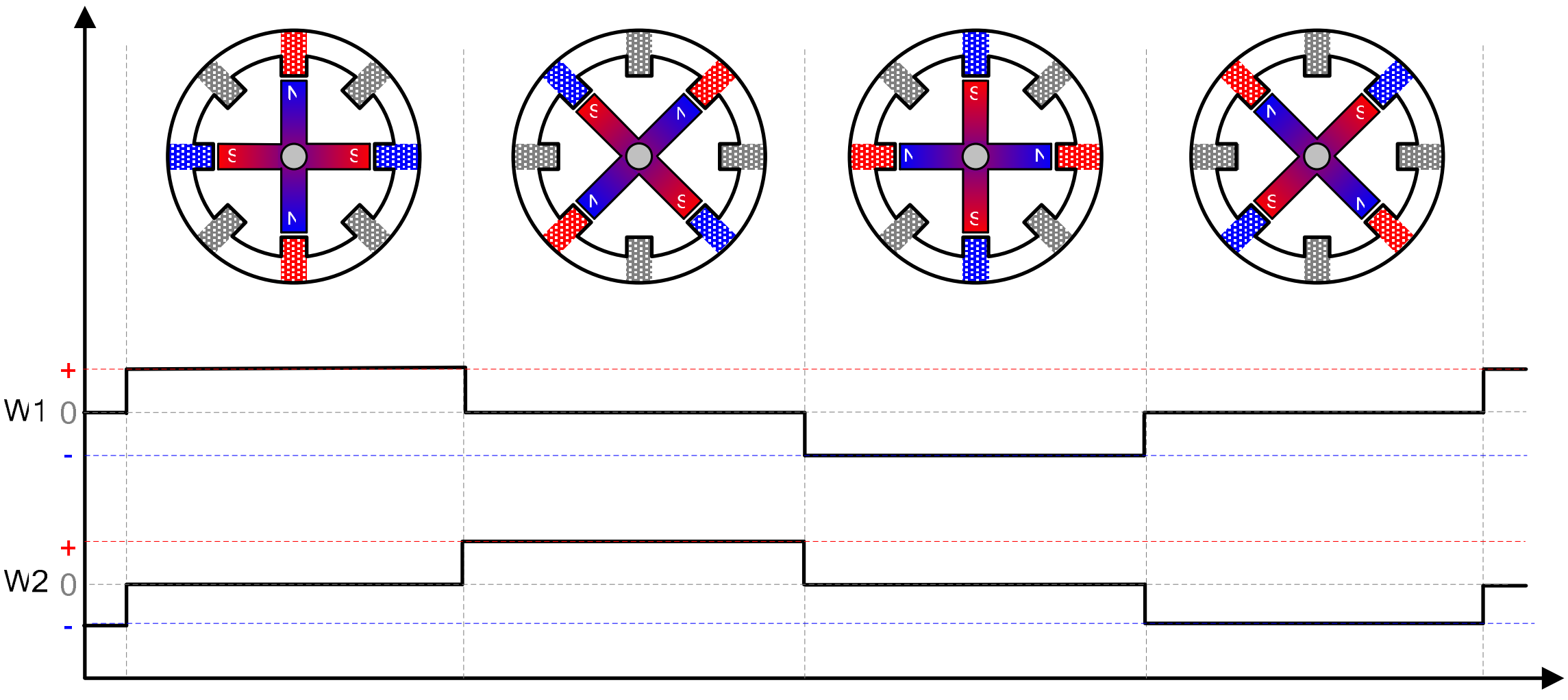

Abbildung 22 - Steuerung und Schrittfolge des Hybrid-Schrittmotors

Der Rotor nimmt aufgrund der Dauermagneten im unbestromten Zustand eine Vorzugslage ein (siehe linkes Bild in Abbildung 22). Damit bildet sich ein Rastmoment aus.

In der Grundstellung (siehe linkes Bild in Abbildung 22) wird die Wicklung W1 (obere, untere, linke und rechte Wicklung) bestromt. Die gegenüberliegenden Spulenpaare erzeugen, im Gegensatz zum Permanentmagnet-Schrittmotor, dieselbe magnetische Polarität. Der Rotor richtet sich mit seinem dauermagnetischen Feld im elektromagnetischen Feld aus. Damit erzeugt der Schrittmotor ein Haltemoment.

Kommutiert der Phasenstrom auf die Wicklung W2 (siehe zweites Bild in Abbildung 22) wandern die Nord- und Südpole, im Beispiel, jeweils um 45° im Urzeigersinn. Der Rotor folgt den wandernden Magnetfeldern und richtet sich neu aus.

Im Folgenden (siehe drittes Bild in Abbildung 22) wird wiederum die Spule W1 bestromt. Allerdings mit der im ersten Schritt entgegengesetzter Polarität.

Als Viertes wird die Spule W2 (siehe rechtes Bild in Abbildung 22) bestromt, jedoch entgegengesetzt zum Schritt Zwei.

Der folgende Ansteuerschritt entspricht wiederum der Grundstellung.

Im Gegensatz zum erläuterten Beispiel bieten die eingesetzten Hybrid-Schrittmotoren 100 bis 2000 Schritte ohne elektronische Zusatzmaßnahmen. Entsprechend der Schrittzahl muss die beispielhaft beschriebene Steuersequenz wiederholt ausgeführt werden. Für eine volle Umdrehung eines Hybrid-Schrittmotors mit einer Auflösung von 100 Schritten muss die Sequenz, welche vier Schritte repräsentiert, beispielsweise 25 Mal ausgeführt werden.

### 3.3.2 Kenngrößen des Schrittmotors und ihre Bedeutung

Schrittmotoren unabhängig vom Typ oder der Ausführung können über folgende Kenngrößen charakterisiert werden:

a. <u>Drehmoment:</u>

   Das Moment des Motors, welches bei unterschiedlichen Drehzahlen erzeugt wird, bezeichnet man als Drehmoment.

   Für die Bewegung der mechanischen Komponenten ist mit zunehmender Masse ein höheres Drehmoment notwendig, um diese bewegen zu können.

b. <u>Haltemoment:</u>

   Ein Moment, welches den Schrittmotor im Stillstand halten kann, ohne dass es eine kontinuierliche Drehung des Rotors hervorruft, wird Haltemoment genannt.

   Je stärker die Kräfterückkopplung der mechanischen Komponenten auf den Motor ist, desto größer muss das Haltemoment sein, damit der Rotor nicht aus seiner Lage gezwungen wird.

c. Motortemperatur
   Die Motortemperatur bezieht sich auf einen bestimmten Messpunkt des Motorgehäuses. Sie darf während des Betriebes nicht überschritten werden, da ansonsten irreversible Schäden entstehen können.

d. Nennspannung
   Mit der Nennspannung ist festgelegt, welche Spannung im stationären Fall an den Motor angelegt werden muss, um den Phasenstromnennwert zu erreichen. Bei einer Konstantstromansteuerung des Motors darf die Nennspannung nicht mit der Betriebsspannung verwechselt werden.

e. Phasenstrom
   Der Phasenstrom bezieht sich auf das Nennmoment des Motors. Mit diesem Strom kann der Schrittmotor im Dauerbetrieb genutzt werden, ohne ihn thermisch zu überlasten. Er kann dem entsprechenden Datenblatt entnommen werden.

   Ist die anzutreibende Masse bereits in Bewegung kann der Phasenstrom elektronisch abgeregelt werden, da nur das bewegungserhaltende Drehmoment aufgebracht werden muss.

   In der Beschleunigungs-, Brems- oder Positionshaltephase der Masse kann der Phasenstrom zur Erhöhung des Drehmoments gesteigert werden.

   Ein höherer Phasenstrom für ein größeres Drehmoment hat eine stärkere Erwärmung des Schrittmotors zur Folge.

f. Phasenzahl
   Die Phasenzahl entspricht den Anschlüssen der Spulen. Bei einem 4-Strang-Biplolar-Schrittmotor beispielsweise gibt es vier Phasen.

g. Rotorträgheitsmoment
   Zum Trägheitsmoment der Last addiert sich das Trägheitsmoment des Schrittmotorrotors. Es begrenzt somit die maximal mögliche Beschleunigung. Dieser physikalischen Größe muss bei der Ansteuerung des Schrittmotors Rechnung getragen werden.

h. Schrittwinkel
   Der Schrittwinkel gibt an, welchen Drehwinkel der Motor ohne elektronische Zusatzmaßnahmen auflösen kann.

   Ein kleiner Schrittwinkel hat eine höhere Schrittauflösung, also eine genauere Positionierung zur Folge.

i. Wicklungsinduktivität
Sie ist bei der Wahl der Betriebsspannung von Bedeutung. Die Winklungsinduktivität bestimmt die Geschwindigkeit des Strom Auf- und Abbaus und beeinflusst deshalb die dynamischen Eigenschaften des Schrittmotors.

j. Wicklungswiderstand
Der Wicklungswiderstand ist der ohmsche Widerstand des Motors. Er kann gegebenenfalls Auswirkungen auf die Auslegung der Ansteuerelektronik haben.
Je kleiner der Wicklungswiderstand ist, desto geringer ist die Erwärmung des Motors.

Die Kenngrößen sind für die Steuerungshardware und die Dimensionierung der mechanischen Komponenten, wie dem Getriebe, von Wichtigkeit.

Dreh- und Trägheitsmomente der an den Schrittmotor angeschlossenen Mechanik stellen den Haupteinfluss auf den Motor dar. Das Antriebsmoment des Schrittmotors muss größer sein, als die Masseträgheitsmomente der mechanischen Komponenten.

Eine Erwärmung des Schrittmotors während des Betriebs ist unvermeidlich. Je mehr Drehmoment zum Antrieb der Mechanik benötigt wird (in Relation zum Motordrehmoment), desto stärker erwärmt sich der Motor. Wärme bedeutet jedoch eine Zunahme des Spulenwiderstandes. Die Folge ist ein höherer Energieverbrauch durch steigende Phasenströme.

Erwärmung bedeutet Energieverlust, welche nicht mehr in Antriebskraft umgesetzt werden kann. Wird jedoch viel Drehmoment benötigt, z.B. beim Beschleunigen der Sensorplattform, muss der Phasenstrom erhöht werden. Hoher Phasenstrom ist jedoch gleichbedeutend mit starker Erwärmung des Schrittmotors. Das Dreh- und Haltemoment des Schrittmotors beispielsweise kann durch ein Untersetzungsgetriebe (siehe Kapitel 3.2) erhöht werden. Ebenso kann der Phasenstrom bei gleich bleibender Leistung durch die Nutzung eines Getriebes verringert werden.

Spannung und Phasenstrom sind deshalb bei der Dimensionierung der Leistungselektronik (siehe Abschnitt 6.5.3) der Schrittmotorsteuerung einzubeziehen. Diese Parameter sollten durch entsprechende Schaltungen überwacht und gesteuert werden.

Zwischen den Kenngrößen bestehen also wichtige Zusammenhänge welche zu beachten sind.

### 3.3.3 Berechnungsgrößen

Der Schrittwinkel $\alpha$ berechnet mit Hilfe der Polzahl p und der Phasenzahl (Stranganzahl).

Vollschrittbetrieb: $\alpha = \frac{360°}{2p * m}$

Halbschrittbetrieb: $\alpha = \frac{360°}{4p * m}$

Aus diesen beiden Größen lässt sich ebenfalls die Schrittzahl S bestimmen, welche für eine vollständige (360°) Drehung des Rotors nötig ist.

Vollschrittbetrieb: $S = 2p * m$

Halbschrittbetrieb: $S = 4p * m$

Die Drehzahl n errechnet sich aus der Schrittfrequenz $f_s$ dividiert durch die Schrittzahl S ($S = 2p * m$).

Vollschrittbetrieb: $n = \frac{f_s}{2p * m}$

Halbschrittbetrieb: $n = \frac{f_s}{4p * m}$

Die Berechnungsgrößen des Schrittmotors entsprechen den Eingangsgrößen für das im Kapitel 3.2 beschriebene Getriebe. Sie werden im Allgemeinen auf dem Typenschild des Motors oder im Datenblatt angegeben und müssen nur berechnet werden, falls dies nicht der Fall ist.

# 4 Trainingsprogramm des künstlichen neuronalen Netzes

„Den Zusammenschluss von intelligenten Paradigmen (Computing Techniques) wie Fuzzy-Logik (FL), Neuronale Netzwerke (NN), wahrscheinlichkeitsbedingte Schlussfolgerungen, Chaostheorie, genetische Algorithmen und Teile der Lerntheorie, die mit den überall vorherrschenden Unbestimmtheiten und den Undefiniertheiten der realen Welt arbeiten, nennt man Soft-Computing." [Aliev 2000]

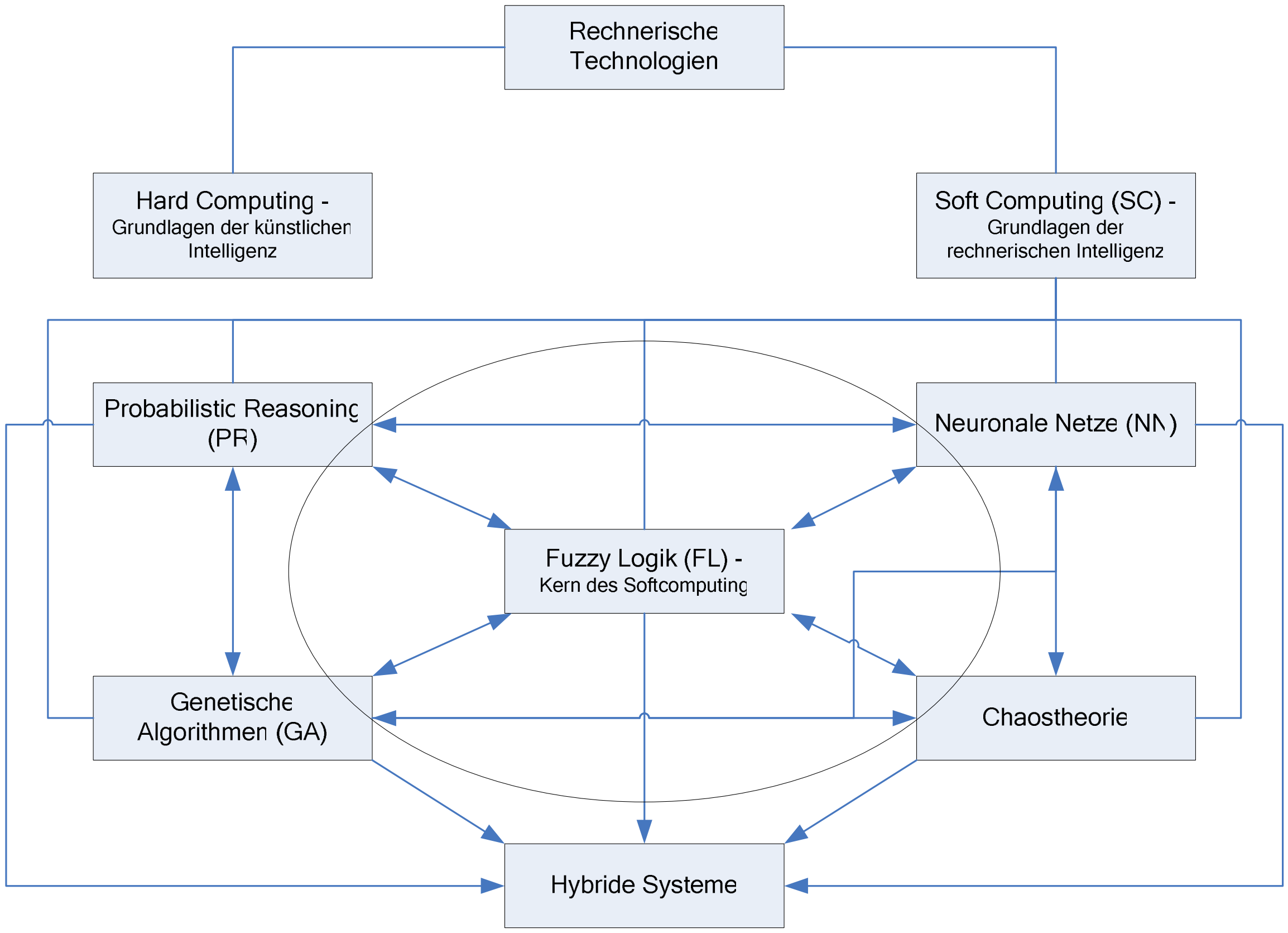

Abbildung 23 - Struktur der Rechenbasis für die künstliche Intelligenz [nach Aliev 2000]

Aus dem umfangreichen Gebiet des Soft-Computings wird im Umgebungserfassungssystem ein künstliches neuronales Netz (KNN) verwendet. In den folgenden Kapiteln wird die Art und die Auswahlkriterien für das KNN beschrieben.

## 4.1 Neuronale Netze im Überblick

Es existieren verschiedene Netzmodelle, welche durch ihre Eigenschaften für bestimmte Aufgaben besonders gut geeignet sind.

| KNN-Modell | Typische Anwendung | Stärken | Schwächen |
|---|---|---|---|
| Hopfield/Kohonen | Rekonstruktion von Daten/Bildern aus Fragmenten | großformatige Implementierung möglich | lernen ist nicht möglich, Gewichte müssen gesetzt werden |
| Perzeptron | Erkennung von Schreibmaschinen-buchstaben | ältestes künstliches neuronales Netz | keine Erkennung komplexer Strukturen, empfindlich gegen Veränderungen |
| Mehrschicht-Perzeptron/Delta-Regel | Mustererkennung | einfaches KNN, allgemeiner als Perceptron | keine Erkennung komplexer Strukturen |
| Back-Propagation | weiter Anwendungsbereich (von Sprachsynthese bis Kreditvergabe) | das am meisten verbreitete Netz funktioniert gut und lernt leicht | gesteuertes Lernen durch eine Vielzahl von Beispielen |
| Boltzmann-Maschine | Muster-Erkennung (Radar, Sonar) | einfaches Netz, das Rauschen verwendet, um ein globales Energieminimum zu erhalten | lange Trainingszeiten |
| Counter-Propagation | Bildkompression, statistische Analyse, Kreditvergabe | einfaches Mehrschicht-Perceptron, aber weniger leistungsfähig als Back-Propagation | große Anzahl von Prozessorelementen und Verbindungen nötig |
| Self-Organizing-Map | bildet eine geometrische Region auf einer anderen ab | bessere Leistung als viele algorithmische Techniken | intensives Training |
| Neocognitron | Erkennung handgeschriebener Buchstaben | raffiniertes Netz, das komplexe Muster erkennen kann | große Anzahl von Prozessorelementen und Verbindungen nötig |

Tabelle 4 - Die wichtigsten KNN-Modelle und ihre Eigenschaften

Alle Modelle haben Vor- und Nachteile, wodurch es kein ideales KNN gibt. Welche Art von KNN für die jeweilige Anwendung genutzt wird, hängt also von der Art der Anwendung ab.

Im Fall des Umgebungserfassungssystems eignet sich ein Mehrschicht-Perzeptron (Multilayer-Perceptron, in folgenden MLP genannt).

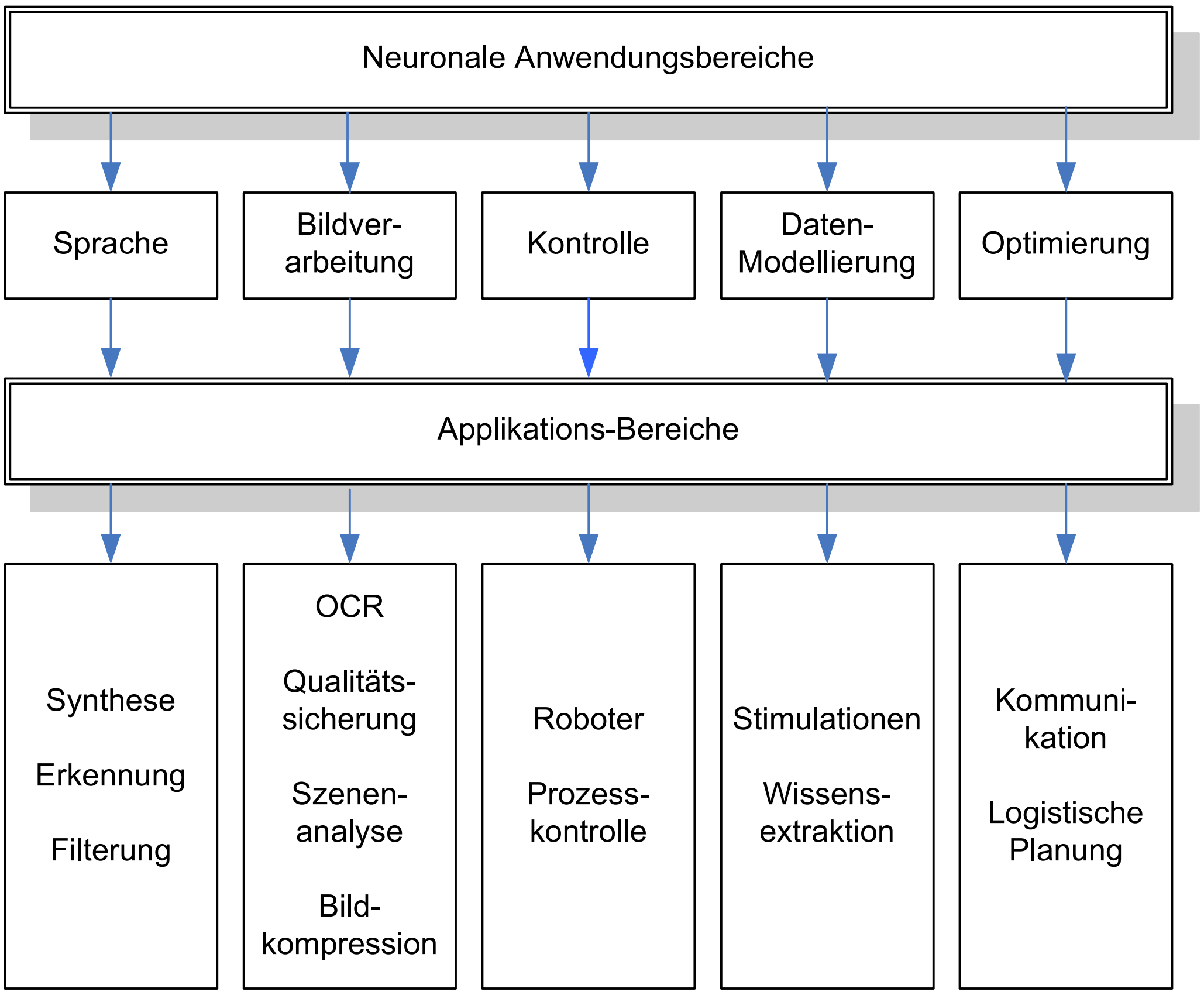

Abbildung 24 - Anwendungsbereiche künstlicher neuronaler Netze [nach Hamilton 1993]

Das MLP soll zur Konvertierung des Spannungswertes des Infrarot-Sensors in die äquivalente Entfernung genutzt werden. In den Microcontroller wird dabei das trainierte Netz integriert (eingefrorenes Wissen).

Das neuronale Netz wird extern auf einem Computer mit den optimierten Messdaten trainiert. In den Microcontroller muss somit nur die Berechnungsroutine für die Outputs implementiert werden.

## 4.2 Entscheidungskriterium für ein neuronales Netz

Wie schon im Kapitel 2.1.1 („Infrarotsensoren“) beschrieben wurde, weisen die eingesetzten Infrarot-Distanzmesssensoren von Sharp keine lineare Kennlinie auf.

Es gibt mehrere Möglichkeiten den Wert des Analog-Digitalwandlers in die Entfernung umzuformen.

Die einfachste Lösung ist die Erstellung einer mathematischen Transferfunktion. Dabei kann man unter Zuhilfenahme von ausgewählten Stützpunkten die Funktion durch Reihenfunktionen oder Gleichungssysteme berechnen. Eine einfachere Erstellungsmöglichkeit der Funktionsgleichung bietet Microsoft Excel unter dem Menüpunkt „Trendlinie hinzufügen“. Zur Gewinnung der Trainingsdaten wurde diese Technik eingesetzt. Eine genaue Beschreibung folgt im Kapitel 4.2.1.

Die interpolierte Transferfunktion sollte möglichst exakt an den gemessenen Werten liegen. Diese Funktion ist ausschlaggebend für die Genauigkeit der Entfernungsbestimmung. Sollte die mathematische von der realen Funktion zu stark abweichen, ist dies eine nicht zu vernachlässigende Fehlerquelle. Diese Fehler pflanzen sich nicht nur in der lokalen Karte (siehe Kapitel 5.1) fort, sondern addieren sich in der globalen Karte (siehe Kapitel 5.2) auf. Sie müssen über eine Ausgleichsfunktion eliminiert werden.

Die Transferfunktion muss zwecks Datenverarbeitung in den Microcontroller integriert werden. Dazu müssen neben der eigentlichen Berechnungsformel alle benötigten mathematischen Bibliotheken und Funktionen integriert werden.

Sollte man jedoch einen anderen oder mehrere Sensoren einsetzen, wird es notwendig mehrere Transferfunktionen einzusetzen. Der Nachteil ist, dass bei jeder Änderung oder Erweiterung das Programm geändert oder erweitert werden muss. Danach sind eine erneute Übersetzung des Programms (Kompilierung) und eine Neuprogrammierung des Microcontrollers unumgänglich. Diese Vorgehensweise ist sehr aufwendig, da das Modul im System oft nicht frei zugänglich ist.

Mathematische Formeln lassen sich ebenfalls nicht in externen Bausteinen (z.B. $I^2C$-EEPROM) speichern, da sie Registeroperationen oder Adresssprüngen im Microcontrollerprogramm entsprechen. Die Berechnungsalgorithmen müssen somit fest im Microcontrollerprogramm integriert werden, jedoch können die Formelparameter extern geladen werden.

Ein solcher Lösungsansatz wäre der Einsatz eines künstlichen neuronalen Netzes. Dabei wird, da sich die Funktionen nicht ändern, ein bereits trainiertes Netz eingesetzt. Es entspricht den Parametern eines Gleichungssystems, welche aus einem microcontrollerfremden Speicher (z.B. $I^2C$-EEPROM) geladen werden können. Somit muss das eigentliche Microcontrollerprogramm beim Einsatz neuer Sensoren nicht mehr verändert werden.

Die gemessenen Sensordaten werden dem KNN als Eingaben präsentiert. Nach der Ausführung des Forward-Passes, liefert das Netz als Ausgabe die zum Input passende Entfernung (siehe Abbildung 25).

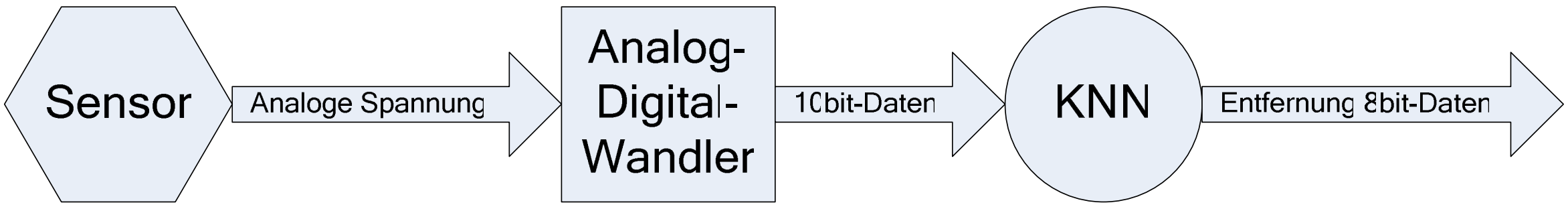

Abbildung 25 - Struktur der Sensordaten-Entfenungs-Wandlung

Jedes der zehn Bits des AD-Wandlers (in Abbildung 26 ADC[]) repräsentiert ein Eingabeneuron des Netzes. Der Entfernungswert wird von den acht Ausgabeneuronen (in Abbildung 26 E[]) binär ausgegeben. Die Einzelbits müssen danach aneinander gereiht werden, um den Sensor-Objekt-Abstand als Byte weiterverarbeiten zu können.

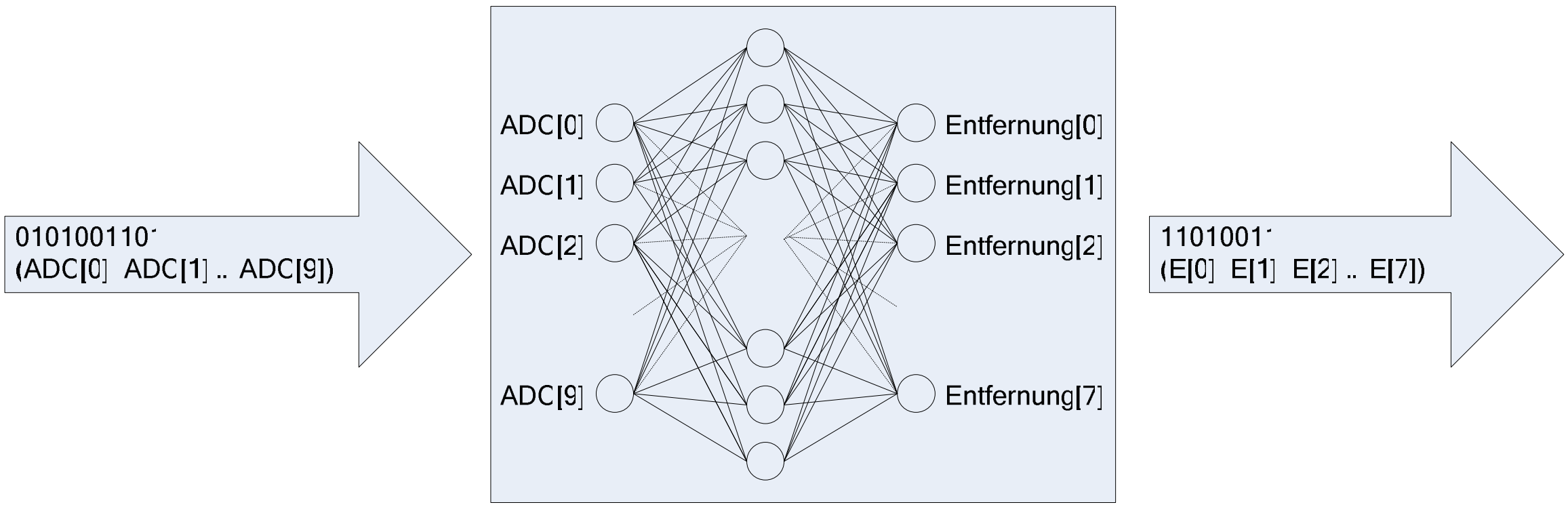

Abbildung 26 - Datenverarbeitung durch das künstliche neuronale Netz

### 4.2.1 Gewinnung der Trainingsdaten

Bevor das künstliche neuronale Netz trainiert werden kann, müssen die Trainingsdaten gewonnen und aufbereitet werden.

Es werden von den jeweiligen Sensoren Messreihen (in Abbildung 27 blau dargestellt) aufgenommen. Diese sind fehlerbehaftet. In Abbildung 27 ist eine solche Messreihe exemplarisch für den GP2D12 dargestellt.

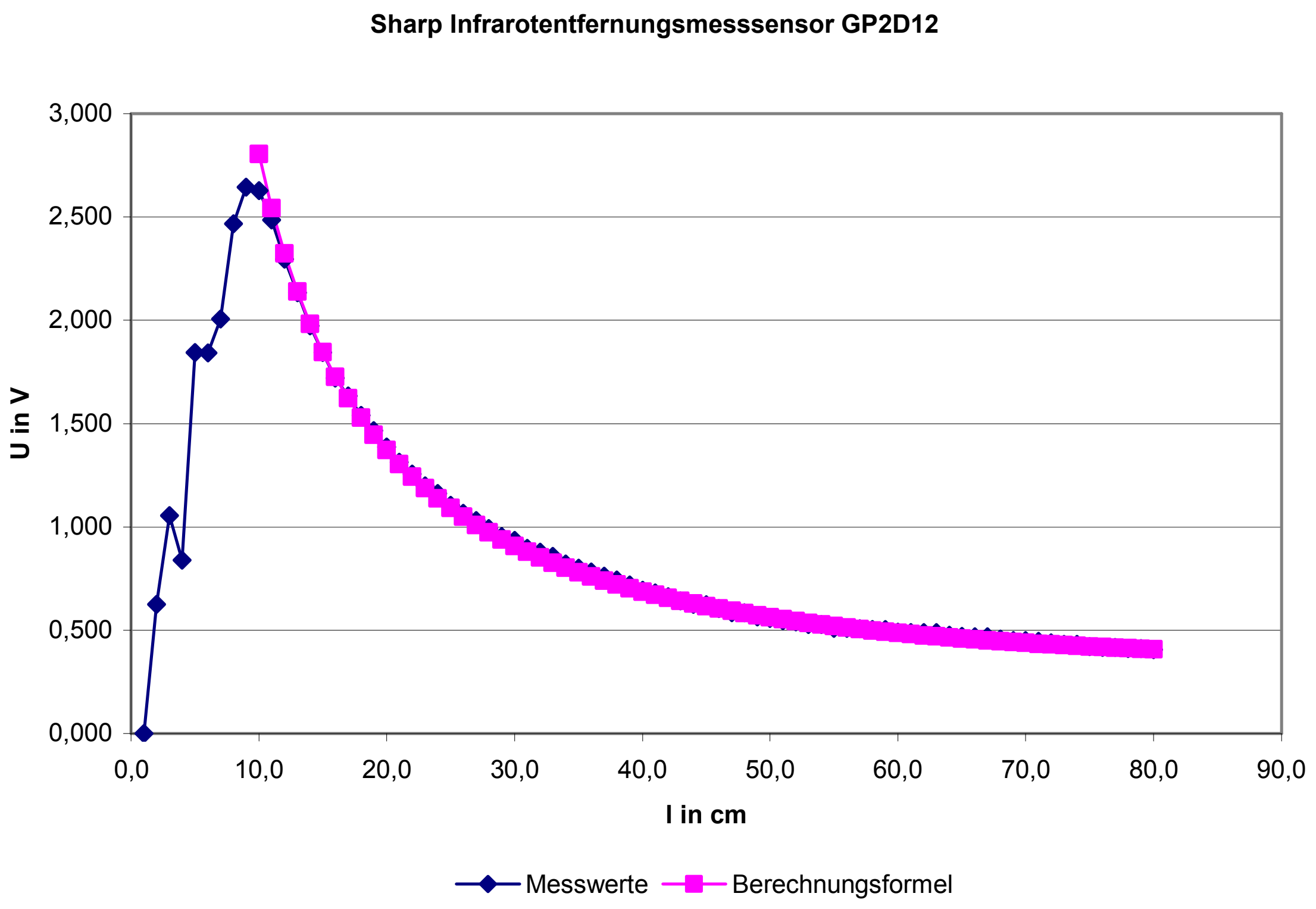

Abbildung 27 - Messwerte des GP2D12

Die Messreihen müssen im nächsten Schritt geglättet, also von Messfehlern befreit werden. Dazu wird eine Funktion (in Abbildung 27 rosa dargestellt) approximiert, die den realen Messwerten angepasst wird. Die Erstellung der Formel erfolgt am einfachsten über eine Reihenentwicklung. Dafür werden beliebige Wertepaare als Stützstellen gewählt. Es sollten mehrere Reihen mit verschiedenen Stützpunkten entwickelt werden, da die Wertepaare fehlerbehaftet sein können. Es entstehen Funktionen, welche sich ausschließlich durch ihre Koeffizienten unterscheiden. Diese liegen in einem Streuungsbereich. Es wird für jeden Koeffizienten der Durchschnittskoeffizient gebildet. Mit diesen wird die optimierte Funktion erstellt, welche fehlerminimiert ist.

Mit Hilfe dieser Berechnungsformel werden die eigentlichen Trainingsdaten erstellt. Diese sind von den Messfehlern bereinigt. Das künstliche neuronale Netz wird mit den optimierten Daten trainiert.

Nach der Analog-Digital-Wandlung liegen die Sensordaten im 10bit-Format vor. Damit sind $2^{10} = 1024$ Eingangsmuster möglich. Für die Entfernung wird eine Auflösung von minimal 1cm gewählt. Damit werden beispielsweise maximal 200 verschiedene Ausgabemuster für den GP2D120 (max. 200cm, normal 150cm Messweite) benötigt. Es werden acht Ausgabeneuronen ($2^8 = 256 > 200$) eingesetzt.

Im Durchschnitt fallen vier Eingabemuster zu einer Ausgabeklasse zusammen. In Tabelle 5 ist eine exemplarische Musterklasse rot dargestellt.

<table>
<tr><th colspan="10">Eingabedaten</th><th colspan="8">Ausgabedaten</th></tr>
<tr><th>$I_0$</th><th>$I_1$</th><th>$I_2$</th><th>$I_3$</th><th>$I_4$</th><th>$I_5$</th><th>$I_6$</th><th>$I_7$</th><th>$I_8$</th><th>$I_9$</th><th>$O_0$</th><th>$O_1$</th><th>$O_2$</th><th>$O_3$</th><th>$O_4$</th><th>$O_5$</th><th>$O_6$</th><th>$O_7$</th></tr>
<tr><td>0</td><td>1</td><td>0</td><td>1</td><td>0</td><td>0</td><td>0</td><td>0</td><td>1</td><td>1</td><td>0</td><td>1</td><td>0</td><td>0</td><td>0</td><td>0</td><td>0</td><td>0</td></tr>
<tr><td>0</td><td>1</td><td>0</td><td>1</td><td>0</td><td>0</td><td>0</td><td>1</td><td>0</td><td>0</td><td rowspan="4">0</td><td rowspan="4">1</td><td rowspan="4">0</td><td rowspan="4">0</td><td rowspan="4">0</td><td rowspan="4">0</td><td rowspan="4">0</td><td rowspan="4">1</td></tr>
<tr><td>0</td><td>1</td><td>0</td><td>1</td><td>0</td><td>0</td><td>0</td><td>1</td><td>0</td><td>1</td></tr>
<tr><td>0</td><td>1</td><td>0</td><td>1</td><td>0</td><td>0</td><td>0</td><td>1</td><td>1</td><td>0</td></tr>
<tr><td>0</td><td>1</td><td>0</td><td>1</td><td>0</td><td>0</td><td>0</td><td>1</td><td>1</td><td>1</td></tr>
<tr><td>0</td><td>1</td><td>0</td><td>1</td><td>0</td><td>0</td><td>1</td><td>0</td><td>0</td><td>0</td><td>0</td><td>1</td><td>0</td><td>0</td><td>0</td><td>0</td><td>1</td><td>0</td></tr>
</table>

Tabelle 5 - Aufbereitete Trainingsdaten

Durch den unterschiedlichen Anstieg der Transferfunktion fallen in der Realität mehr Muster in Bereichen mit starkem Anstieg (Nahbereich) in eine Klasse, als in Abschnitten mit einem geringen Anstieg (Fernbereich).

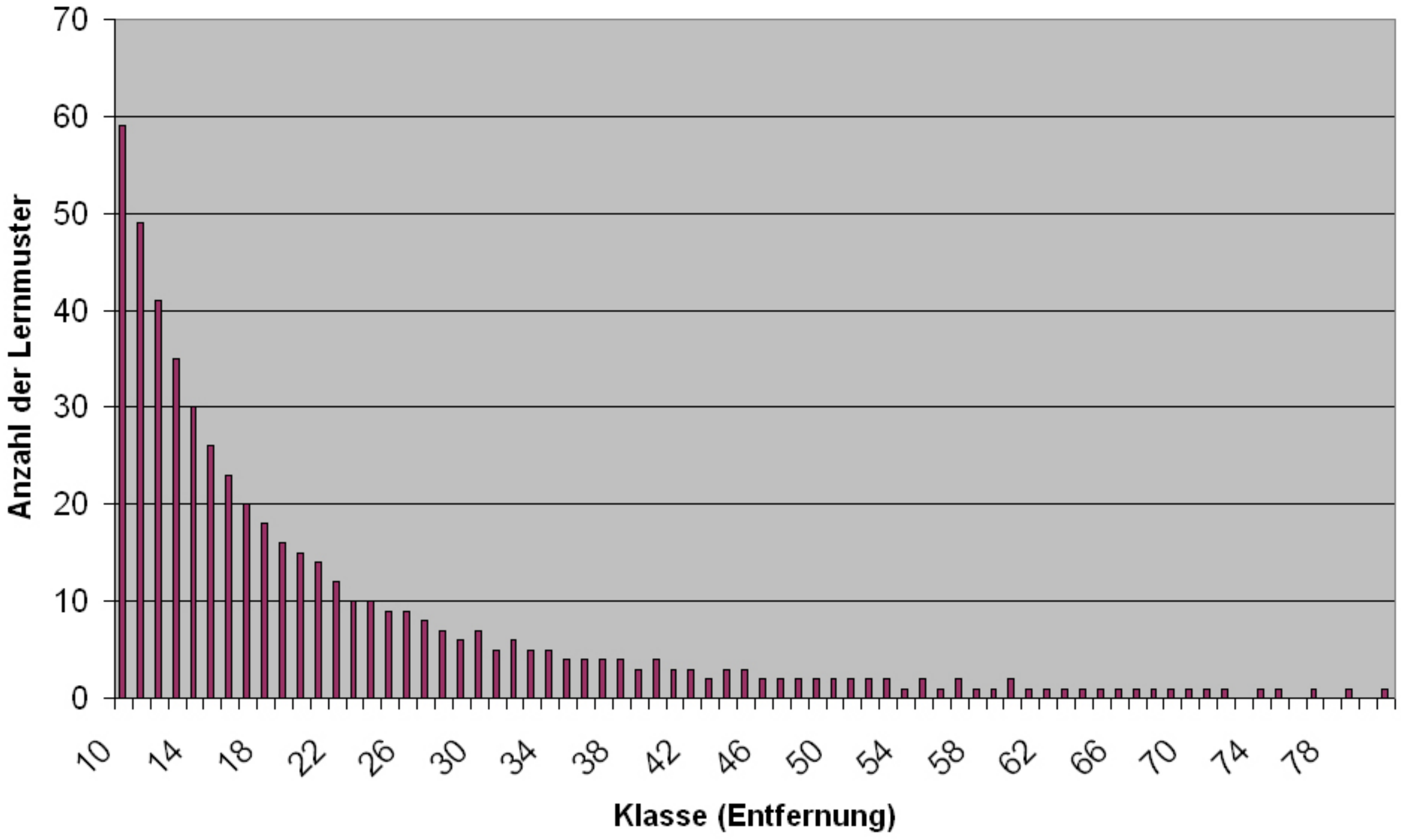

Abbildung 28 - Anzahl der Muster pro Trainingsklasse am Beispiel des GP2D12

In Abbildung 28 sind die Musteranzahlen der Klassen für die Entfernung von 10cm bis 80cm (Klasse 10 bis 80) dargestellt. Alle nicht gültigen ADC-Werte fallen zur Klasse „0“ zusammen, wodurch diese mit 498 Mustern die größte Klasse ist. Zu den ungültigen Werten zählen die ADC-Werte des Messbereiches welche den Messentfernungen von 0cm bis 9cm und ab 81cm entsprechen.

In der Trainingssoftware werden die Daten der Trainingsklassen in einem Array gespeichert. Die acht Output-Bits stellen beim Training die Sollwerte zu den zehn Input-Bits dar.

Die programmtechnische Umsetzung ist im Quellcode des Trainers detailliert kommentiert.

### 4.2.2 Training des Multilayer-Perceptons

Als Trainingsmethode für das Multilayer-Perceptron wird der Back-Propagation-Algorithmus verwendet. Er wird für diese Art des künstlichen neuronalen Netzes an häufigsten eingesetzt.

„Zum Training mehrschichtiger neuronaler Netze ist die Prozedur der kleinsten Quadrate zu verallgemeinern, um die Einstellung der Gewichtskoeffizienten der zu den latenten Neuronen gehörenden Verbindungen durchzuführen.“ [Aliev 2000] Damit dies erreicht werden kann, ist ein Forward Pass zur Berechnung der Zustände aller Neuronen in Netz und ein Backward Pass zur Modifizierung der Gewichte notwendig. Im Backward Pass kommt der Back-Propagation-Algorithmus zum Einsatz. Der Trainingszyklus ist in Abbildung 29 im Überblick dargestellt.

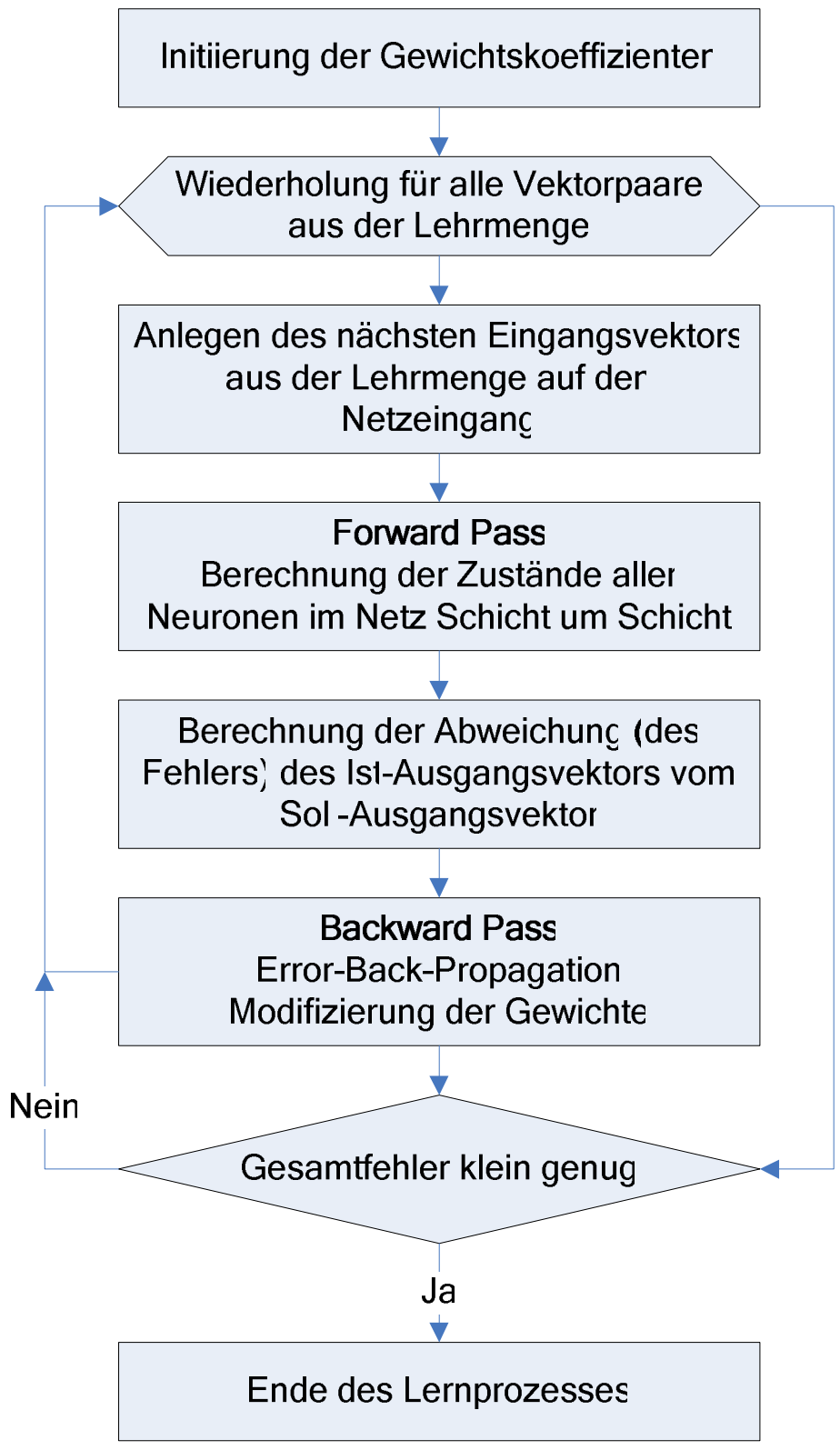

Abbildung 29 - Trainingszyklus des Multilayer-Perceptrons

Im Microcontroller wird das trainierte Netz eingesetzt und somit nur der Forward-Pass (Berechnung der Outputs) integriert. Der komplette Rechenablauf des Error-Back-Propagation-Algorithmus wird nur in der Trainingssofware benötigt. Während der Nutzung des KNN ändert oder erweitert sich die Lehrmenge nicht. Die Forward-Pass-Prozedur zur Sensordatenauswertung im Microcontroller wird im Kapitel 6.6.3 detailliert erläutert.

### 4.2.3 Konzeption des Trainers

Aus den in den vorangegangenen Abschnitten beschriebenen Grundlagen entstand ein Trainingsprogramm. Es wurde in Visual Basic.NET erstellt. Dessen Funktion ist die Berechnung und Erzeugung der Trainingsdaten mit Hilfe der Transferfunktion, das Training des MLP's und die Übertragung der Gewichtsmatrix in den EEPROM des Umgebungserfassungssystems.

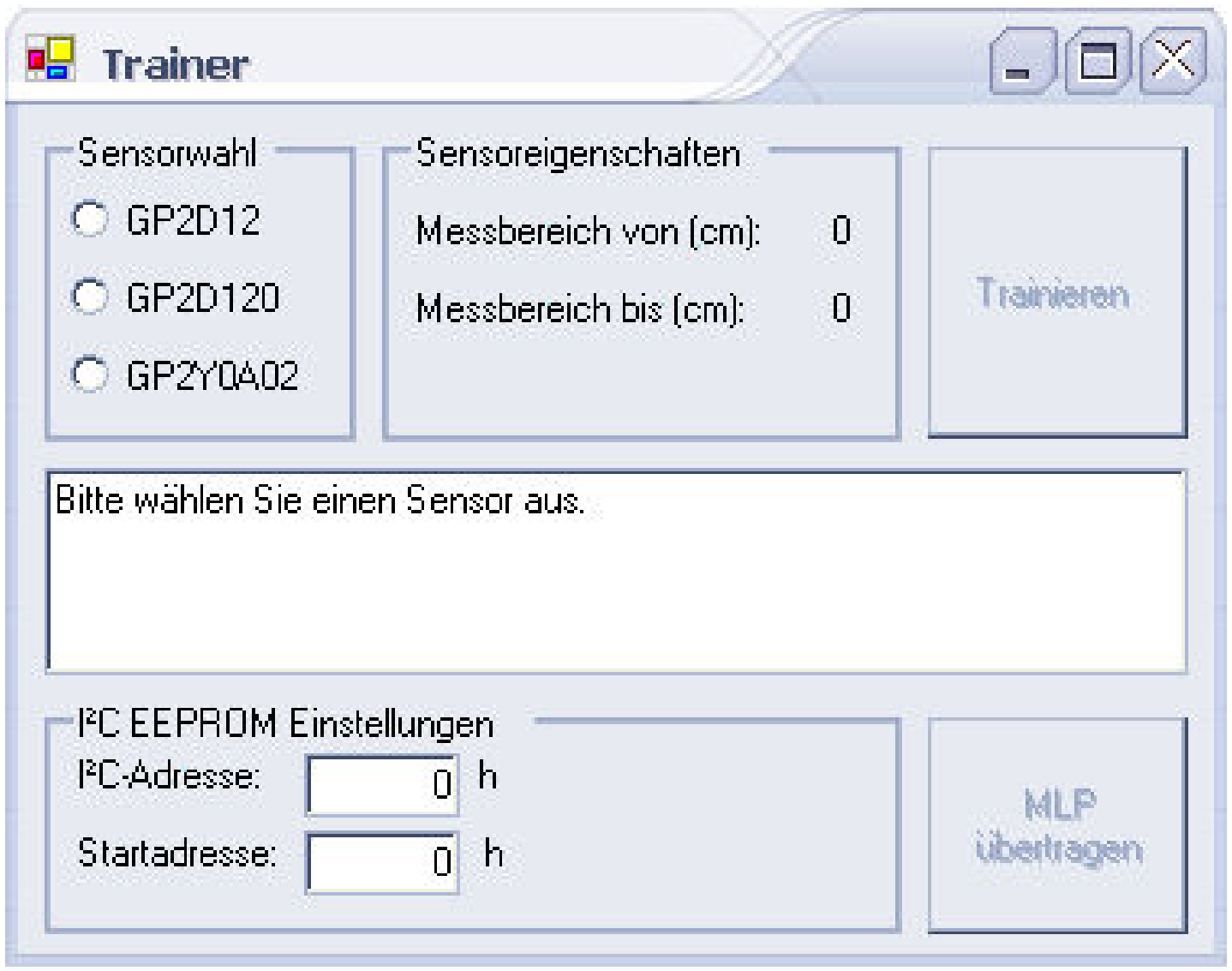

Abbildung 30 - Trainings- und Übertragungsprogramm für das MLP

Im Trainer sind die Transferfunktionen, also die Berechnungsformeln der idealen Messwerte, integriert. Durch die Auswahl im Feld „Sensorwahl" wird die zum IR-Sensor-Typ gehörende Funktion selektiert und die Messbereiche angezeigt. Wird eine Entfernung unterhalb des Minimums oder über dem Maximum gemessen, liefert es die Entfernung „0". Diese Werte werden bei einer Messung mit den IR-Sensoren bei der Umgebungserfassung nicht berücksichtigt. Die generierten Trainingsdaten wurden bereits im Abschnitt 4.2.1 dargestellt.

Der in der Abbildung 29 gezeigte Trainingszyklus wird beim Anklicken der Schaltfläche „Trainieren" aktiviert.

Im Folgenden ist der im Trainer realisierte Forward-Pass dargestellt:

```
For L = 1 To Layeranzahl
     For n = 1 To MaxNeurons
          For c = 1 To MaxGewichte
               Output(L + 1, n) = Output(L + 1, n) +_
               Output(L, c) * Gewicht(L, n, c)
          Next c
          If Output(L + 1, n) < 0.5 Then
               Output(L + 1, n) = 0
          Else
               Output(L + 1, n) = 1
          End If
     Next n
Next L
```

Die Forward-Pass-Prozedur ist detailliert im Quellcode der Datei „MLP.VB“ kommentiert. Sie dient der Berechnung des Outputs zu dem angelegten Input.

Zum Erlernen der Trainingsdaten ist neben dem Forward-Pass der Backward-Pass notwendig. Mit ihm werden die Gewichte des KNN's durch den Back-Propagation-Algorithmus angepasst.

```
For x = 1 To 8
     Fehler(Layeranzahl, x) = Output(Layeranzahl, x) *_
     (1 - Output(Layeranzahl, x)) *_
     (Ziel(x) - Output(Layeranzahl, x))
Next x
For L = Layeranzahl To 1 Step -1
     For n = 1 To MaxNeurons
          For c = 1 To MaxGewichte
               Gewicht(L, n, c) = Gewicht(L, n, c) +_
               Fehler(L + 1, n) * Output(L, c)
          Next c
     Next n
     For n = 1 To MaxNeurons
          For c = 1 To MaxGewichte
               Fehler(L, n) = Fehler(L, n) +_
               Fehler(L + 1, c) * Gewicht(L, c, n)
          Next c
          Fehler(L, n) = Fehler(L, n) * Output(L, n) *_
          1 - Output(L, n))
     Next n
Next L
```

# 5 Testprogramm für den Umgebungsscanner

Ein voll- oder teilautonomer mobiler Roboter muss sich im Raum orientieren können. Dazu muss er exakt seine Umgebung vermessen, um sich eine Umgebungskarte zu erstellen oder seine Position in einem vorhandenen Plan bestimmen zu können. Die an einem bestimmten Ort vermessene Umgebung repräsentiert den lokalen Grundriss um den Roboter. Das Testprogramm stellt diese lokale Umgebung dar. Die dazu notwendigen Daten liefert das Umgebungserfassungsmodul.

An dem System fest installierte Sensoren liefern zwar über der entsprechenden Winkelkoordinate Entfernungsdaten von Objekten, jedoch nicht genug Daten um die Umgebung komplett zu erfassen. Diese Entfernungsmesssensoren dienen viel mehr dem Erkennen von Objekten während der Bewegung und somit der Kollisionsvermeidung. Die Daten der unbeweglich angebrachten IR-Sensoren können durch das Testprogramm ebenfalls abgefragt werden.

Um eine detaillierte Umgebungskarte anfertigen zu können, muss die Umgebung exakt abgetastet werden (siehe Kapitel 5.1). Diese Daten der lokalen Karte können mit den relativen Bewegungsdaten des Roboters in eine globale Umgebungskarte umgerechnet werden (im Kapitel 5.2 erläutert). Diese wird jedoch nicht durch das Umgebungserfassungssystem erstellt, sondern wird mit Hilfe eines gesonderten Programms erzeugt.

Mit Hilfe dieses globalen Plans wird eine spätere Wegstreckenplanung und -optimierung möglich. Die Erstellung der globalen Karte wird durch den Steuerrechner ausgeführt. Wie der Aufbau erfolgen soll, wird an dieser Stelle nur grob umrissen, da die Datenweiterverarbeitung zur globalen Umgebung nicht mehr Bestandteil dieser Diplomarbeit ist. Die Beschreibung ist vielmehr als Ausblick gedacht.

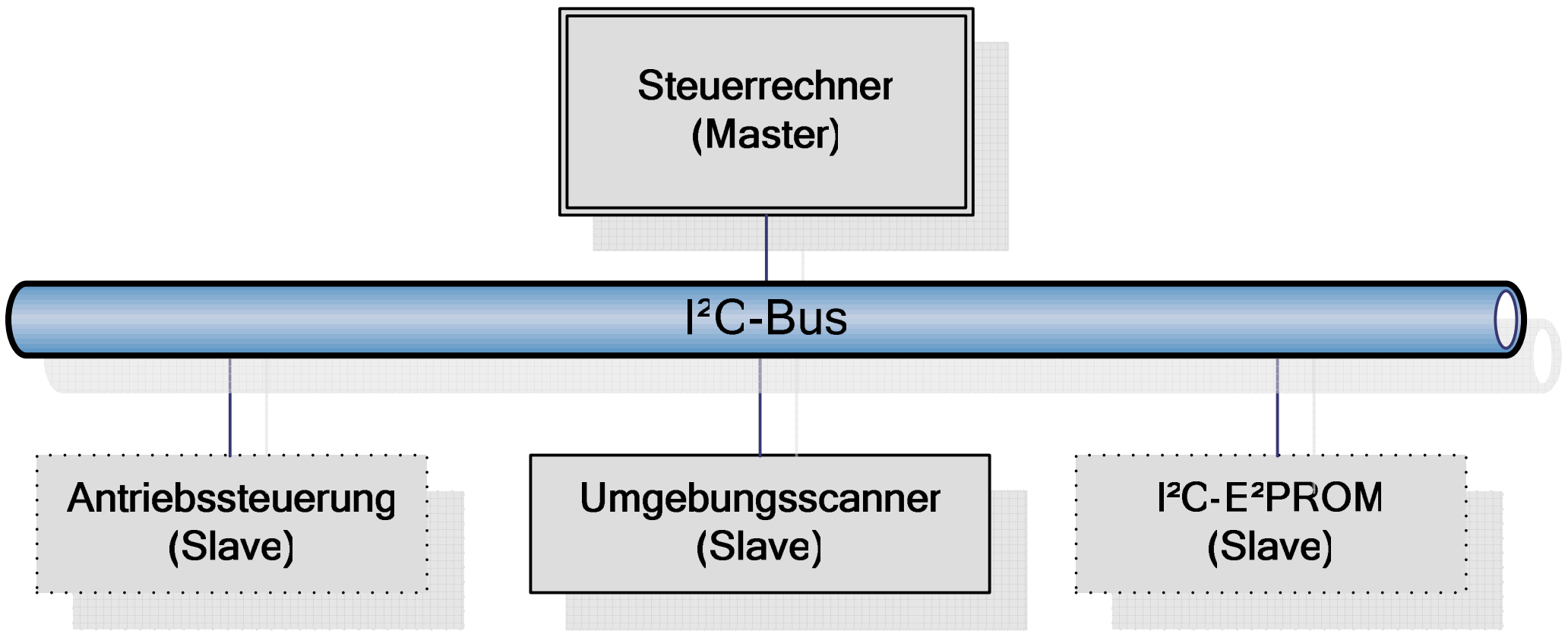

Abbildung 31 - Umgebungsscanner mit Steuerrechner

# 5.1 Lokale Karten

## 5.1.1 Mathematische Grundlagen

Die Sensoren können nicht alle Objekte in einem Winkel $\Delta\varphi$, von beispielsweise 180°, gleichzeitig erfassen. Daher muss die Umgebung in Teilsegmenten $\Delta\varepsilon$ erfasst werden.

Der Vermessungsvorgang des Scannerumkreises erfolgt während des Stillstandes des mobilen Roboters. Dadurch kann auf die Bewegungskompensation der gemessenen Daten verzichtet werden. Der durch die Bewegung entstehende Verschiebungsvektor muss so nicht von den Messdaten subtrahiert werden. Fehler, welche beispielsweise durch Vibrationen des Messsensors während der Bewegung entstehen, werden somit ebenfalls vermieden.

Zur exakten Aufnahme der Objektdaten, muss ein Messsensor mit möglichst kleinem Objekterfassungswinkel $\Delta\varepsilon$ gewählt werden.

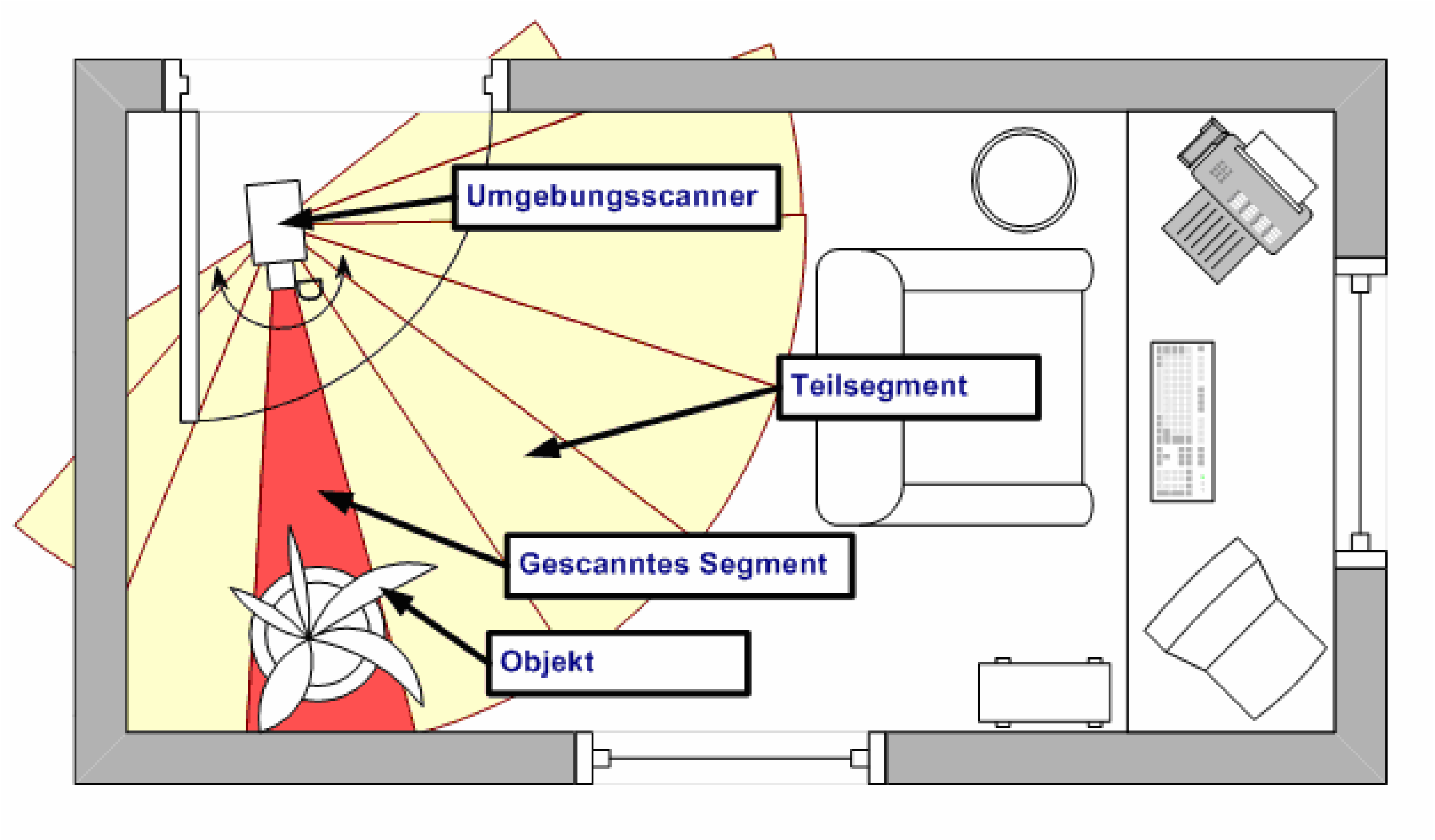

Abbildung 32 – Erfassung von Objekten im aktiven Teilsegment

Zur Datenerfassung des gewünschten Umgebungsbereiches wird der Messsensor drehbar montiert (Sensorkopf). Die eigentliche Umgebungsabtastung erfolgt dann in $n * \Delta\varepsilon = \Delta\varphi$ Schritten.

Die Segmentgröße $\Delta\varepsilon$ und der zu vermessende Bereich $\Delta\varphi$ hängt von den Einstellungen der im Kapitel 6.5.3 beschriebenen Schrittmotorsteuerung ab.

Aus den Größen von $\Delta\varepsilon$ und $\Delta\varphi$ resultiert die Dauer der Vermessung. Je größer der Messbereich $\Delta\varphi$ und je kleiner der Erfassungswinkel $\Delta\varepsilon$ ist, desto länger dauert die Messung. Ein Hauptgrund dafür ist die mit kleiner werdendem $\Delta\varphi$ steigende Da-tenmenge, welche verarbeitet werden muss. Dadurch erhöht sich jedoch die Messauflösung.

Die Messdauer $t_{Messdauer}$ kann durchfolgende Formeln abgeschätzt werden:

$$t_{Messdauer} = t_{Anfahrtszeit_Startposition} + t_{Messung} + t_{Anfahrtszeit_Nullposition}$$

$$t_{Anfahrtszeit_Startposition} = t_{Anfahrtszeit_Nullposition} = 0{,}05 * \frac{\Delta\varphi}{2} * 0{,}85 = 0{,}02125 * \Delta\varphi$$

$$t_{Messdauer} = (0{,}05 + 0{,}01) * \frac{\Delta\varphi}{\Delta\varepsilon} = 0{,}06 * \frac{\Delta\varphi}{\Delta\varepsilon}$$

Die Zeiten zum Anfahren der Start- ($t_{Anfahrtszeit_Startposition}$) und Nullposition ($t_{Anfahrtszeit_Nullposition}$) sind gleichgesetzt, da in den meisten Fällen ein identischer Winkelbereich ($\Delta\varphi/2$) rechts und links der Nullposition vermessen wird. Der Multiplikator 0,85 entspricht einer Zeitersparnis von 15%, welche bei der Nutzung von Beschleunigungs- und Bremsrampen beim Anfahren einer Position entsteht.

In den Formeln für die Anfahrts- und Messzeiten ist der Faktor 0,05 enthalten. Er entspricht einer Programmverzögerung von 0,05 Sekunden, welcher zum mechanischen Ausführen eines Schrittes benötigt wird. Die Messdauerfunktion enthält weiterhin 0,01s die der Microcontroller zur Entfernungsberechnung benötigt.

Setzt man die Teilfunktionen in den Ansatz ein, klammert $\Delta\varphi$ aus und vereinfacht, erhält man die Gesamtberechnungsfunktion der Messdauer:

$$t_{Messdauer} = \Delta\varphi\left(\frac{0{,}06}{\Delta\varepsilon} + 0{,}0425\right)$$

Vor der eigentlichen Koordinatenberechnung des Hindernisses muss die gewünschte Rasterung der Karte $l_r$ festgelegt werden. Ein feineres Raster bedeutet zwar eine höhere Auflösung der lokalen Karte, jedoch auch eine höhere Datenmenge.

Der Speicherplatzbedarf d (in Bit) der Kartendaten lässt sich folgender Maßen berechnen:

$$d = \left(\frac{2 * l_m}{l_r}\right)^2 \quad [d \in N]$$

Ein Rasterfeld benötigt also ein Bit Speicherplatz (Hindernis/ kein Hindernis).

Der Ausrichtungswinkel des Sensorkopfes $\alpha$ und die gemessene Entfernung $l_m$ werden zur Weiterverarbeitung im Steuerrechner in x/y-Koordinaten umgerechnet. Dadurch erfolgt die eben beschriebene Umgebungsrasterung. Der Sachverhalt ist in Abbildung 33 graphisch dargestellt.

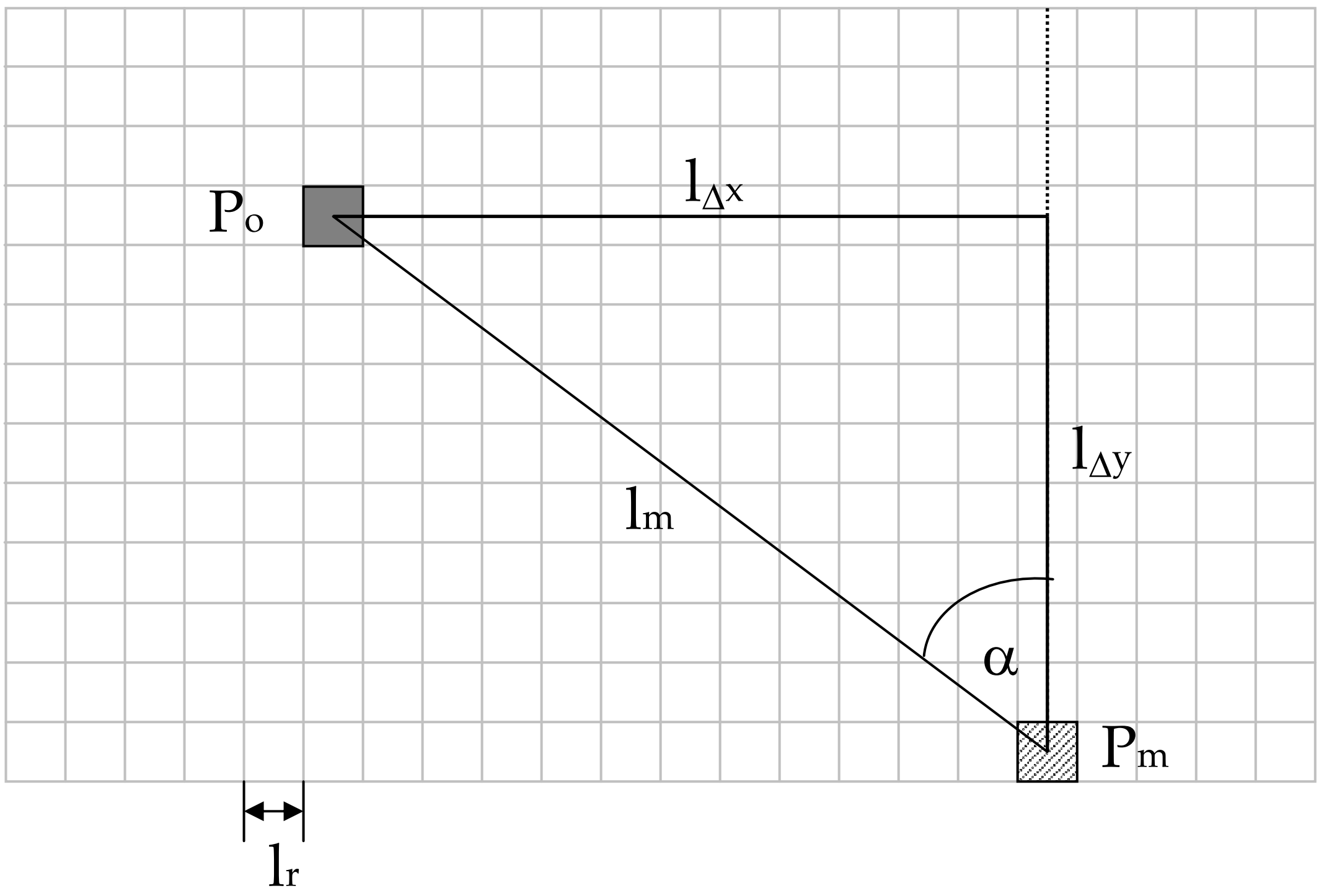

$P_m$…Messpunkt mit $P_{mx}$ und $P_{my}$

$P_o$…Objekt mit $P_{ox}$ und $P_{oy}$

$l_r$…Rasterbreite

$l_m$…Gemessene Entfernung zum Objekt

$l_{\Delta x}$…x-Koordinatendifferenz zwischen $P_{mx}$ und $P_{ox}$

$l_{\Delta y}$…y-Koordinatendifferenz zwischen $P_{my}$ und $P_{oy}$

Abbildung 33 - Koordinatenberechnung der lokalen Karte

Zur Berechnung der Teilstrecken $l_{\Delta x}$ und $l_{\Delta y}$ werden folgende Formeln benutzt:

$$l_{\Delta x} = \frac{\sin\alpha * l_m}{l_r} \quad \left[l_x \in Z\right]$$

$$l_{\Delta y} = \frac{\cos\alpha * l_m}{l_r} \quad \left\lfloor l_y \in Z \right\rfloor$$

Die Koordinaten des Objektes $P_o$ im lokalen Koordinatensystem lassen sich mit $x_{P_o} = x_{P_m} + l_{\Delta x}$ und $y_{P_o} = y_{P_m} + l_{\Delta y}$ berechnen. Stellt der Punkt $P_m$ den Koordinatenursprung des lokalen Rasters, also die Drehachse des Scannerkopfes dar, ist $x_{P_m} = 0$ und $y_{P_m} = 0$.

Um negative Koordinaten in der lokalen Karte zu vermeiden, wird der Ursprung verschoben. Damit kann die Karte als Array gespeichert werden, da dessen Laufindexe nicht negativ sein dürfen.

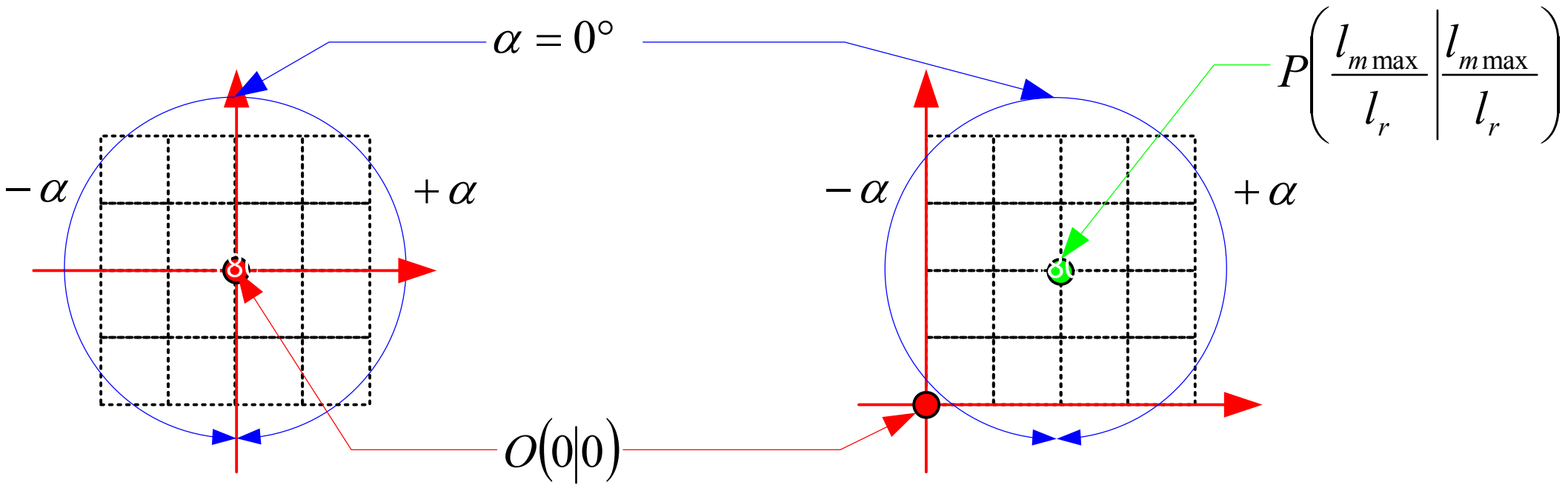

Abbildung 34 - Verschiebung des Koordinatenursprungs

Die Koordinaten des Objektpunktes $P_o$ in der lokalen Karte errechnen sich mit folgender Formel:

$$P_o\left(\frac{l_{m\max} + \sin\alpha * l_m}{l_r}\middle|\frac{l_{m\max} + \cos\alpha * l_m}{l_r}\right)$$

Dabei ist $l_{m\max}$ die maximale Reichweite des Messsensors, $l_m$ die gemessene Entfernung und $\alpha$ der Drehwinkel des Sensorkopfes. Für $\alpha$ sind Werte bis $\pm 180°$ gültig.

### 5.1.2 Konzeption des Testprogramms

Das Testprogramm übernimmt die Funktion des Masters und dient in erster Linie der Funktionskontrolle des Moduls und der Darstellung der Messdaten. Es sendet die Befehle an das Modul, welches die entsprechenden Kommandos umsetzt.

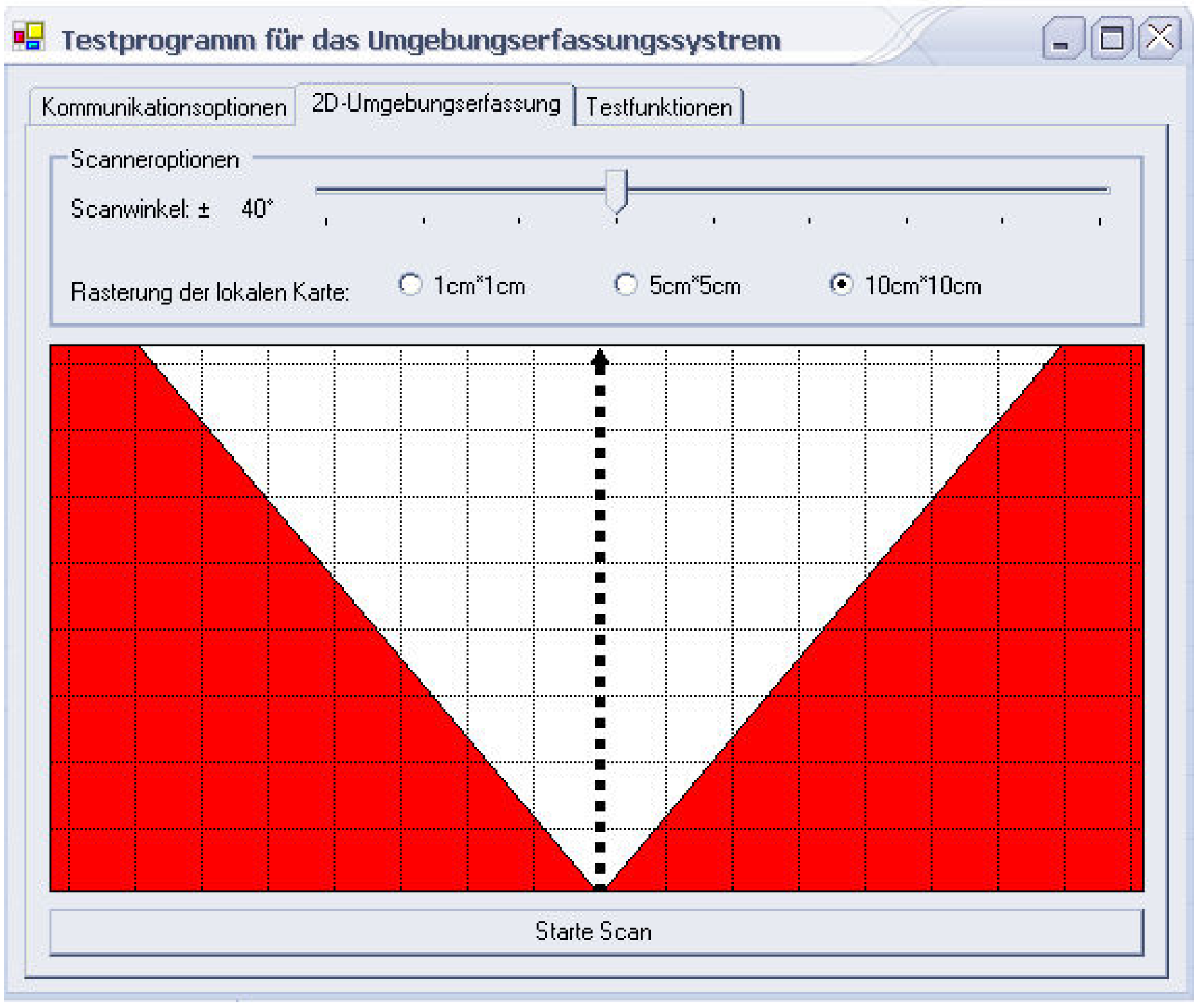

Abbildung 35 - Test- und Diagnoseprogramm für das Umgebungserfassungsmodul

Abbildung 35 zeigt die Endversion des Programms. Die dargestellte Registerkarte beispielsweise dient der Erfassung einer lokalen Karte.

Mit Hilfe des Testprogramms sind die Darstellung einer lokalen Umgebung, sowie die Nutzung von Diagnosefunktionen möglich. Die Entfernungswerte der einzelnen Sensoren können abgefragt werden, wodurch unterbrochene Sensorkabel oder defekte Sensoren erkannt werden können. Zum Test der Sensorkopfmotoren besteht die Möglichkeit die Drehbewegungen manuell zu steuern.

Alle Funktionen und Elemente des Programms sind im Quellcode und auf der CD detailliert beschrieben.

## 5.2 Globale Karten

Die komplette reale Umgebung muss in Daten einer globalen Karte umgesetzt werden. Der Steuerrechner des Roboters erstellt ein globales Weltmodell. Es wird aus den Daten berechnet, welche die einzelnen Module liefern. Dazu zählen die Bewegungsdaten, welche das Antriebsmodul zur Verfügung stellt und den Scandaten (lokale Karte), die das Umgebungserfassungsmodul bei einem Scanzyklus liefert.

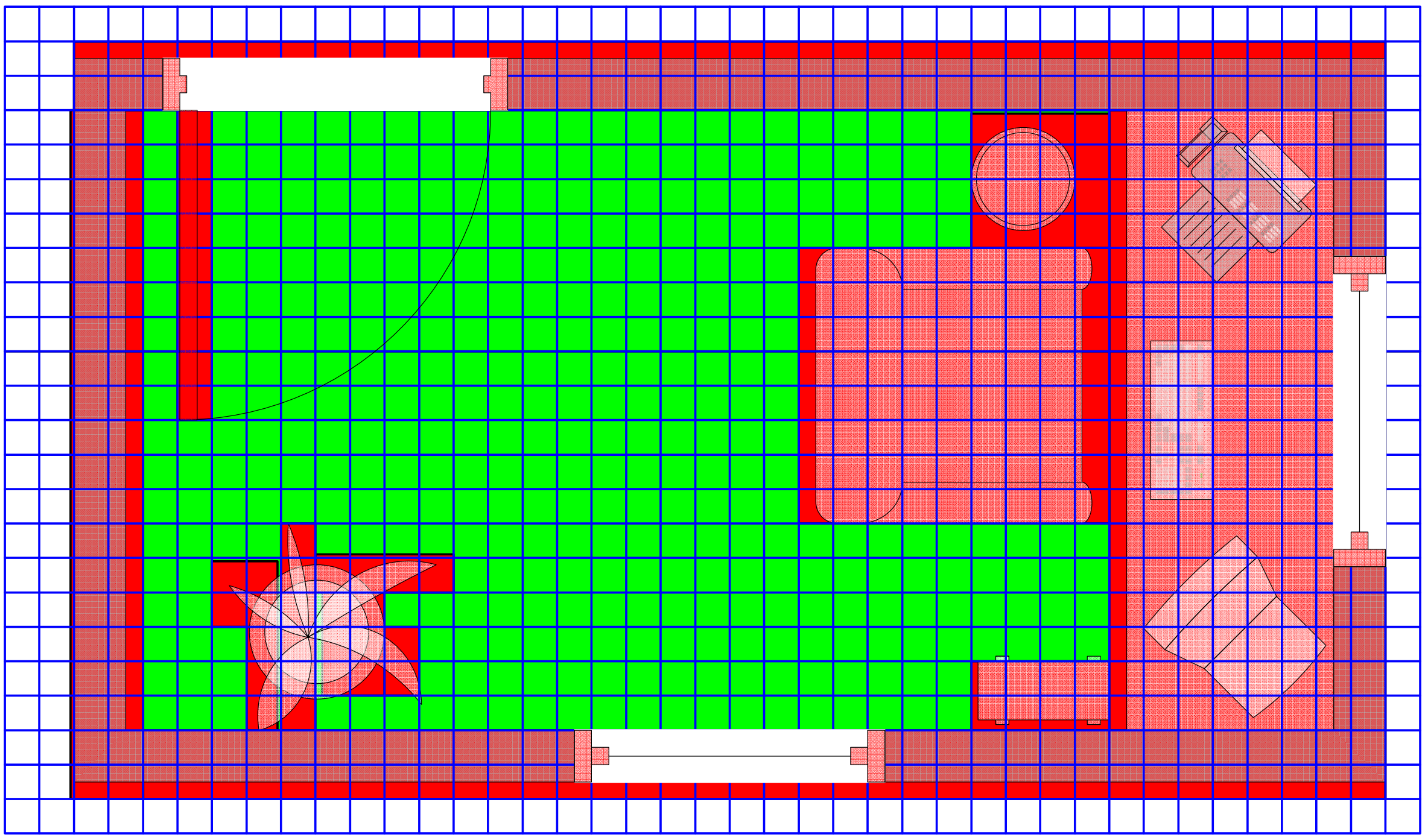

Abbildung 36 - Reale Umgebung und gerasterte Karte (rot-Hindernis/grün-frei)

Die Umgebung des Roboters ist meist größer, als die Reichweite seiner Messsensoren. Es ist deshalb nötig mehrere lokale Umgebungskarten (in Abbildung 37 grün eingezeichnet) zu einem globalen Plan der Umgebung (rot dargestellt) zusammen zu fassen.

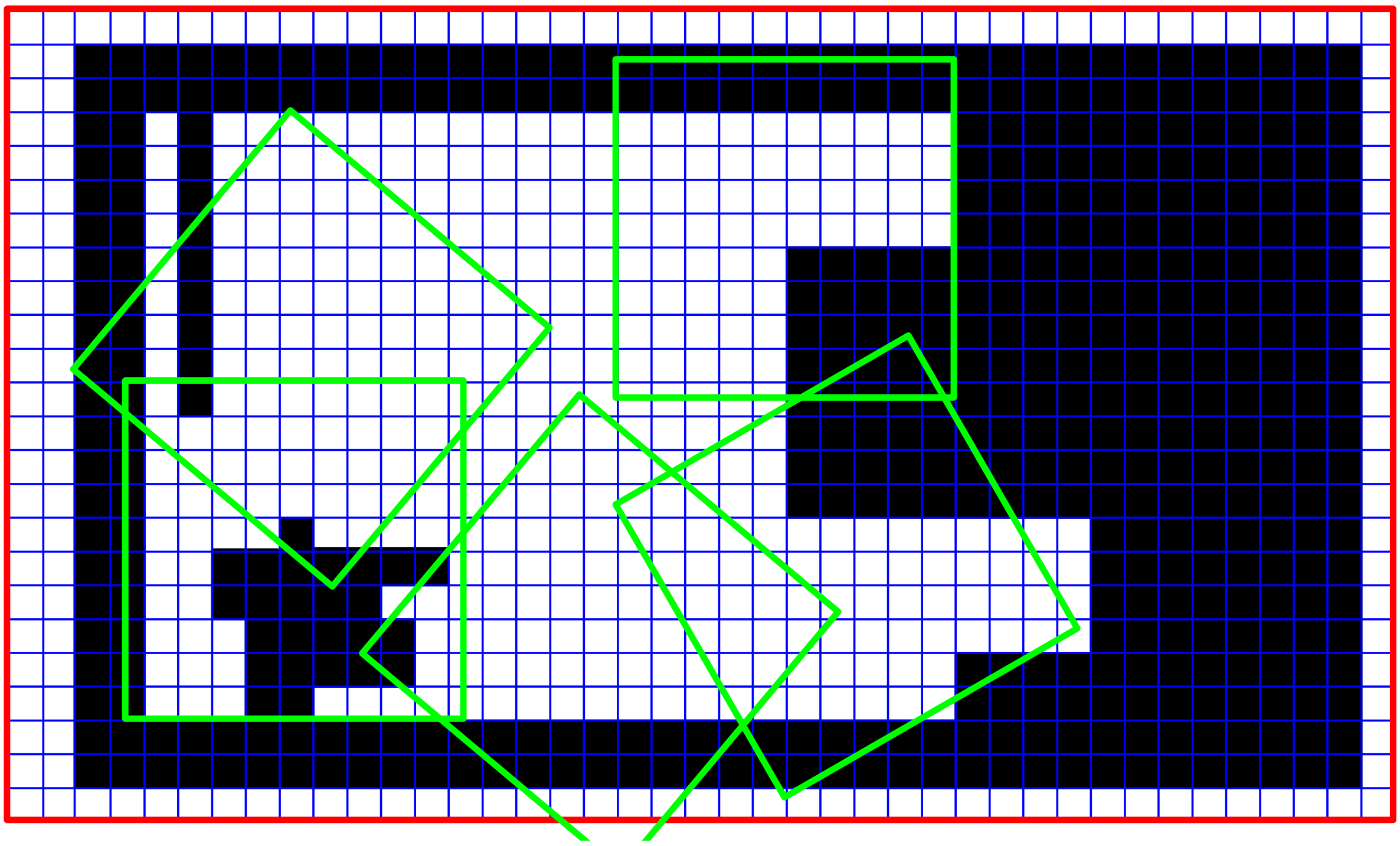

Abbildung 37 - Karte der gerasterten Umgebung

Die Karte der Umgebung umfasst nach der Kombination aller lokalen Pläne alle festen Hindernisse (unbewegliche Hindernisse) und Wände. Dadurch ist die reale Umgebung in Bereiche mit und ohne Barrieren eingeteilt. In der Abbildung 37 sind die Rasterfelder mit Objekten schwarz dargestellt.

Zur Kombination der lokalen Karten zum globalen Weltmodell stehen verschiedene Möglichkeiten zur Verfügung. Allen gemeinsam ist, dass entweder die absolute Position und die Ausrichtung im lokalen Raster bekannt sein muss, beispielsweise mit Hilfe von GPS und einem Kompasssensor oder der relativen Bewegung zum letzten Standpunkt.

### 5.2.1 Vor- und Nachteile verschiedener Kombinationsmethoden

Durch die Anfälligkeit des GPS-Systems, vor allem in Innenräumen wird es hier selten eingesetzt. Eine Positionsgenauigkeit von zwei Metern ist für diesen Zweck der Umgebungserfassung in Gebäuden nicht ausreichend. Die Einordnung von Messdaten mittels GPS-Positionsdaten findet ausschließlich im Außenbereich Anwendung. Hier ist die erreichte Genauigkeit ausreichend.

In Gebäuden, welche lokal sehr begrenzt sind, werden in erster Linie odometrische Verfahren eingesetzt. Sie nutzen die Bewegungsdaten des Roboters, um die Umgebungsdaten miteinander zu kombinieren. Dies hat zur Folge, dass sich die Positionierungsungenauigkeiten, welche bei der Bewegung entstehen, mit fortschreitender Weglänge aufaddieren.

Durch diverse Korrekturalgorithmen können die Fehler jedoch minimiert werden. Dazu zählt beispielsweise die Soll-/Ist-Positionskorrektur mittels Landmarken, wel-che in der Umgebung fest angebracht sind. Deren Lage muss allerdings bekannt sein, z.B. durch die Anordnung in einem Raster.

Neuere Verfahren speichern die Umgebung in Objektbäumen ab, wodurch die Position beim erreichen eines bekannten Ortes abgeglichen werden kann.

Neben den beiden grob umrissenen Verfahren stehen noch andere Algorithmen zur Verfügung. Sie können in entsprechender Fachliteratur nachgelesen werden.

### 5.2.2 Kombination der lokalen Karten zur globalen Karte

Die Nutzung der Bewegungsdaten, eines Kompasssensors und geeigneter Matching-Algorithmen ist eine weit verbreitete Methode zur Umgebungsdatenkombinierung.

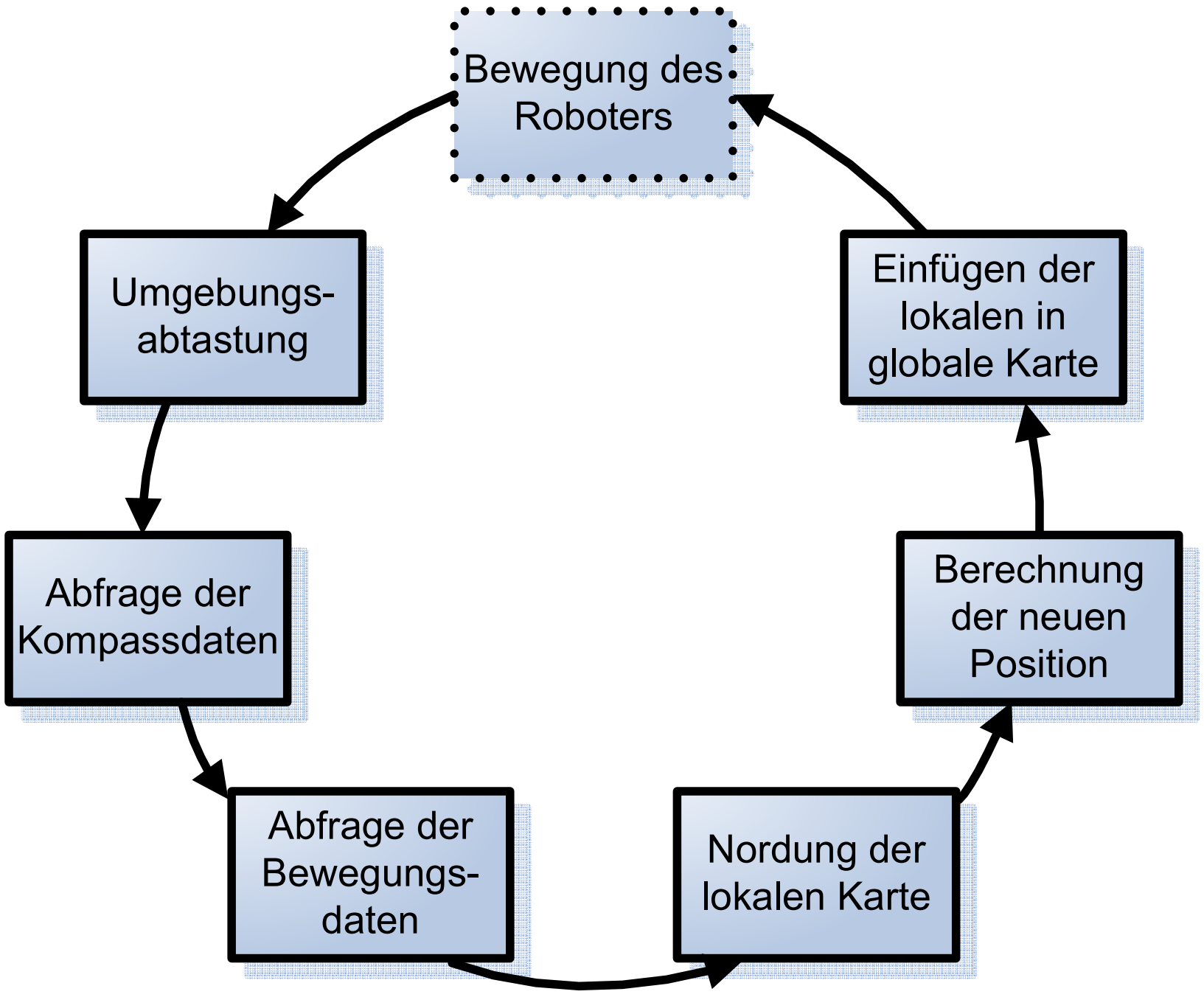

Abbildung 38 - Erstellungszyklus der globalen Karte

Das Schema zur Erstellung der globalen Karte ist in Abbildung 38 dargestellt. Der Zyklus wird dabei in unterschiedliche Phasen geteilt, welche in verschiedenen Modulen des Roboters abgearbeitet werden.

Die Umgebungsabtastung wird von dem im Kapitel 6 beschriebenen Umgebungserfassungssystem übernommen. Es erfasst die lokale Karte, welche im Abschnitt 5.1 detailliert beschrieben wurde.

Um die lokale Karte in Bezug zu den anderen bereits erfassten Umgebungsdaten bringen zu können, muss der Steuerrechner die Kompass- und Bewegungsdaten der entsprechenden Module abrufen.

Zur Nordung der lokalen Karte wird der Ausrichtungswinkel des Roboters gemessen. Diese Daten liefert ein Kompasssensor. Er ist eine elektronische Version eines Kompasses. Die Himmelsrichtungen werden per Magnetfeldsensorvollbrückenschal-tungen durch das Erdmagnetfeld bestimmt. Je nach verwendetem Typ kann die Genauigkeit bis $\pm 0{,}2°$ betragen.

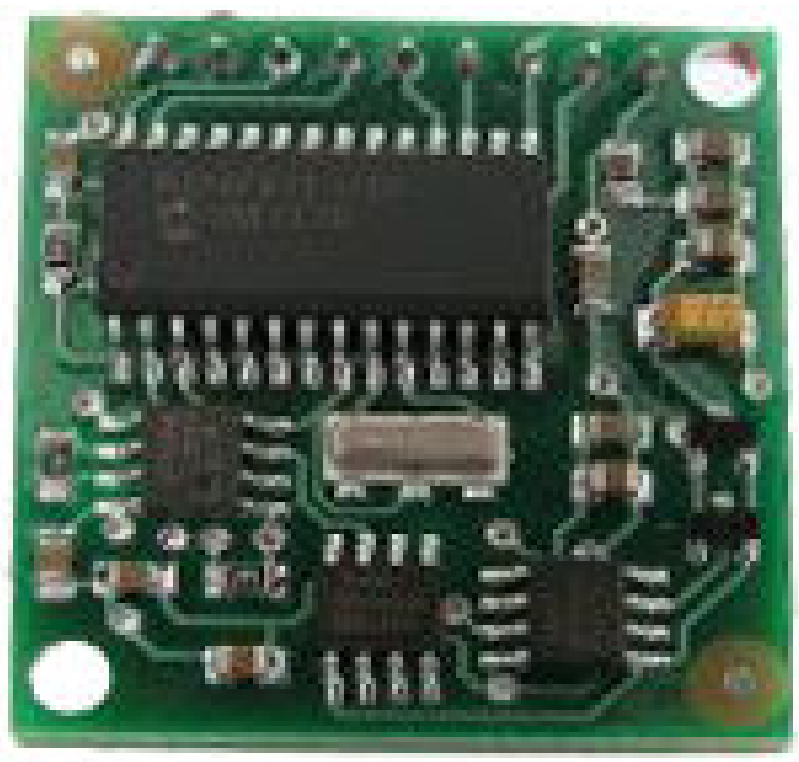

Abbildung 39 - Kompassmodul CMPS03

Durch den Einsatz eines Sensors mit entsprechend hoher Winkelauflösung kann diese Fehlerquelle minimiert werden. Dadurch wird die Genauigkeit der globalen Karte gesteigert. Entsprechende Sensormodule sind ab circa 70€ erhältlich.

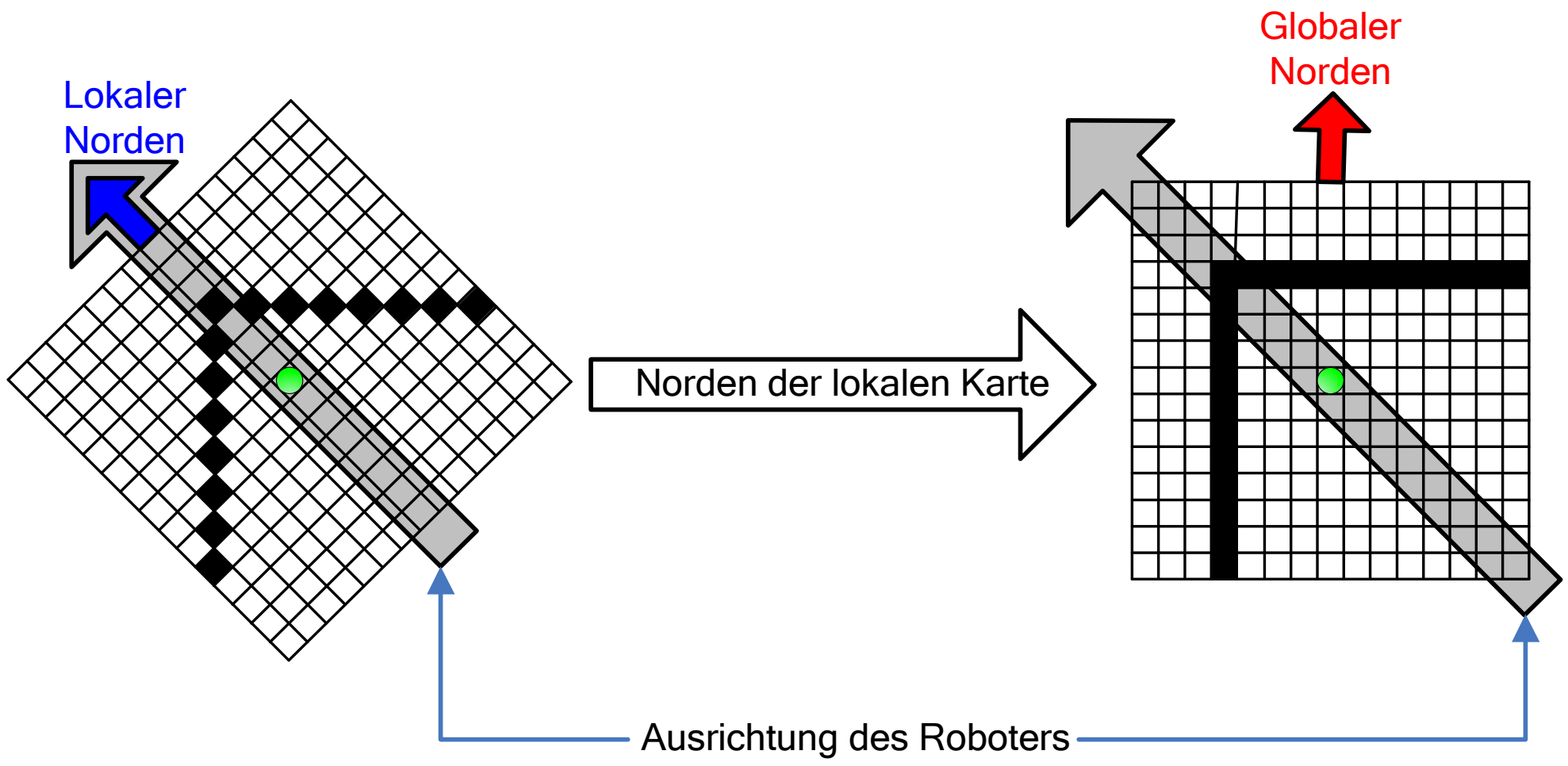

Abbildung 40 - Ausrichtung der lokalen Karte zum globalen Norden

Die Umrechnung der einzelnen Kartenpunkte der lokalen Matrix in das globale System erfolgt mit folgender Formel (erweiterte Koordinaten):

$$\begin{pmatrix} x' \\ y' \\ 1 \end{pmatrix} = \begin{pmatrix} \cos\alpha & \sin\alpha & 0 \\ -\sin\alpha & \cos\alpha & 0 \\ 0 & 0 & 1 \end{pmatrix} * \begin{pmatrix} x \\ y \\ 1 \end{pmatrix}$$

Dabei ist $P(x/y)$ der Originalpunkt und $P'(x'/y')$ der nach dem globalen System ausgerichtete Punkt. Der Winkel $\alpha$ befindet sich zwischen der Hauptachse des Roboters (lokaler Norden, Front des Roboters) und dem Norden des globalen Systems. Dabei ist ausschließlich die Ausrichtung des lokalen Koordinatensystems veränderlich. Je nach Orientierung des Roboters verändert sich der Winkel $\alpha$.

Mit Hilfe der Rotationsmatrix muss jeder Punkt der Speichermatrix der lokalen Karte gedreht werden. Das Ergebnis der Prozedur ist die zum globalen Norden ausgerichtete lokale Karte.

Nach Anwendung der Rotationsmatrix ruft der Hostrechner die Bewegungsdaten aus dem Antriebsmodul ab. Aus der relativen Position zur vorhergehenden Lage kann die notwendige x-y-Verschiebung in der globalen Karte errechnet werden. Danach werden die lokalen Kartendaten in die globale Umgebungsmatrix kopiert.

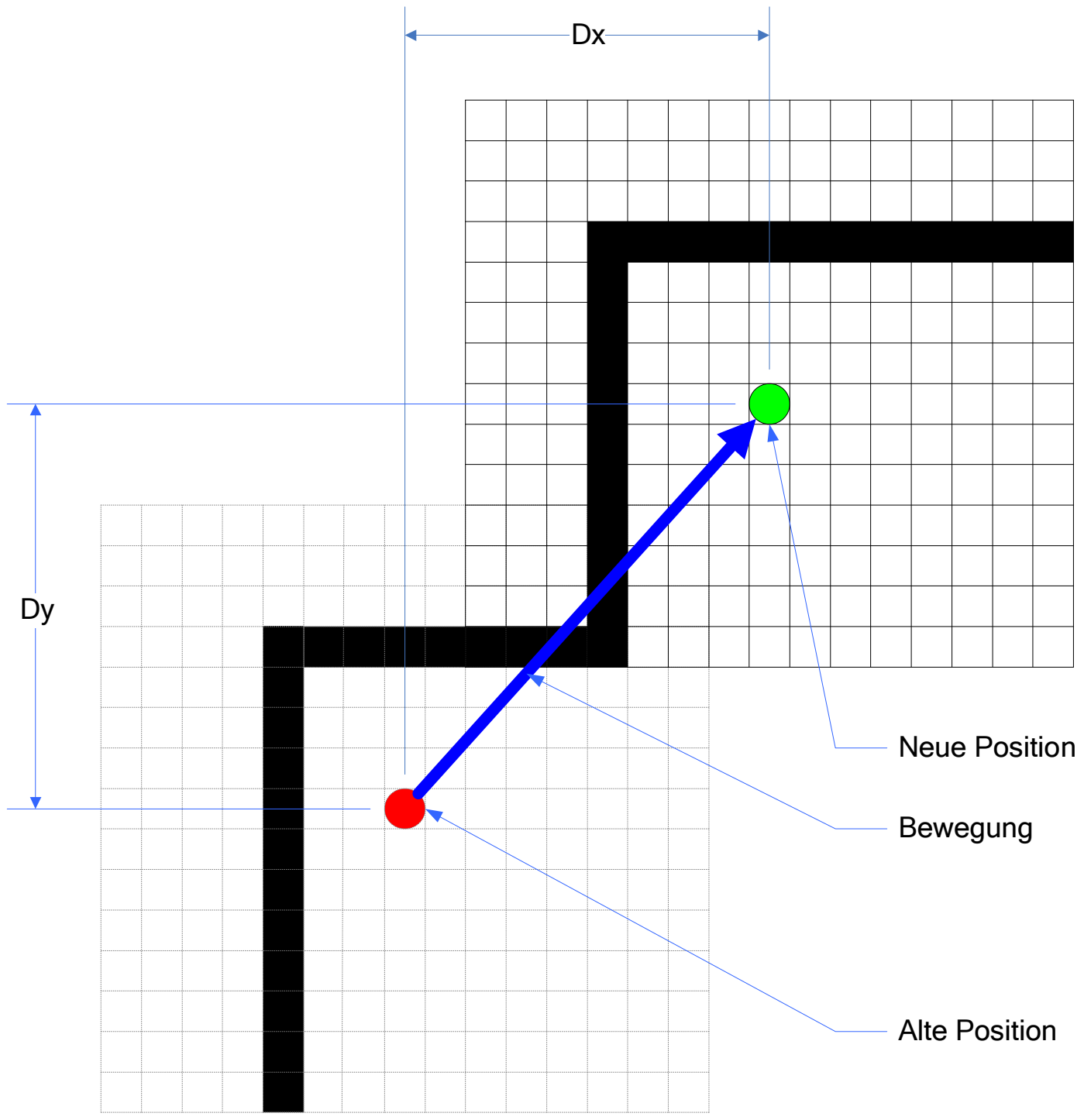

Abbildung 41 - Einfügen der lokalen in globale Kartendaten

Zur Berechnung der globalen Koordinaten für die lokale Karte wird eine Translationsmatrix verwendet:

$$\begin{pmatrix} x' \\ y' \\ 1 \end{pmatrix} = \begin{pmatrix} 1 & 0 & \Delta x \\ 0 & 1 & \Delta y \\ 0 & 0 & 1 \end{pmatrix} * \begin{pmatrix} x \\ y \\ 1 \end{pmatrix}$$

In diese Matrix muss die x- und y-Differenz ($\Delta x, \Delta y$) zwischen der alten und neuen Position eingebunden werden.

Für eine zweidimensionale Umgebungskarte gilt, dass die dritte Dimension, also die z-Koordinaten nicht berücksichtigt wird.

Die Umrechnungsmatrix der lokalen in die globalen Koordinaten lässt sich in folgender Matrix zusammenfassen:

$$\begin{pmatrix} x_{global} \\ y_{global} \\ 1 \end{pmatrix} = \begin{pmatrix} \cos\alpha & \sin\alpha & \Delta x \\ -\sin\alpha & \cos\alpha & \Delta y \\ 0 & 0 & 1 \end{pmatrix} * \begin{pmatrix} x_{lokal} \\ y_{lokal} \\ 1 \end{pmatrix}$$

# 6 Konzeption der Modulhardware und -software

## 6.1 Entfernungsmessung mit Infrarotsensoren

Um eine kostengünstige und dennoch exakte Umgebungsvermessung ausführen zu können, eignen sich die in Kapitel 2.1.1 erwähnten Infrarotsensoren. Die Verwendung an einem Roboter wurde bereits in den Kapiteln 2.3 und 3.1 beschrieben.

### 6.1.1 Entscheidungskriterien

Ausschlaggebende Kriterien für die Wahl des Sensors waren:

- Preis
- Reichweite
- Verfügbarkeit als Fertigmodul, um die Entwicklungszeit einer IR-Sensorschaltung zu sparen
- Entfernungsausgabe als analoge Spannung oder als Daten in digitaler Form
- Messgenauigkeit und Störungsunempfindlichkeit
- Positive Erfahrungsberichte aus bestehenden Projekten

Die verwendeten Infrarotmesssensoren sind die einzigen Fertigmodule, welche in der Endversion des Umgebungserfassungssystem genutzt werden.

### 6.1.2 Einsatz im Scanner

Der Infrarotmesssensor ist die Messvorrichtung im Umgebungserfassungssystem. Er ist auf einer Positionierungsplattform (siehe Kapitel 3) montiert. Beide Elemente zusammen bilden den Scannerkopf. Er ist im Kapitel 3.1 detailliert beschrieben.

Die exakte Ausrichtung des Scannerkopfes erfolgt durch einen Schrittmotor. Dieser wird durch die im Kapitel 6.4 beschriebene Hardware gesteuert. Das Modul mit der dazugehörigen Software berechnet ebenfalls die vom Infrarot-Sensor gemessene Entfernung und die lokalen Hindernisdaten (siehe Kapitel 5.1).

## 6.2 Schrittmotor des Sensorkopfes

Im Abschnitt 3.3 wurden bereits die Grundlagen des Schrittmotors erläutert. Für den Einsatz im Umgebungserfassungssystem muss ein ausreichend, jedoch nicht überdimensionierter Motor gewählt werden.

### 6.2.1 Auswahlkriterien

Neben dem notwendigen Drehmoment sind ein geringer Stromverbrauch und eine Betriebsspannung im Bereich des Gesamtsystems wichtige Kriterien. Die Dimensionierung dieser physikalischen Parameter ist durch den begrenzten Energievorrat des Gesamtsystems essentiell.

Die Systemspannung liegt bei 12V. Die benötigte Betriebsspannung muss bei einer geringeren Spannung liegen. Durch die eingesetzte elektronische Regelung (siehe Kapitel 6.4.3) kann der Schrittmotor direkt an den 12V betrieben werden.

Da der Antrieb des Sensorkopfes über ein Untersetzungsgetriebe erfolgt, ist eine Schrittauflösung von 100 Schritten/Umdrehung (3,6°) ausreichend. Durch die große Untersetzung des Antriebs ist lediglich ein geringes Motordrehmoment notwendig.

Die Wahl fiel deshalb auf einen Minebea-Schrittmotor mit folgenden Eigenschaften:

| Eigenschaft | Wert |
|---|---|
| Nennspannung | 7V |
| Leistungsaufnahme | ca. 8W |
| Schrittwinkel | 3,6° |

| Eigenschaft | Wert |
|---|---|
| Strangstrom | 0,58A |
| Strangwiderstand | 12Ω |
| Preis | ca. 7,90€ |

Tabelle 6 - Eigenschaften des eingesetzten Schrittmotors

### 6.2.2 Einsatz im Scanner

Wie bereits im Kapitel 3 beschrieben, wird der Schrittmotor als Antriebseinheit des Sensorkopfes verwendet. Das konzipierte Modul unterstützt die Steuerung zweier Schrittmotoren. Somit können zwei unabhängige Sensorköpfe am Modul betrieben werden oder ein Sensorkopf, welcher als 3D-Vermessungssystem arbeitet.

Die bipolaren 4-Strang Schrittmotoren werden an das Modul mit 4-poligen Mini-DIN-Steckern angeschlossen.

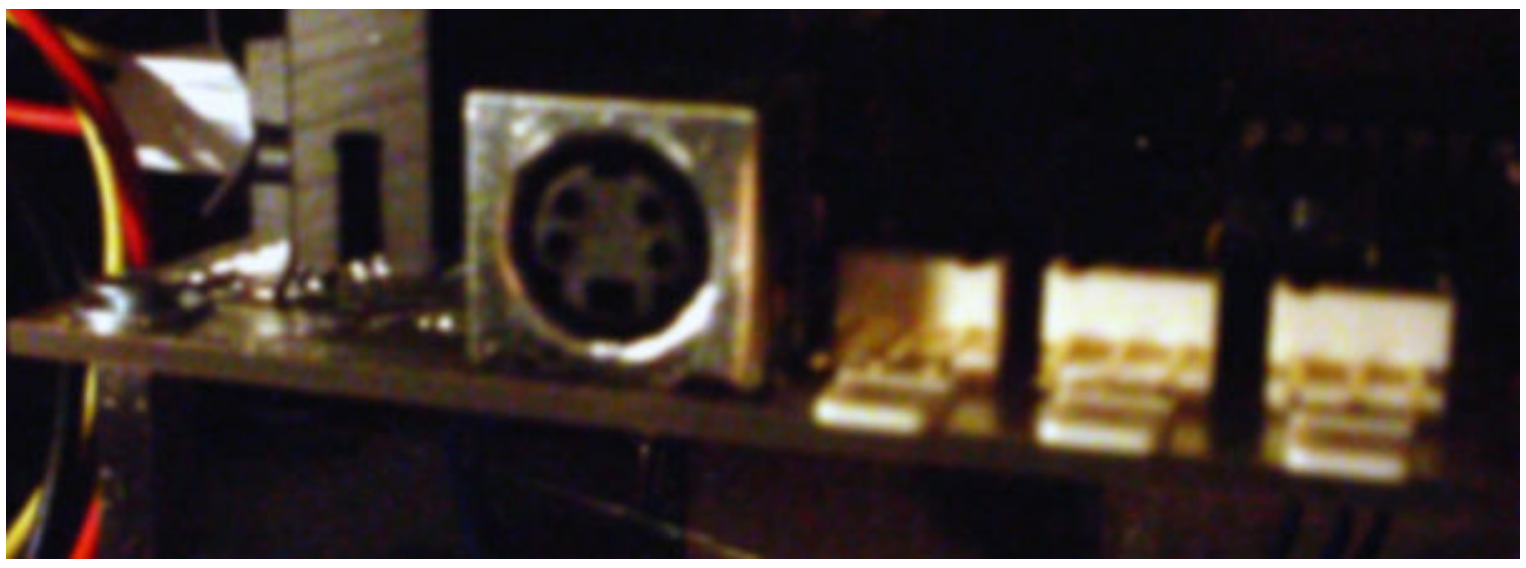

Abbildung 42 - Schrittmotoranschlussbuchsen

## 6.3 Microcontroller im Umgebungsscanner

Der Microcontroller muss folgende Aufgaben bearbeiten:

- Steuerung der Sensorplattform

  Der Microcontroller berechnet die für die Positionierung notwendigen Schritte (siehe Kapitel 3.2.2) und führt die zur Ansteuerung notwendigen Steuersequenzen aus.

  Um die Positionierungszeiten zu minimieren werden die Beschleunigungs- bzw. Bremsrampen berechnet und diese in der Steuersequenz berücksichtigt (siehe Kapitel 6.6.4).

- Verarbeitung der Sensordaten

  Eine weitere Aufgabe der Microcontrollersoftware ist die Verarbeitung der Infrarot-Sensor-Daten (siehe Kapitel 6.6.3). Dazu wird der Forward-Pass für ein Multilayer-Perceptron ausgeführt.

  Das aktivieren und deaktivieren der IR-Sensoren ist in diesem Zusammenhang eine weitere Funktion, die durch die Ansteuerung des 74HCT4514 realisiert wurde (siehe Kapitel 6.4.3).

- Kommunikation mit dem Hauptrechner (Host)

  Zum Befehlsempfang und Weitergabe der durch die Sensoren gewonnen Umgebungsdaten werden Kommunikationsschnittstellen notwendig. Um diese nutzen zu können müssen die entsprechenden Kommunikationsprotokolle implementiert werden.

### 6.3.1 Auswahlkriterien

Der im Umgebungsscanner eingesetzte Microcontroller muss folgende Kriterien erfüllen:

- Schnittstelle für den I²C-Bus
- günstiger Preis
- integrierte AD-Wandler
- Systemresourcen für Erweiterungen
- Energiesparmodul (Sleep-Modus)

Die optimalste Wahl für den Umgebungsscanner stellt der Atmel ATmega8535 dar.

### 6.3.2 Eigenschaften des ATmega8535

Der ATmega8535 ist pinkompatibel mit dem AT90S8535. Die Megaserie bietet zur normalen Ausführung einen erweiterten Befehlsfunktionsumfang und zusätzliche Busschnittstellen (z.B. $I^2C$).

In folgender Tabelle sind die Eigenschaften im Überblick dargestellt:

| Funktion | Wert | Funktion | Wert |
|---|---|---|---|
| Flash (kByte) | 8 | I/O-Pins | 32 |
| EEPROM (Byte) | 512 | $f_{max}$ (MHz) | 16 |
| SRAM (Byte) | 512 | PWM[4]-Kanäle | 4 |
| 8-Bit-Timer | 2 | 16-Bit-Timer | 1 |
| 10-Bit-AD-Kanäle | 8 | SPI[5] | 1 |
| UART | 1 | $I^2C$ | 1 |
| Betriebsspannung | 4,5-5,5V | Preis | ca. 4-5€ |

Tabelle 7 - Funktionsumfang des ATmega8535

### 6.3.3 Einsatz im Scanner

Im Umgebungserfassungssystem (Endversion) ist der ATmega8535 der Host-controller. Er steuert und verwaltet verschiede Subcontroller und Schnittstellen. In folgender Grafik ist die Einbindung des ATmega8535 in die Modulhardware dargestellt:

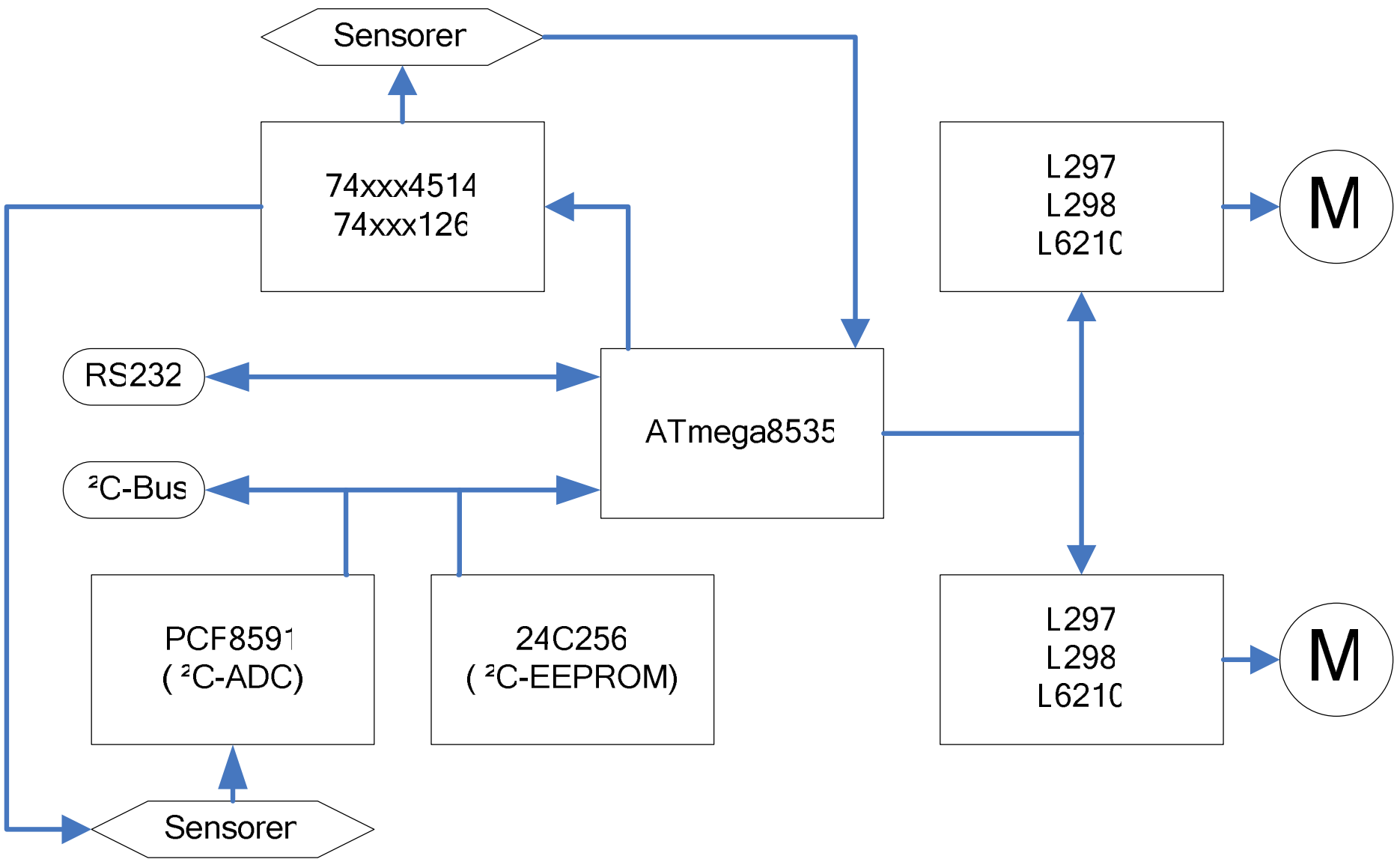

Abbildung 43 - Einbindung des ATmega8535 in Modulhardware

[4] Pulse-Weiten-Modulation (Begriff siehe Glossar)

[5] Serial-Peripheral-Interface

## 6.4 Hardwareentwicklungsstufen

Der Umgebungsscanner ist für den Einsatz in mobilen Robotern konzipiert. Damit werden an die Hardware besondere Ansprüche gestellt. Dazu zählen beispielsweise ein geringer Energieverbrauch und ein Stand-By-Modus, die Unempfindlichkeit gegen Vibrationen, eine universelle Einsetzbarkeit in verschiedenen mobilen Robotern, die Leistungssteuerung der elektromechanischen Komponenten und eine Fremdsignalresistenz der Sensoren.

Die meisten der gerade genannten Forderungen betreffen den Stromverbrauch, eine geringe Wärmeentwicklung, sowie eine hohe Genauigkeit der Schaltungen. Um die Vorraussetzungen zu erfüllen, ist der Einsatz entsprechender ICs und Schaltungen notwendig. Auf diese wird in den folgenden Teilkapiteln eingegangen.

Der geforderte Grundaufbau wird in der folgenden Abbildung dargestellt:

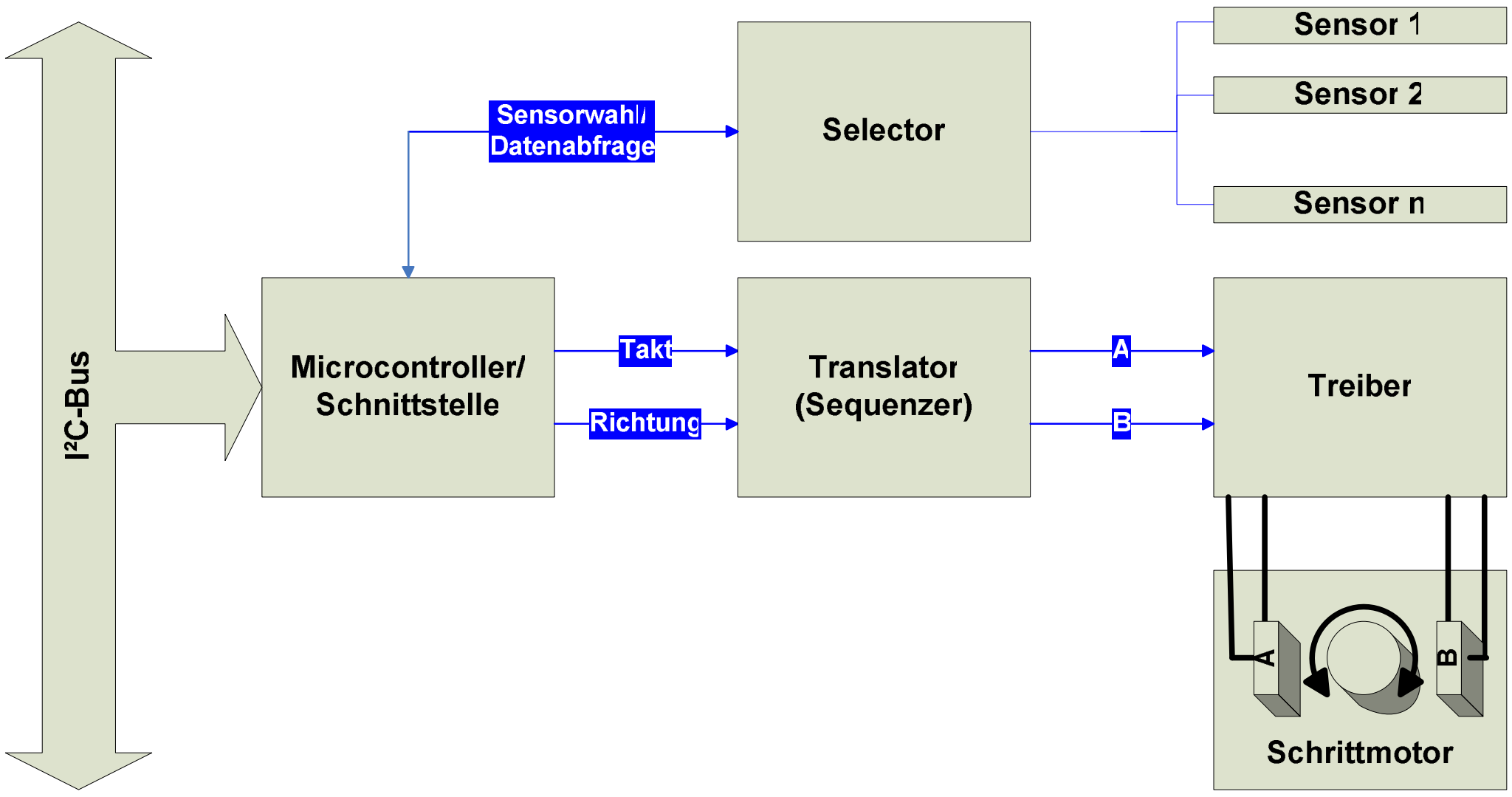

Abbildung 44 - Grundaufbau des Umgebungsscanners

### 6.4.1 Labormuster auf Lochrasterplatine

Das erste Labormuster wurde auf einer Lochrasterplatine erstellt. Es diente zum Test des Microcontrollerprogramms, der Ansteuerprozeduren des Schrittmotors und dem verifizieren der Umgebungserfassungsprozedur. Auf die Softwareanteile wird im Kapitel 6.6 näher eingegangen.

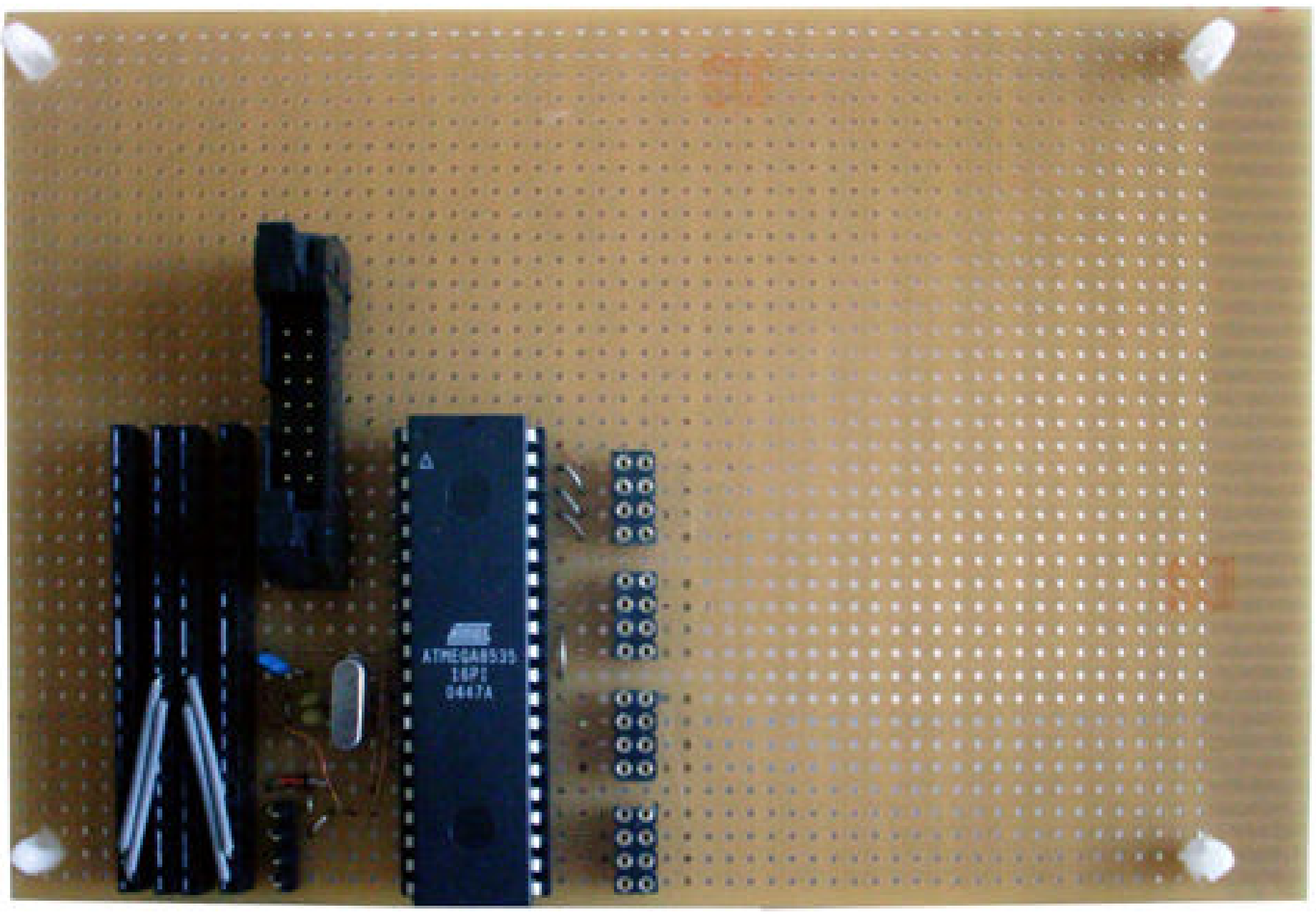

Abbildung 45 - 1. Prototyp zum Test der Steuer- und Kommunikationsprozeduren

Auf der Lochrasterplatine wurden die Schnittstellen, der Microcontroller und die Infrarotsensoranschlüsse aufgelötet.

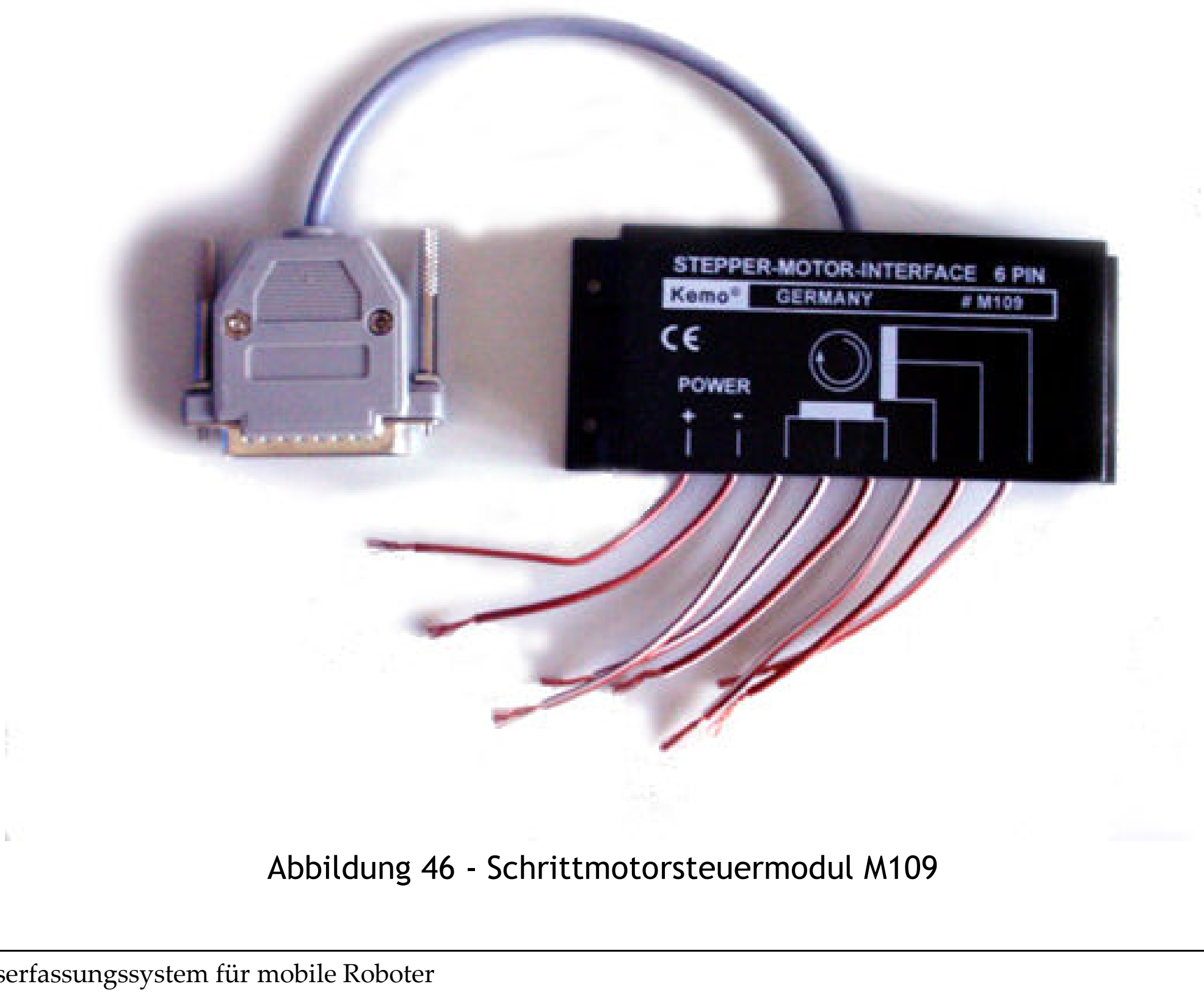

Abbildung 46 - Schrittmotorsteuermodul M109

Die Motorsteuerung wurde mit Hilfe des Schrittmotor-Interface „M109“ der Firma Kemo Electronic realisiert und am LPT-Port eines PCs getestet. Es beinhaltet den Leistungsteil der Steuerung. Im Anschluss wurde das Modul an den Testaufbau des Steuer-Microcontrollers angeschlossen.

Es erfolgten erste Tests der Schrittmotorsteuerprozedur, welche im Kapitel 6.6.4 detailliert beschrieben wird. Eine Auswertung der Sensordaten erfolgte in dieser Prototypenphase im Hostrechner. Die Kommunikation erfolgte ausschließlich über die RS232-Schnittstelle.

Eine Beta-Version der Hostsoftware übernahm dabei sowohl die Steuerung des Sensorkopfes, als auch die Darstellung der Messdaten.

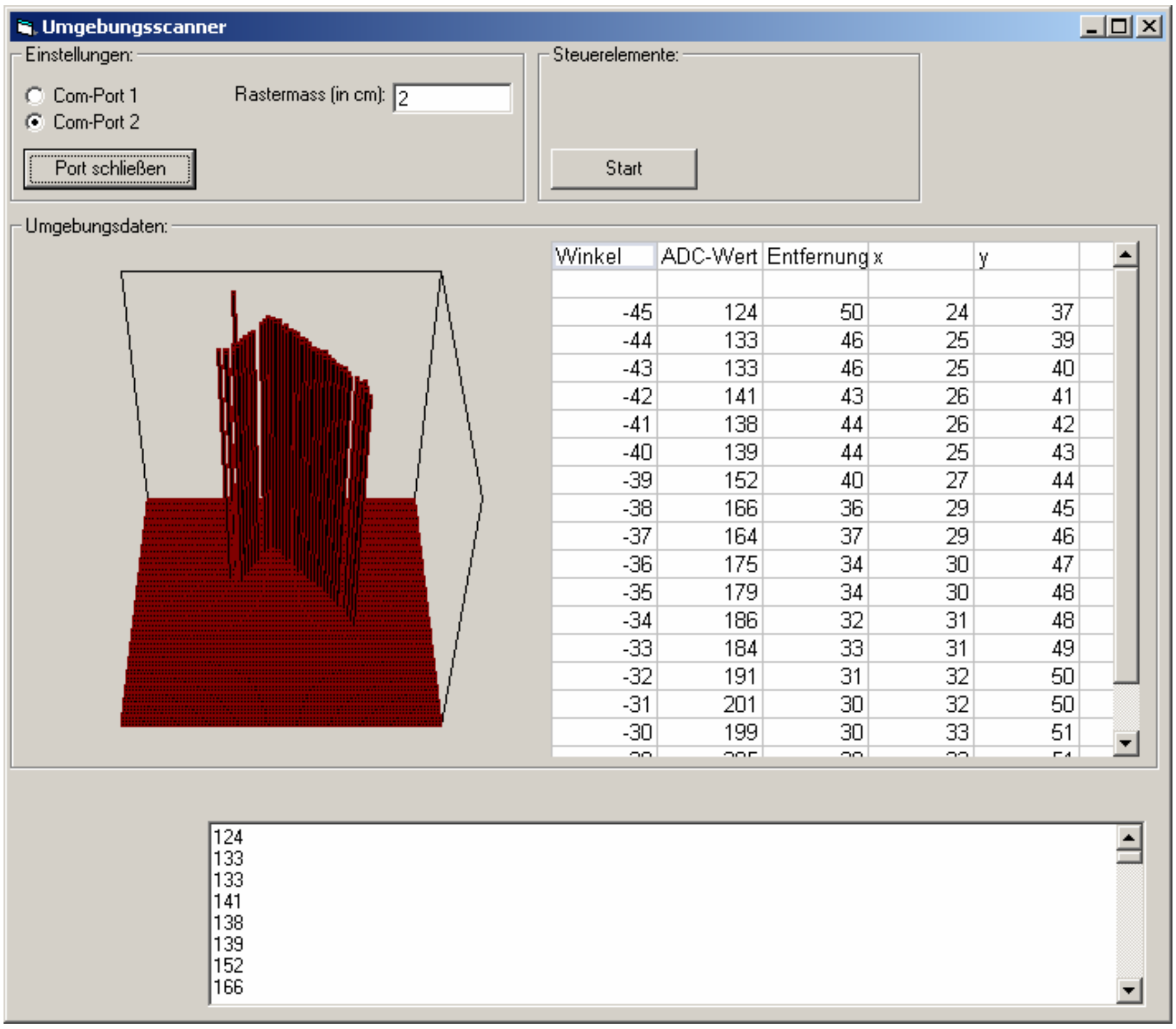

Abbildung 47 - Betaversion der Steuer- und Testsoftware des Umgebungserfassungssystems

### 6.4.2 Gefräste Testplatine

Nach der erfolgreichen Erprobung der Schaltung auf der Lochrasterplatine, erfolgte der Aufbau eines gefrästen Testmusters.

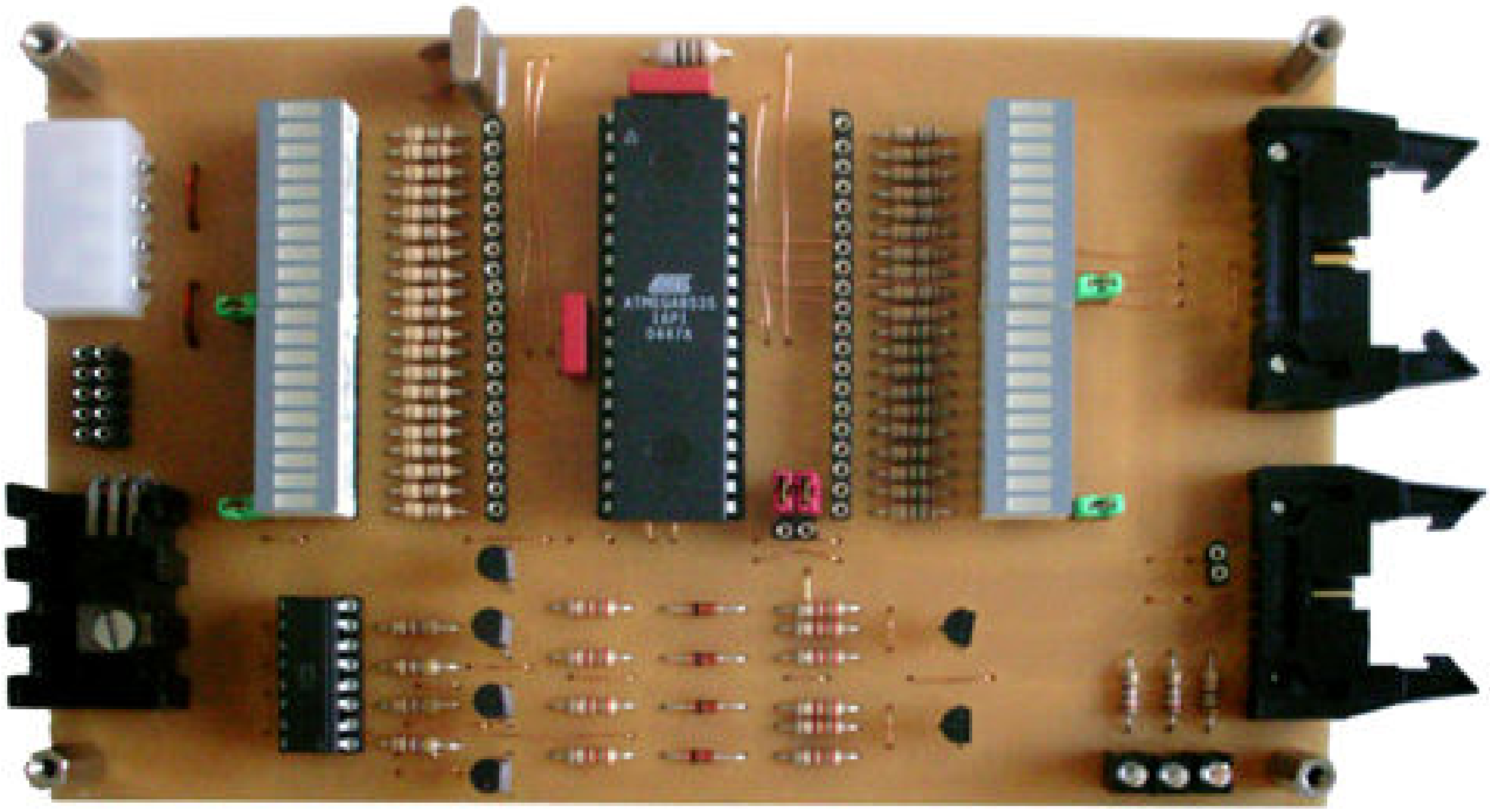

Abbildung 48 - 2. Prototyp mit I²C-Bus-Schnittstelle

Die Experimentierplatine wurde zur Programmierung und Kontrolle des I²C-Datenflusses und dem Test der erweiterten Microcontrollerprozeduren erstellt.

Zur Leistungssteuerung des Schrittmotors zum Antrieb des Sensorkopfes wurde in dieser Phase wieder der „M109“ verwendet.

### 6.4.3 Endversion des Moduls

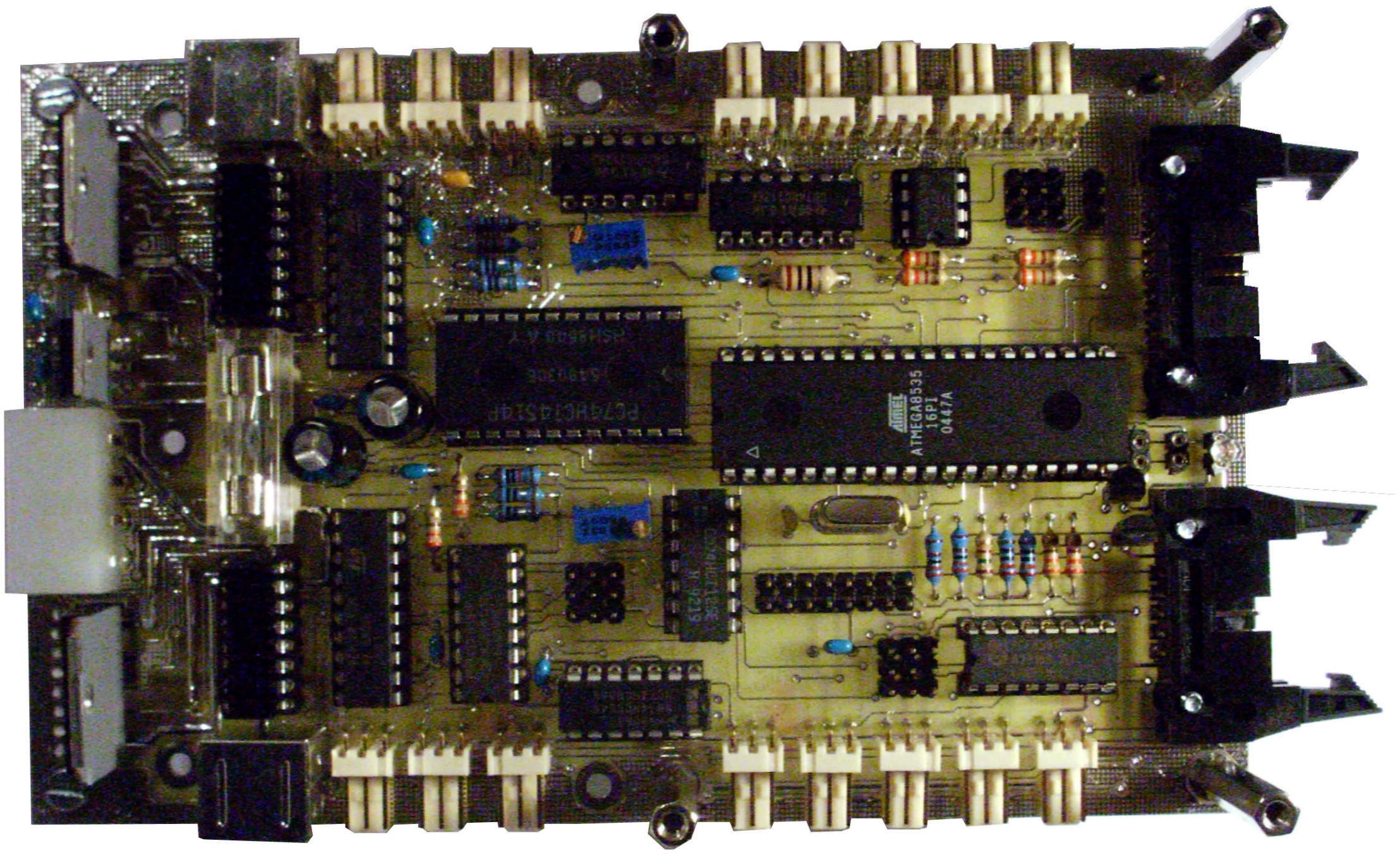

Abbildung 49 - 3. Prototyp (Endversion) der Platine des Umgebungserfassungssystems

In den bisherigen Testaufbauten wurde das „M109"-Schrittmotor-Interface als Leistungsstufe für die Schrittmotoren genutzt. Allerdings ist der Leistungsumfang des Moduls, im Bezug auf die Überwachung des Schrittmotors und die Stromregelung sehr begrenzt.

Das Interface sollte zur Steigerung des Leistungsumfangs der endgültigen Schaltung durch leistungsfähigere Elemente ersetzt werden. Diese werden im Kapitel 6.5 erläutert.

Die Endversion (in Abbildung 49 ohne Kühlkörper abgebildet) verfügt über ein verbessertes Energiemanagement sowohl bei den IC's, als auch bei den Sensoren und Motoren.

## 6.5 Leistungsumfang der Endversion

Im folgenden Kapitel (6.5) wird die Modulhardware des Umgebungserfassungssystems detailliert beschrieben. Der Abschnitt 6.6 stellt die im Microcontroller integrierte Modulsoftware dar.

Eine Bedienungsanleitung und die Übersicht der Befehle, welche im Umgebungserfassungssystem implementiert sind, finden sich auf dem beiliegendem Datenträger und im Quellcode des Microcontrollerprogramms.

### 6.5.1 Anbindung und Energiemanagement der Sensoren

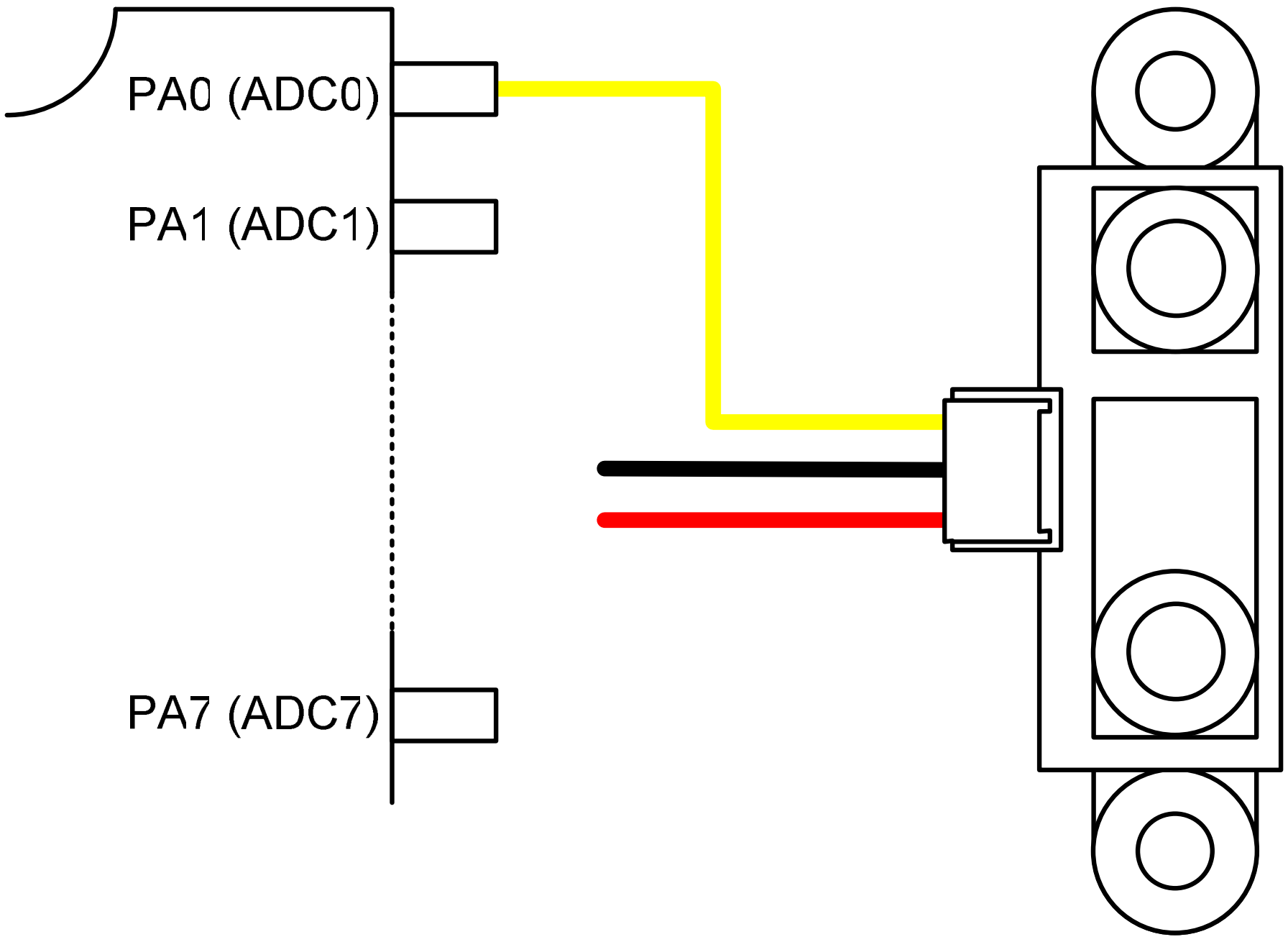

Abbildung 50 - Anschluss der Infrarotentfernungsmesssensoren an den Microcontroller

Die vom Sensor gemessene Entfernung wird als analoge Spannung ausgegeben. Durch diese Vorverarbeitung der Messdaten, mit Hilfe der Sensorelektronik, (wie in im Kapitel 2.1.1 beschrieben) gestaltet sich die Verbindung zum Microcontroller sehr einfach. Es wird der $V_o$-Ausgang des Sensors (analoger Signalausgang) mit einem ADC-Pin des ATmega8585 oder des $I^2C$-ADC-Bausteines verbunden. In Abbildung 50 ist der Sachverhalt dargestellt.

Der Anschluss ist sowohl für die fest am Roboter montierten, als auch für die am Umgebungsscanner angebrachten Sensoren identisch. Es können ebenfalls verschiede IR-Sensoren mit analogem Ausgang eingesetzt werden, da sie untereinander anschlusskompatibel sind.

Zur Energieersparnis im Gesamtsystem und zur Vermeidung von Überlagerungen der IR-Strahlung wurde ein Energiemanagementsystem im Modul integriert. Je nach Betriebsart, Umgebungserfassung oder Hinderniserfassung während der Fahrt, können die nicht benötigten Sensoren deaktiviert werden.

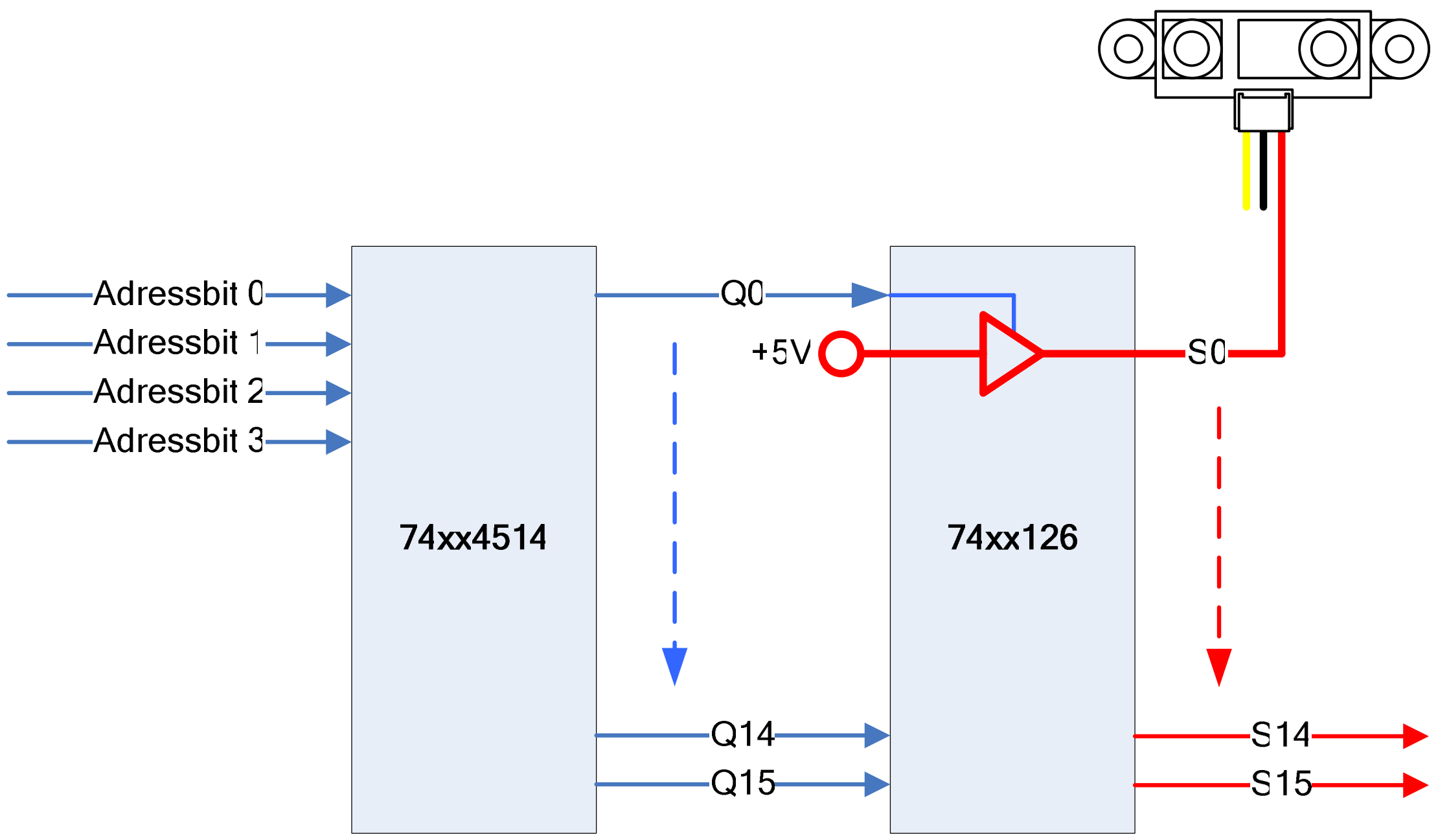

Abbildung 51 - Blockschaltung des Energiemanagementsystems der IR-Sensoren

Wie das Blockschaltbild (Abbildung 51) zeigt, sind zur Sensorkontrolle lediglich zwei IC's notwendig.

Der 74xx4514 ist ein 4-zu-16-Leitungsdecoder. Es werden 4-Adress-Bits vom Microcontroller an den Adresseingängen des 74xx4514 angelegt. Damit liegt am adressierten Ausgang ein Hi-Pegel an.

Mit dem Hi-Pegel des 74xx4514 wird die entsprechende Leitung des 74xx126 durchgeschaltet. Der 74xx126 ist ein Vierfach-Puffer und Leitungstreiber. Im Energiemanagementsystem wird er als Leitungstreiber eingesetzt. Wird die Leitung geschalten, liegen am entsprechenden Ausgang die 5V des Einganges an, ansonsten das Massepotential. Beim aktivieren des jeweiligen Leitungstreibers wird der daran angeschlossene IR-Sensor mit seiner Betriebsspannung versorgt. Er wird somit aktiviert und die Entfernung kann am entsprechenden ADC-Pin abgefragt werden. Wird der jeweilige IR-Sensor nicht adressiert, befindet er sich im inaktiven Zustand.

Sobald die Umgebungsabtastung erfolgt, ist nur der IR-Sensor des Sensorkopfes aktiviert. Alle anderen werden deaktiviert.

Während der Fahrt werden die fest am Roboter montierten Sensoren im „Round-Robin“-Verfahren abgefragt. Dadurch wird verhindert, dass die Messung des gerade aktiven IR-Sensors durch Fremdinfrarotstrahlung der anderen IR-Sensoren beeinflusst wird.

### 6.5.2 Bussysteme und Schnittstellen

CSP[6]-Port:

Da die Funktionen der Kontroll-, Steuer- und Programmierschnittstelle im laufenden Betrieb nicht benötigt werden und darüber hinaus auch Platz auf der Platine kosten, wurden sie durch einen CSP-Port ersetzt.

Die Programmierschnittstelle dient zum Übertragen des Programms in den Microcontroller. Dabei können alle ISP[7]-fähigen Atmel-Microcontroller mit Hilfe einer Programmiersoftware, wie beispielsweise Ponyprog programmiert werden. Ein teures Starterkit, wie zum Beispiel das STK500 von Atmel ist dabei nicht nötig.

In den CSP-Port ist ebenfalls eine serielle Schnittstelle im RS-232-Standard integriert, die sowohl zur Datenkommunikation mit dem Umgebungserfassungssystem, als auch als Testinterface dienen kann. Sollten die Testfunktionen genutzt werden, müssen diese in das Controllerprogramm integriert werden. Dies ist jedoch nur im Entwicklungsstadium des Programmes von Nutzen.

Die Programme und die CSP-Port-Beschreibung befinden sich auf dem beiligendem Datenträger.

I²C-Bus (Interner Integrierter Schaltungsbus):

*Zwei Leitungen, 'serial data' (SDA) und 'serial clock' (SCL), übertragen Informationen zwischen den Bausteinen die am Bus angeschlossen sind. Jeder Baustein ist über eine einzigartige Adresse ansprechbar (ob es ein Microcontroller, LCD Treiber, Speicher oder Tastaturinterface ist) und kann entweder als Sender oder Empfänger operieren, abhängig von der Funktion des Bausteines.* (Übersetzung Hesselbach nach [Philips 2002]) Bei dem I²C-Bus handelt es sich um einen biderektionalen Bus.

Für diesen von Philips Semiconductors  entwickelten Bus mit entsprechendem I²C-Protokoll sind verschiedene Bausteine verfügbar. Das Spektrum reicht von Microcontrollern über Port- und Speichererweiterungen bis hin zu Uhrenbausteinen.

Jedem Teilnehmer ist eine 7-Bit Adresse zugeordnet, gefolgt von einem Lese/Schreib-Byte. Damit können Daten an I²C-fähige ICs eindeutig adressiert werden.

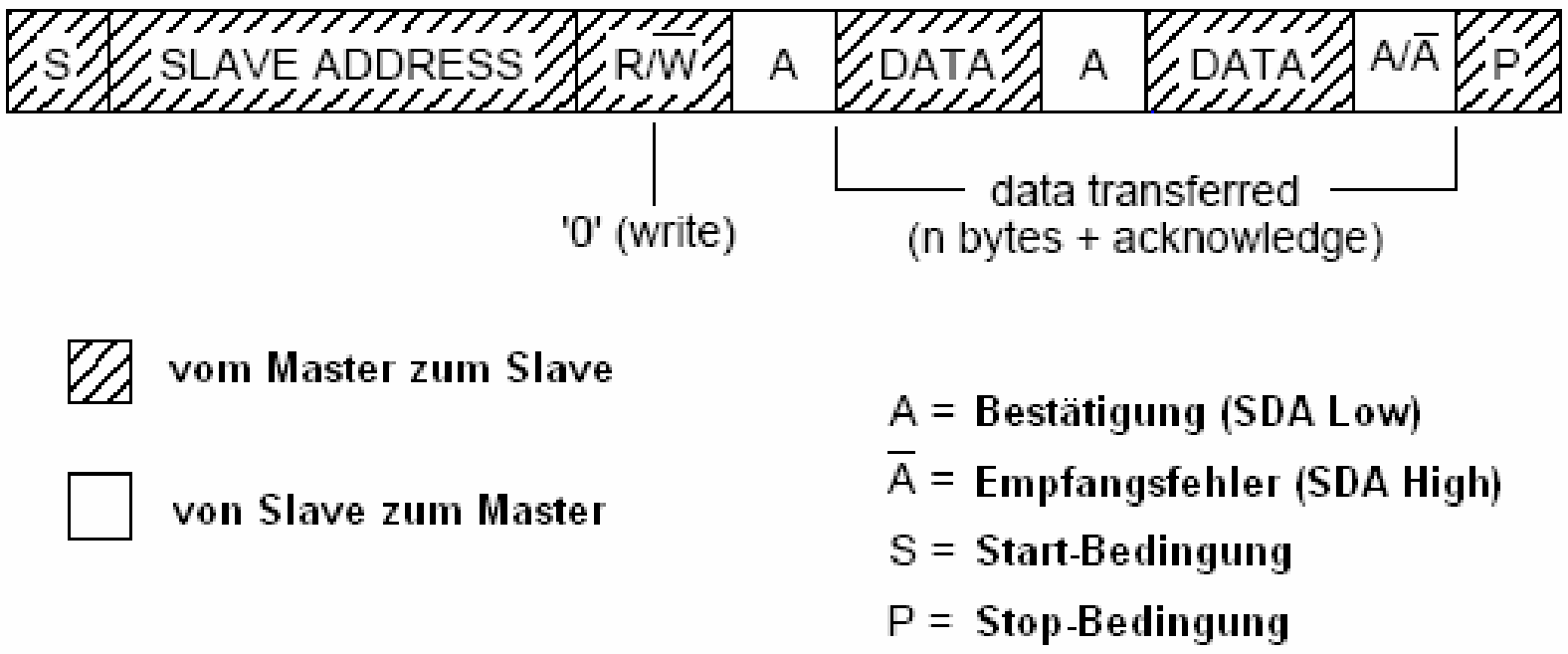

Abbildung 52 - I²C-Frame

[6] Com-Port, Serielles-Test-Interface und Programmierport

[7] In-System-Programmer/In-System-Programmable

Dabei sind zwei Gruppen von acht Adressen (0000XXX und 1111XXX) für besondere Zwecke reserviert. Eine genaue Beschreibung findet sich in Tabelle 8.

| Slave-Adresse | R/W-Bit | Beschreibung |
|---|---|---|
| 0000 000 | 0 | generelle Ruf-Adresse |
| 0000 000 | 1 | Start-Byte |
| 0000 001 | X | CBUS-Adresse |
| 0000 010 | X | reserviert für verschiedene Bus-Formate |
| 0000 011 | X | reserviert für zukünftige Erweiterungen |
| 0000 1XX | X | Hs-Mode Master-Code |
| 1111 0XX | X | 10-Bit-Slave-Adressierung |
| 1111 1XX | X | reserviert für zukünftige Erweiterungen |

Tabelle 8 - reservierte I²C-Adressen

Nach dem Adressteil wird der Acknowledge-Clock-Impuls vom Slave empfangen. Im Anschluss sendet der Master im Wechsel ein Daten-Byte und wartet auf den Acknowledge-Clock-Impuls vom Slave. Das gesamte Frame aus Adress- und Datenteil wird von einem Start- und einer Stop-Bedingung eingeschlossen.

Bei einem Single-Master-System ist ein Baustein am Bus der Master, welcher verschiedene Slaves kontrolliert. Der I²C-Standard lässt aber auch mehrere Master (Multi-Master-System) an einem I²C-Bus zu.

Ein Baustein kann Daten senden, empfangen (wie zum Beispiel eine Flüssigkeitskristallanzeige (LCD)) oder beides (wie eine Real Time Clock (RTC)). Diesen bezeichnet man dann als Transceiver.

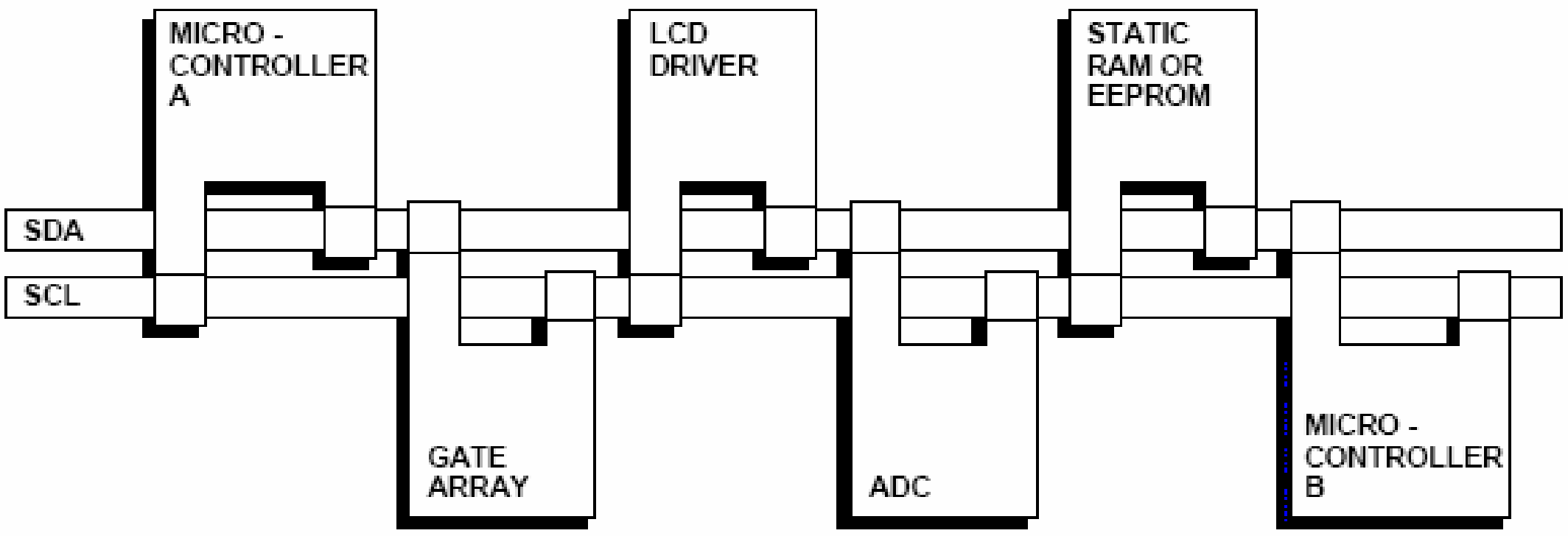

Abbildung 53 - I²C-ICs am I²C-Bus [Philips 2002]

Der Master gibt die Datenflussrichtung und die Übertragungsgeschwindigkeit mit der SCL-Leitung vor. Beide Übertragungsleitungen werden mit PullUp-Widerständen versehen, um diese auf ein Hi-Pegel zu ziehen. Die mit Open-Drain-Schaltungen angeschlossenen Bus-Teilnehmer ziehen die Leitungen im aktiven Zustand gegen Masse (Low-Pegel).

### 6.5.3 Leistungstreiberschaltung der Schrittmotoren

Wie im Kapitel 6.4.3 angerissen, wurde das „M109“-Schrittmotorinterface durch eine Schaltung ersetzt.

Um den nötigen Platz im Platinenlayout zu minimieren, wurden Integrierte Schaltkreise (IC’s) verwendet. Zur Steuersignalgenerierung für die Motorphasen, Leistungssteuerung und Regelung der Schrittmotoren werden drei IC’s verwendet.

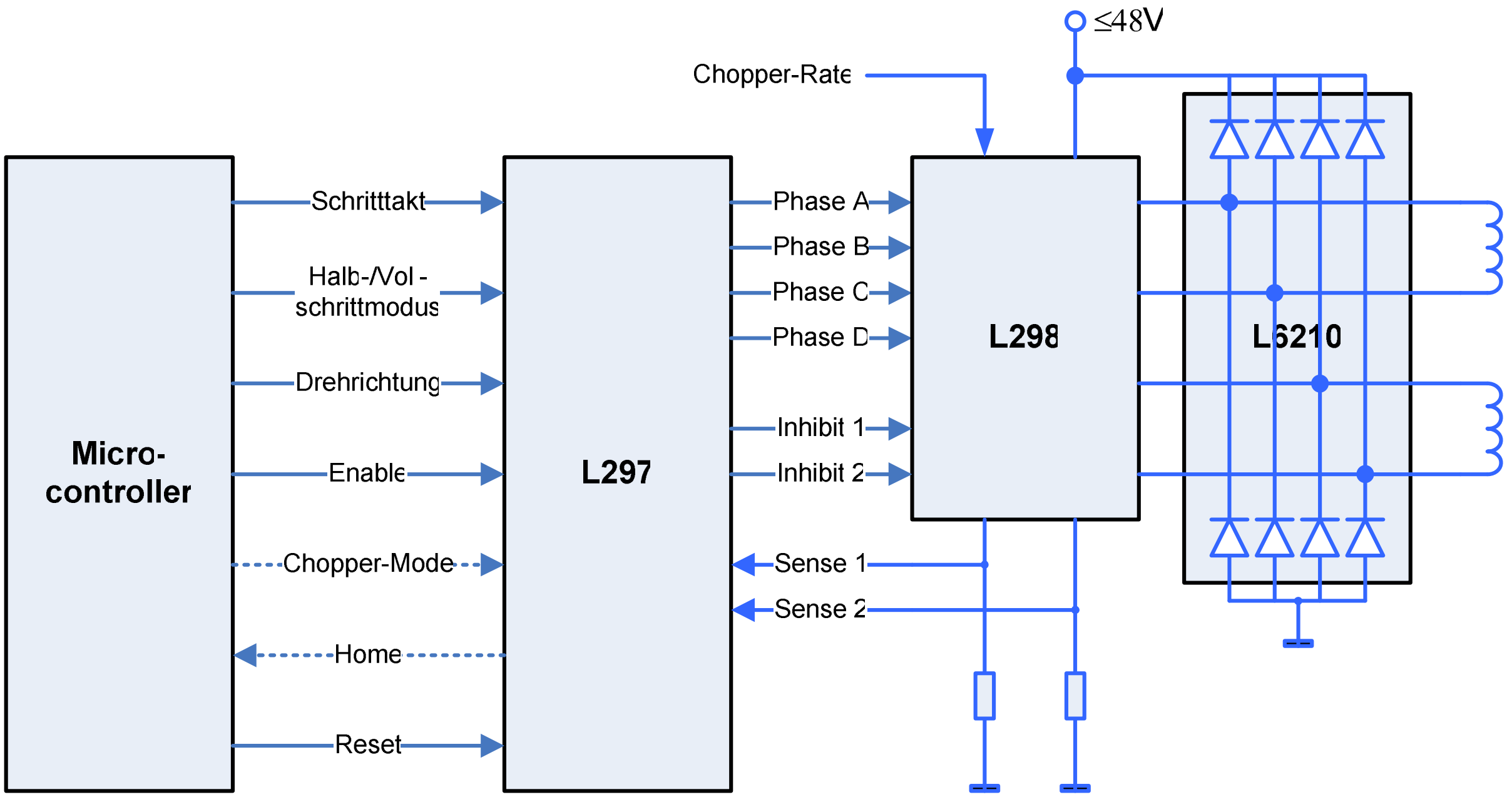

Abbildung 54 - Blockschaltbild der Schrittmotorsteuerung und des Leistungstreibers

Der L297 übernimmt in der Schaltung die Funktion des Sequenzers im Translator und des Choppers (Zerhacker).

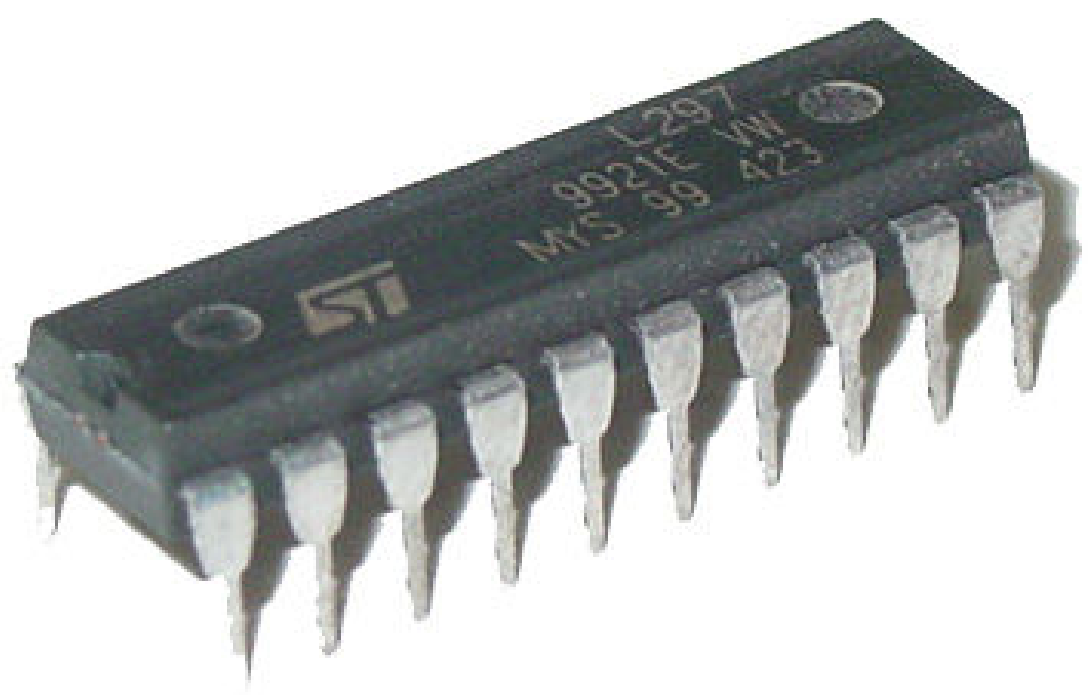

Abbildung 55 - L297 Schrittmotor-Controller

Die Phasenpegel werden aus den Schrittimpulsen und dem Richtungssignal des Microcontrollers generiert. Durch das Schritttaktsignal wird ein Schritt ausgeführt. Die Frequenz der Schrittfolge wird durch den Microcontroller oder –prozessor vorgegeben. Abhängig vom Pegel des Schrittmoduseinganges (in Abbildung 54 das Signal

„Halb-/Vollschrittmodus") wird an den Phasenausgängen (Phase A bis D in Abbildung 54) die Steuersequenz für den Vollschrittmodus (siehe Anhang) oder für den Halbschrittmodus (siehe - Ansteuerungssequenz eines bipolaren Schrittmotors (Halbschrittbetrieb) im Anhang) generiert. Die Drehrichtung des Schrittmotors wird dabei über das Drehrichtungssignal bestimmt.

Damit die Microcontrollersignale vom L297 verarbeitet werden, muss er mit einem Hi-Pegel am Enable-Eingang aktiviert werden. Wird der Schrittmotor nicht benötigt, werden der L297 und damit der Schrittmotor deaktiviert. Um die Kalibrierung der Motorposition nicht zu verändern, wird vor der Deaktivierung ein Low-Pegel am Reset-Eingang angelegt, wodurch der Schrittmotor in eine Grundstellung gebracht wird. Dort wird er durch die magnetischen Kräfte der Permanentmagnete gehalten.

Die beiden Signale „Chopper-Mode" und „Home" sind für den Schrittmotorbetrieb nicht zwingend notwendig.

Das „Home" Signal dient der Kontrolle und Synchronisation des Schrittmotors. Durchläuft er die Grundstellung kann durch den Open-Collector-Ausgang ein Signal an den Microcontroller übergeben werden, wodurch das Erreichen der Grundstellung signalisiert wird.

Zur Auswahl des internen oder externen Choppers dient der „Chopper-Mode"-Eingang. Das englische Wort „Chopper" bedeutet übersetzt „Zerhacker". Seine Funktion ist die Unterbrechung der Phasenströme mit einer bestimmten Zerhackerfrequenz. Das Abschalten des Phasenstroms beendet den Aufbau des elektromagnetischen Feldes der Statorspulen. Bei Einschalten des Phasenstroms wird das Feld erneut aufgebaut. Mit der Chopperschaltung, welche im L297 integriert ist, wird die Stärke des elektromagnetischen Feldes der Statorspulen und somit die auf den Rotor wirkenden Kräfte reguliert. Somit kann in der Beschleunigungsphase der mechanischen Komponenten ein stärkeres Feld bereitgestellt werden, als es während der Bewegungsphase notwendig ist. In der Bewegungsphase werden, durch die schwächeren elektromagnetischen Felder, weniger hohe Phasenströme benötigt. Dadurch wird Energie eingespart.

Bei der Nutzung des internen Choppers zur Generierung der pulseweitenmodulierten Steuerspannung (siehe „PWM" im Glossar) lässt sich die Chopper-Frequenz mit $f = 1/(0{,}69RC)$ berechnen. R und C geben dabei den Widerstand und die Kapazität des Kondensators des RC-Netzwerkes an. Im Datenblatt des L297 sind Werte von 33nF und 22kΩ empfohlen. Damit ergibt sich eine Frequenz von rund 20kHz.

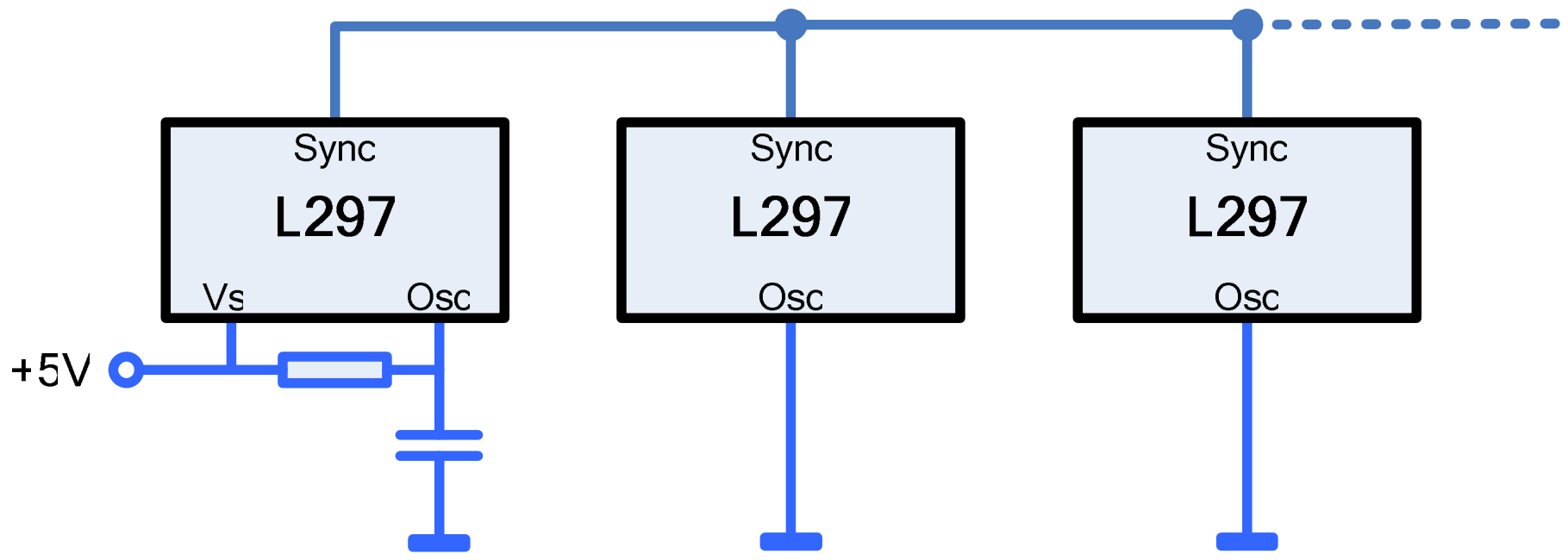

Abbildung 56 - Chopper-Anschaltung und Chopperfrequenzsynchronisation

Sollten mehrere L297-IC's genutzt werden, ist das RC-Netzwerk nur bei einem L297 notwendig. Der OSC-Eingang der anderen IC's wird auf Masse gelegt und die Bausteine über den Sync-Eingang verbunden (siehe Abbildung 56).

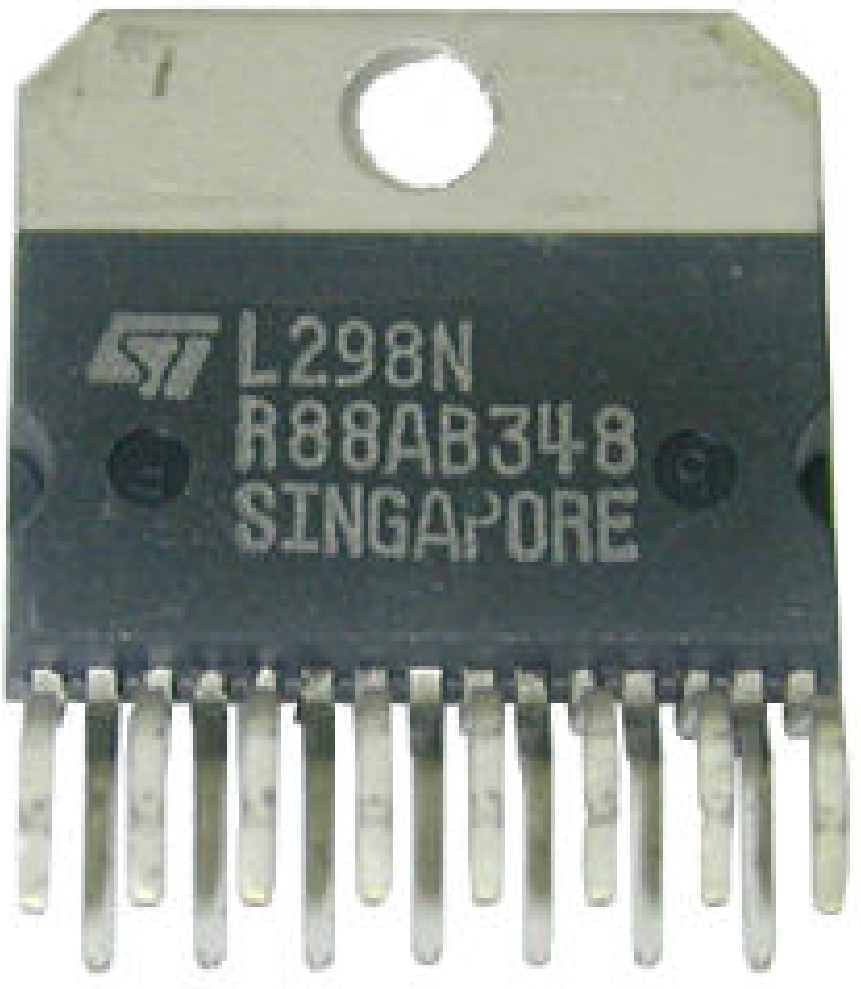

Abbildung 57 - Dual-Vollbrückentreiber

Die eigentliche Stromsteuerung der Statorwicklungen des Schrittmotors übernimmt der L298. Die Leistungsstufe hat in erster Linie die Signale des L297 zu verstärken. Dabei liefert der L298 über die „Sense"-Leitungen einen Abtaststrom an den L297 zurück. Der Controller (L297) erkennt dadurch ob in der Endstufe ein instabiles Verhalten, wie undefinierte Schwingungen oder ein Überlastungsfall auftritt. Sollte einer dieser Fälle eintreten, kann der L297 die Leistungsstufe L298 und damit den Schrittmotor deaktivieren.

Im störungsfreien Betrieb wird durch die Regelelektronik in der Beschleunigungsphase ein hohes Beschleunigungsmoment erzeugt. Dies ist notwendig, um den mit dem Schrittmotor verbundenen Sensorkopf, welcher eine träge Masse darstellt, an zu treiben. Während des Synchronlaufes (Bewegungserhaltung) oder des Stillstandes sorgt der Regelkreis aus L297 und L298 für eine optimale Leistungsversorgung des Motors. Eine Erwärmung wird durch die automatische Reduzierung des Halte- bzw. Steuerstroms vermieden.

Die Steuerströme des L298-Leistungstreibers werden, bevor sie den Motor erreichen, durch den L6210 geleitet. Der L6210 ist eine Dual-Schottky-Dioden-Brücke. Er dient zur Eliminierung von Spannungsspitzen in den Steuerströmen. Diese können durch Rückkopplungseffekte der Schrittmotoren oder durch ein instabiles Verhalten der Endstufe entstehen. Ohne die Schottky-Dioden-Brücken könnte es zu einer Zerstörung der Schrittmotoren kommen.

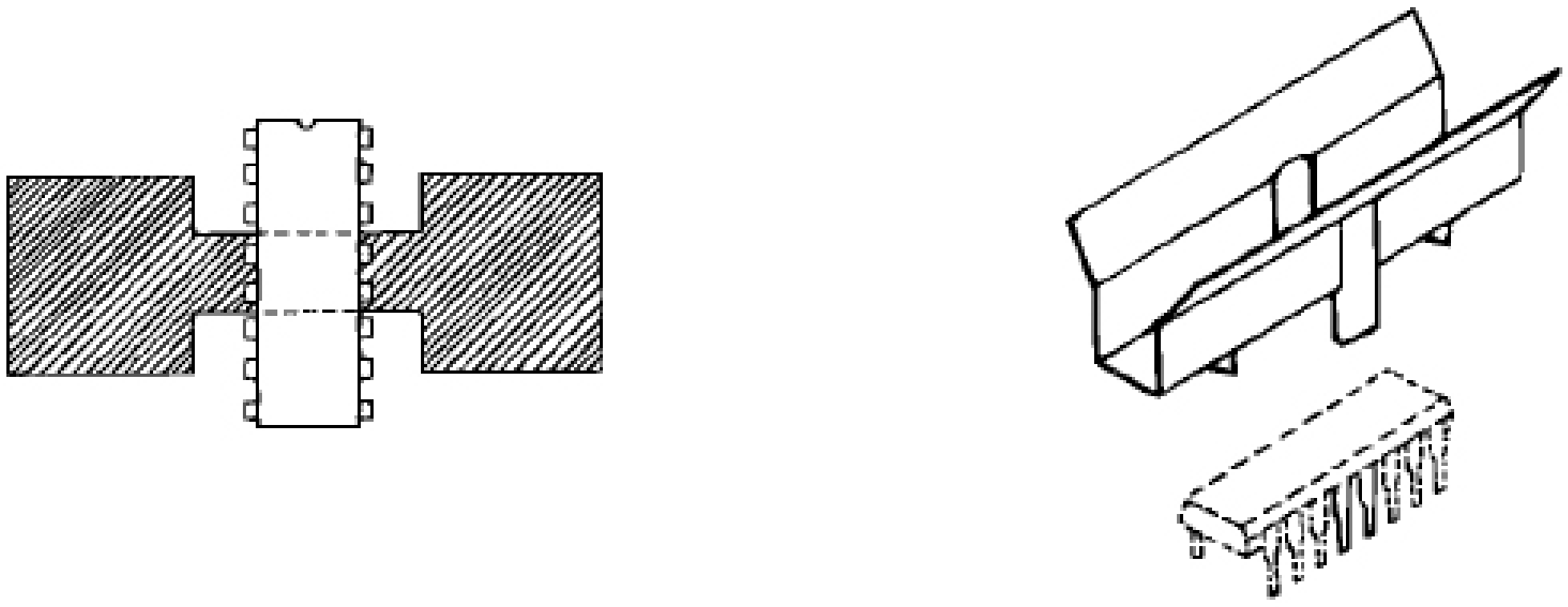

Abbildung 58 - L6210 mit Heatsink oder Kühlkörper

Die Brückenschaltung kann zwar durch separate Schottky-Dioden realisiert werden, jedoch wird dafür mehr Platinenfläche benötigt als bei der Nutzung des L6210. Da die Schottky-Dioden-Brücken in der IC-Bauform räumlich kompakt realisiert sind, muss auf das Gehäuse ein Kühlkörper geklebt oder auf der Platine eine Heatsink realisiert werden. Dies ist für die Wärmabführung aus dem IC unerlässlich. Wird die Wärmeenergie nicht abgeführt, ist eine Zerstörung des L6210 wahrscheinlich.

### 6.5.4 I²C-Bausteine

Neben dem ATmega8535, welcher als I²C-Baustein genutzt wird, kommen weitere Busbausteine zum Einsatz (Abbildung 59). Dazu zählen ein 24C256 und zwei PCF8591-IC's.

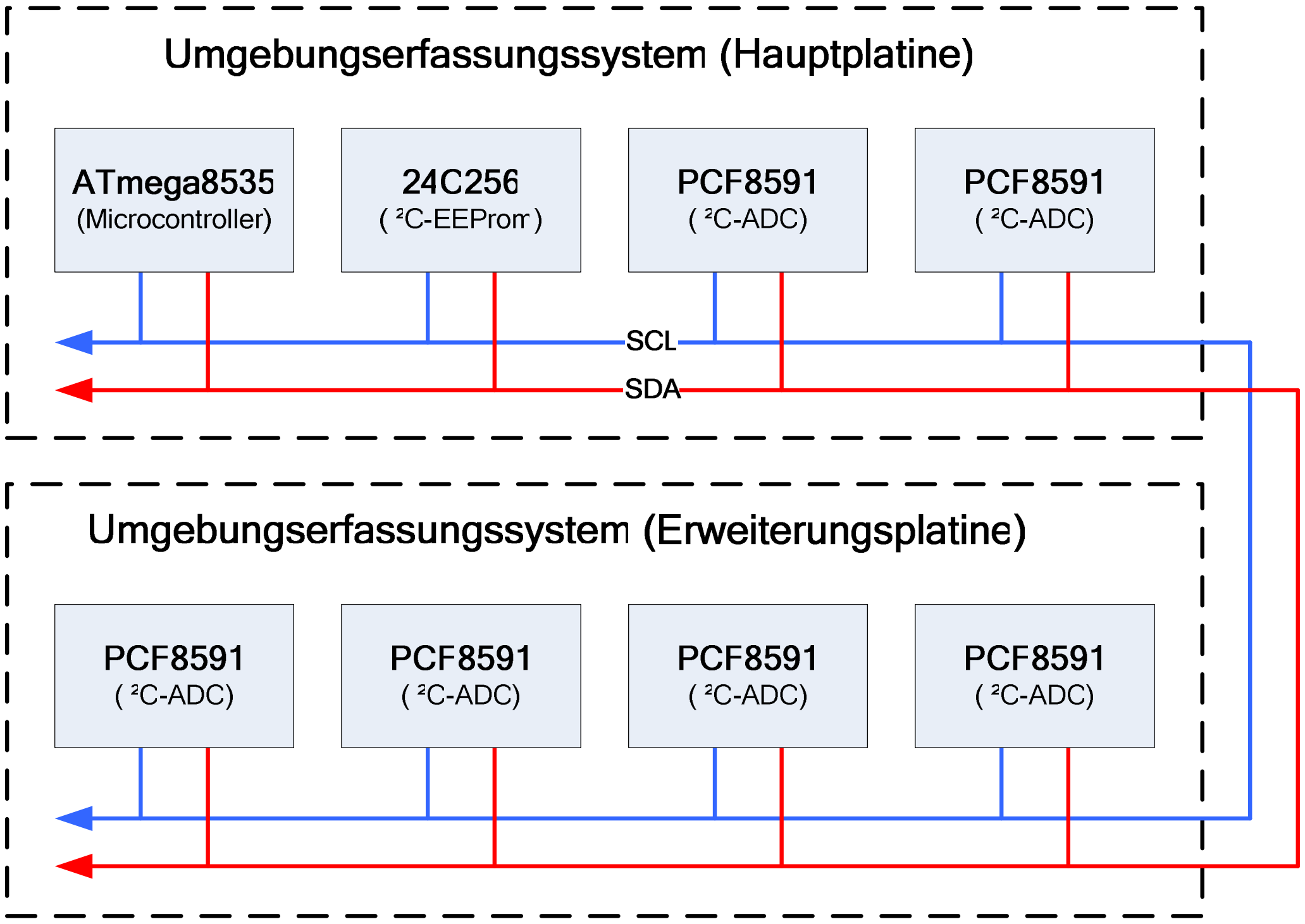

Abbildung 59 - I²C-Busbausteine des Umgebungserfassungssystems

Der 24C256 ist ein 256kByte-I²C-EEPROM. Er zeichnet sich durch eine kleine Bauform (DIP 8) und den geringen Preis (ca. 2€) aus. Es werden 100.000 Schreibzyklen und 40 Jahre Datensicherheit garantiert. Um ein Schreibzugriff über den I²C-Bus zu verhindern, kann der Schreibschutz des EEPROM über einen externen Jumper aktiviert werden. Für eine Page (64Byte) werden maximal 6ms Zykluszeit (Lesen/Schreiben) benötigt.

Aus dem 24C256 werden die Gewichtsmatrizen den KNN's zur Spannungs-/Entfernungswandlung bei der Initialisierung des ATmega8535 gelesen. Bei einer Änderung in der KNN-Matrix muss somit der Microcontroller nicht neu programmiert werden. Die Matritzen können über den I²C-Bus beispielsweise vom PC in den EEPROM geschrieben werden. Weiterhin können die letzte erfasste lokale Karte und verschiedene Initialisierungsvariablen gespeichert werden.

Die PCF8591-IC's sind 8Bit-I²C-Analog-Digital-Wandler. Diese sind notwendig, da der ATmega8535 nur acht ADC-Pins besitzt. Durch die Verwendung von zwei PCF8591 ist der Anschluss von sechzehn Sensoren möglich. Auf dem Expansion-Board stehen weitere vier PCF8591 und damit zusätzliche sechzehn Sensoranschlüsse zur Verfügung.

## 6.6 Modulsoftware

### 6.6.1 Anforderungen

Das Microcontrollerprogramm ist das Betriebssystem des Moduls. Die Aufgabe des Programms ist neben der Steuerung der Hardware die Aufnahme und Verarbeitung der Messdaten der Sensoren.

Dafür gibt es mehrere Gründe. Zum einen wird der Steuercomputer oder der Steuerprozessor durch die Datenverarbeitung im Modul entlastet, wodurch er Ressourcen spart, und zum anderen ist die Datenmenge auf dem $I^2C$-Bus wesentlich geringer. Der Hostcomputer und die Module arbeiten somit ähnlich einem Mehrprozessorsystem. Die Daten werden nicht als Winkel-Enfernungsdatensatz übertragen, sondern als Umgebungsmatrix. Darauf wurde bereits im Kapitel 5.1 eingegangen.

Ein weiterer Aspekt ist die Kontrolle der Hardware. So muss das Programm den Status der angeschlossenen Schrittmotoren und Sensoren an den Steuerrechner melden. Darüber hinaus ist es notwendig bei einer eventuellen Überhitzung der Motoren oder Fehlfunktionen der Sensoren diese zu deaktivieren, um weitere Schäden zu verhindern und an den Steuercomputer eine Fehlermeldung zu senden.

Sollte das Programm in seinem Ablauf gestört sein, beispielsweise durch ungültige Registerzugriffe, so muss es per Watchdog neu gestartet werden (automatischer Reset). Diese Einstellung wird beim Start des Programms initialisiert und kontrolliert in definierten Abständen die Lauffähigkeit des Microcontrollerprogramms.

### 6.6.2 Programmstruktur

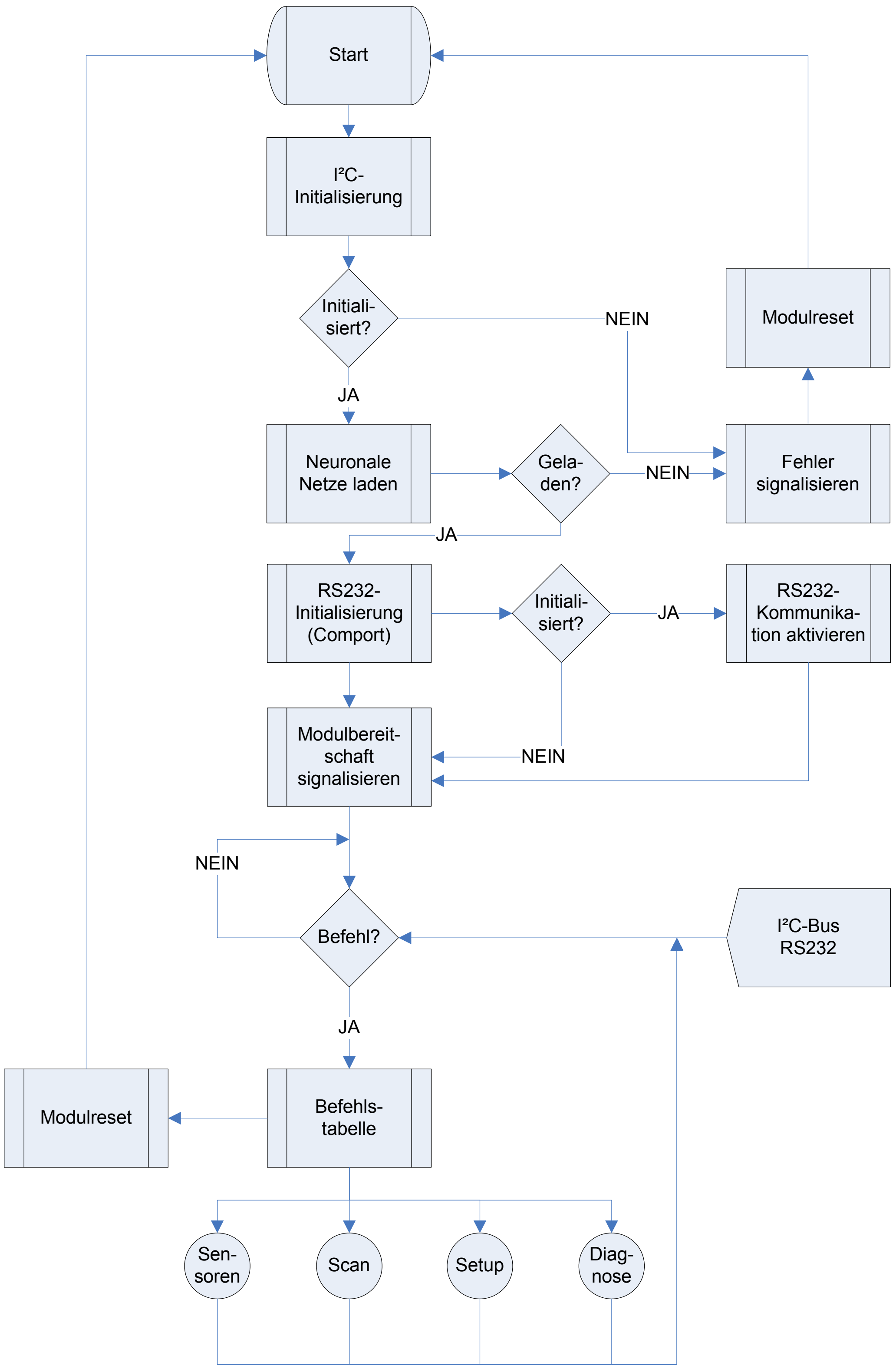

Abbildung 60 - Struktur des Microcontrollerhauptprogramms

Wird das Modul mit Energie versorgt, werden die Kommunikationsschnittstellen initiiert. Sollte dies fehlschlagen, wird ein Modulreset ausgelöst.

Sobald die Kommunikationskanäle etabliert wurden, lädt das Modul das Datenwandlungs-KNN und gegebenenfalls die Initiierungswerte von Variablen aus den I²C-EEPROM.

Ist die Initialisierung abgeschlossen, signalisiert das Modul dem Steuerrechner seine Funktionsbereitschaft. Es durchläuft nun die Endlosbetriebsschleife.

Trifft eine Befehlssequenz ein, wird eine entsprechende Prozedur ausgeführt. Danach kehrt das Modul in die Betriebsschleife zurück und erwartet eine neue Befehlssequenz.

### 6.6.3 Sensordatenauswertung

Die Sensordatenauswertungsprozedur wird sowohl für die fest am Roboter befestigten Infrarotsensoren, wie auch für den Umgebungsscanner (beweglicher Sensor) genutzt. Es sind verschiedene Arten der Datenauswertung denkbar.

Eine sehr einfache Möglichkeit ist die Integration einer Äquivalenztabelle. Dabei handelt es sich um ein Array bestehend aus $2^{ADC-Auflösung}$ Zeilen. Für einen 10bit Analog-Digital-Wandler wird somit ein 1024-zeiliges Array benötigt. Bei einer Messgenauigkeit von einem Zentimeter und einer Reichweite von 150cm, würde man n-Bit mit $2^n > 150$, also n=8 Bit pro Arrayelement, benötigen. Damit ergibt sich ein Speicherbedarf von $Speicherzellengröße * 2^{ADC-Auflösung} = 8 * 2^{10} = 1kByte$. Der Vorteil dieser Methode ist die hohe Wandlungsgeschwindigkeit. Dies wiegt den großen Speicherplatzbedarf nicht auf, da dieser im Microcontroller nur begrenzt zur Verfügung steht. Ein derartiges Array wäre in den ATmega8515 mit seinen 512Byte EEPROM nicht implementierbar. Die Wandlungstabelle müsste in einem externen EEPROM (z.B. I²C-EEPROM) ausgelagert werden. Durch die Busoperationen wäre diese Methode langsamer als bei Speicherung im internen EEPROM.

Die Erstellung einer mathematischen Funktion zur Umrechnung des ADC-Wertes in die dazugehörige Entfernung ist eine zweite Möglichkeit. Dazu ist keine Integration von Arrays und somit kein EEPROM zur Datenspeicherung im Microcontroller notwendig. Der Nachteil dieser Methode ist die Notwendigkeit mathematische Bibliotheken in den Controller einzufügen. Diese belegen sehr viel Platz des Programmspeichers (Flash-Speicher) im Microcontroller.

Beide der eben beschriebenen Verfahren erfordern extrem viel Speicherplatz. Da dieser im Controller sehr begrenzt ist, sind sie nicht zu empfehlen. Eine Methode, welche eine korrekte Umrechnung und einen geringen Speicherplatzbedarf garantiert, wäre die optimale Lösung. Derartige Ansätze werden im Folgenden beschrieben.

Zu diesen Prozeduren gehören die künstlichen, neuronalen Netze. Sie bieten die Möglichkeit die Transferfunktion zwischen ADC-Werten und Entfernung zu approximieren und benötigen dabei nur wenig Speicherplatz. Wegen diesen Vorzügen wurde diese Methode gewählt.

Da sich die Transferfunktion nicht ändert, wurde ein künstliches, neuronales Netz mit eingefrorenem Wissen integriert. Es wurde extern mit den optimierten Daten trainiert, welche aus den Sensormessreihen gewonnen wurden.

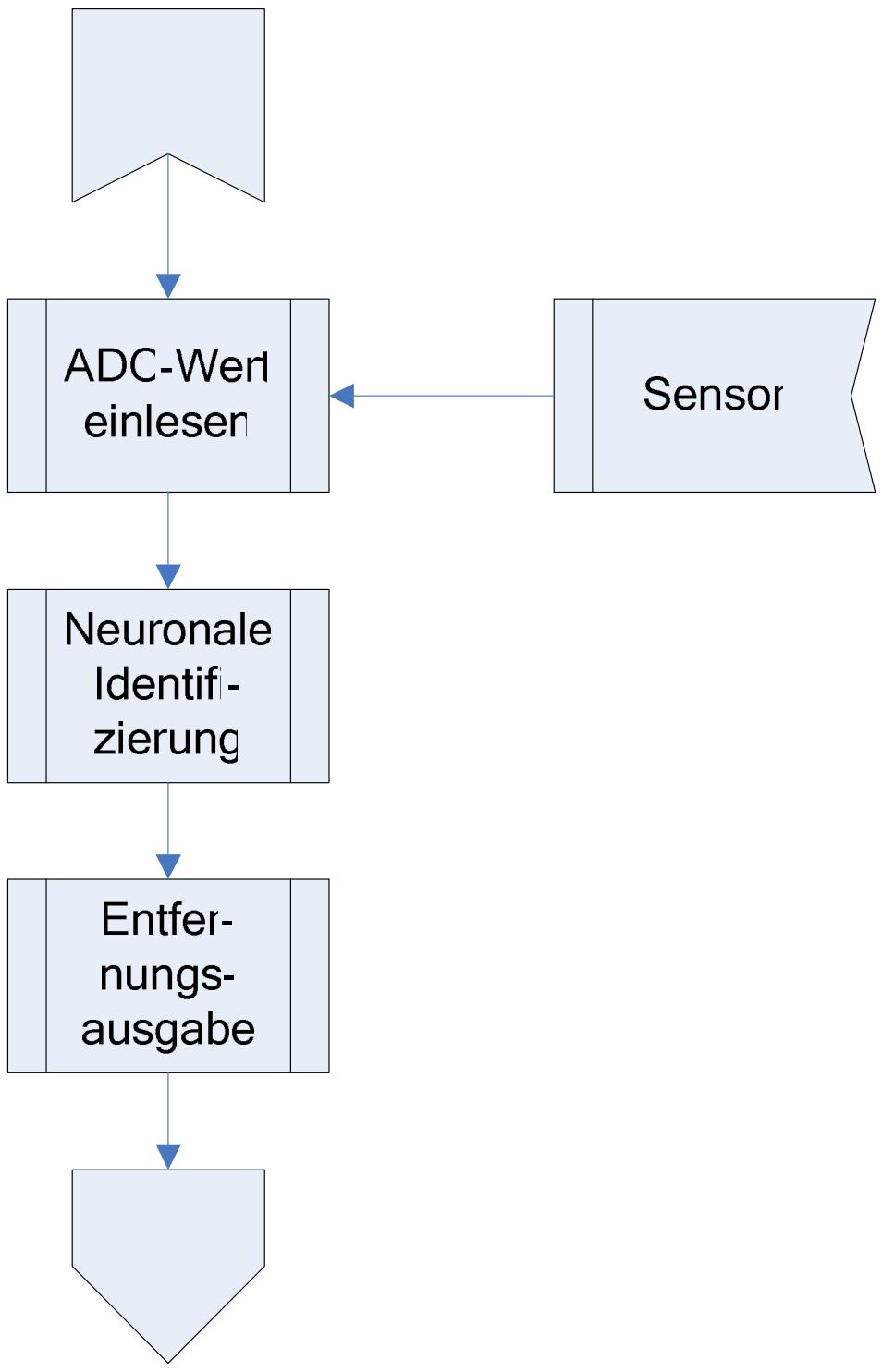

Abbildung 61 - Struktur der Sensorauswertungsprozedur

Die Wandlung der Spannungs- in Entfernungsdaten (im Abbildung 61 mit „Neuronale Identifizierung" bezeichnet) erfolgt durch den Forward-Pass. Das bedeutet die Berechnung des KNN von den Input- zu den Outputschichten. Damit werden die Ausgabe- zu den entsprechenden Eingabedaten berechnet. Die Forward-Pass-Prozedur sieht im Quellcode wie folgt aus:

```
for (L=1;L<=Lmax;L++)
{
    for (n=1;n<=nmax;n++)
    {
        for (c=1;c<=wmax;c++)
        {
            O[L+1,n]=O[L+1,n]+O[L,c]*W[L,n,c];
        }
        if (O[L+1,n]<0.5)
            O[L+1,n]=0;
        else
            O[L+1,n]=1;
    }
}
```

Das im Quellcode mit W[] bezeichnete Array ist die Gewichtsmatrix des neuronalen Netzes. Der Inhalt des Arrays wird, wie in den vorangegangenen Abschnitten beschrieben, bei der Initialisierung aus dem externen EEPROM geladen. Das Array ist wie folgt aufgebaut:

*W[Layernummer, Neuronennummer, Verbindungsnummer]*

Das Array O[] speichert die Outputs der Neuronen in den einzelnen Schichten. Der Inhalt des Output-Array ist rein temporär. Es wird als leeres Array initalisiert. Die Größe wird wie folgt bestimmt:

*O[Anzahl der Layer, Anzahl der Gewichte im größten Layer].*

Die in der Prozedur verwendeten Variablen erklären sich wie folgt:

| Variable/Array | Beschreibung |
|---|---|
| L | Layer |
| Lmax | Anzahl der Layer |
| n | Neuronennummer |
| nmax | Anzahl der Neuronen im größten Layer |
| c | Verbindungsnummer |
| wmax | Anzahl der Gewichte im größten Layer |
| O[] | Speichermatrix der Neuronenausgaben |
| W[] | Gewichtsmatrix des künstlichen neuronalen Netzes |

Tabelle 9 - Variable und Arrays der Microcontroller-Forward-Pass-Prozedur

### 6.6.4 Schrittmotorsteuerung

Die Steuerprozedur der Schrittmotoren hat im Wesentlichen zwei Aufgaben. Zum einen soll eine bestimmte Position angefahren werden können und zum anderen müssen einzelne Winkelschritte während des Scannvorgangs ausgeführt werden.

Das Anfahren einer bestimmten Position ist programmtechnisch mit einer Programmschleife gelöst. Allerdings sind der Schrittmotor, das Getriebe und die Sensorplattform mechanische Komponenten mit Trägheitsmomenten.

Dadurch kann der Schrittmotor nicht mit seiner höchsten Schrittgeschwindigkeit beim Anfahren einer Position angesteuert werden. Durch die Trägheitsmomente der Mechanik würden durch die auf den Rotor wirkenden Kräfte Schritte verloren gehen. Der Rotor könnte sich in den elektromagnetischen Feldern der Statorspulen nicht schnell genug ausrichten. Damit würde man den Vorteil des Schrittmotors, nämlich die Positionsgenauigkeit, einbüßen. Es ist deshalb nötig, eine Beschleunigung und ein Abbremsung des Schrittmotors in die Prozedur einzufügen, um die mechanischen Effekte zu kompensieren. Alternativ wäre ein Start-Stop-Betrieb mit geringer Schrittfrequenz möglich.

Eine geringe Schrittfrequenz zu wählen, so dass der Rotor des Schrittmotors dem Magnetfeld der Statorspulen folgen kann, ist nicht optimal. Bei dieser Art Ansteue-rung würde ein Scanzyklus während der Anfahrzeit der Start- beziehungsweise Nullposition länger dauern als nötig, da diese 50% der Bewegung ausmachen.

Es gibt jedoch die Möglichkeit eine lineare, exponentielle beziehungsweise kubische Beschleunigungs- oder Bremsrampe zu nutzen. Bei allen drei Methoden wird die Schrittfrequenz während der Bewegung erhöht oder verringert. Man kann die Rampenfunktionen sowohl als mathematische Funktionen im steuernden Microcontroller, als auch durch externe Analogschaltungen erzeugen. In der Analogtechnik werden dazu Ladekurven von Kondensatoren und Timer eingesetzt. Im Detail soll darauf an dieser Stelle jedoch nicht eingegangen werden. Entsprechende Schaltungen finden sich in entsprechender Fachliteratur.

a. Lineare Rampe:

Eine lineare Rampe wird in einem Drehzahlbereich angewendet, in dem die Drehmomentkennlinie nicht wesentlich abfällt. Dabei werden der Motor und die Last linear beschleunigt und abgebremst. Für diese Methode muss der Motor ein konstantes Drehmoment aufweisen. Die Flankensteilheit der Ansteuerkurven (Rampenfunktion) hängt dann von diesem ab. Mathematisch lässt sich die lineare Rampe in Folgender Form darstellen:

Beschleunigung: $$n = \frac{\Delta n}{\Delta t * t} + n_s$$

Abbremsung: $$n = \frac{-\Delta n}{\Delta t * t} + n_s$$

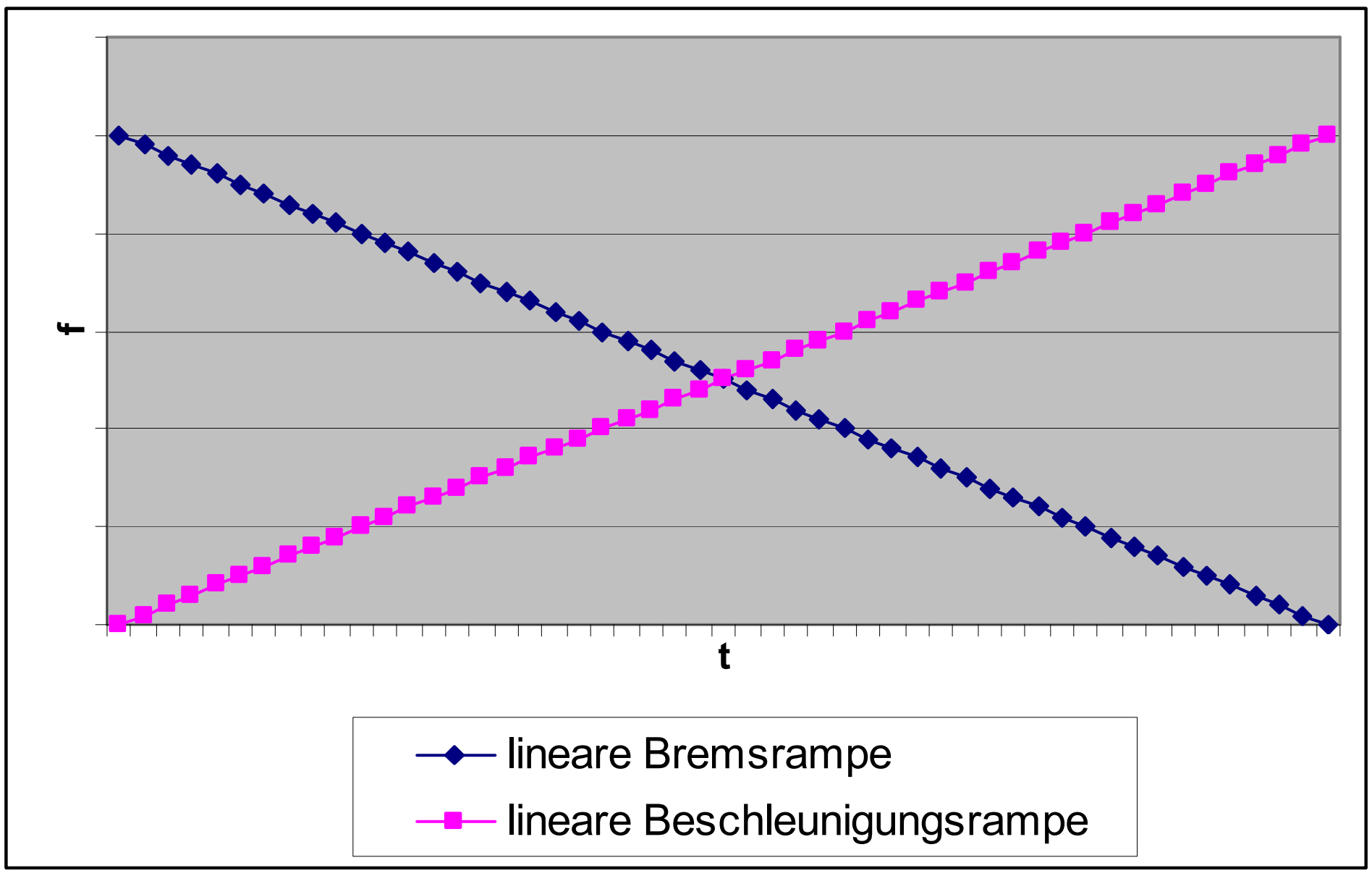

Abbildung 62 - Lineare Rampen

In den Berechnungsformeln stellen n die Drehzahl zum Zeitpunkt t, $n_s$ die Start-Stop-Drehzahl und $\Delta n$ die Drehzahländerung im Zeitintervall $\Delta t$ dar.

b. <u>Exponentielle Rampe:</u>

In Gegensatz zur linearen Rampe erfordert die exponentielle Rampe einen relativ hohen Rechenaufwand. Die Form der Kurve passt sich jedoch an die Drehmomentkurve des Motors an, welche abfallende Drehmomente bei höheren Drehzahlen aufweist.

Im unteren Frequenzbereich weist das Drehmoment annähernd einen linearen Verlauf auf. Im höheren Bereich der Drehmomentkurve verringert sich der Anstieg der Kurve und somit die Beschleunigung. Damit passt sich die Beschleunigung dem bei höherer Drehzahl abnehmenden Beschleunigungsmoment an.

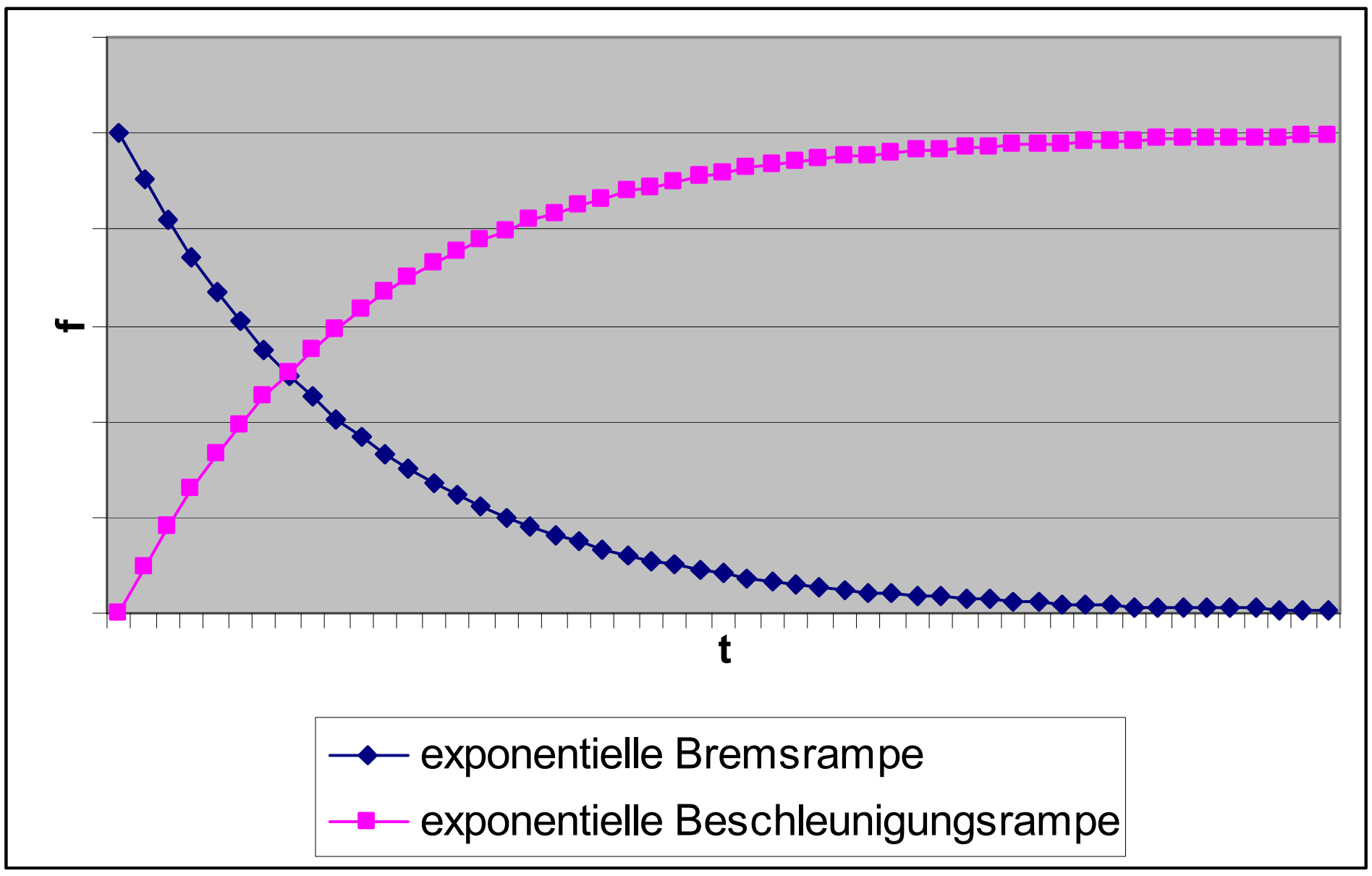

Abbildung 63 - Exponentielle Rampen

Die Rampenberechnung erfolgt unter Berücksichtigung der Dreh-, Beschleunigungs- und Bremsmomente in folgender Weise:

Beschleunigung: $$n = (n_0 - n_s) * e^{\frac{-(t_{br} - t)}{\tau}} + n_s$$

Abbremsung: $$n = (n_0 - n_s) * \left(1 - e^{\frac{-(t_{br} - t)}{\tau}}\right) + n_s$$

Die Funktionen sind der analogen Schaltungstechnik entliehen. Dabei entspricht $n_s$ der Start-Stop-Drehzahl, $n_0$ der angestrebten Endfrequenz, $t_{br}$ der Bremszeit und $\tau$ der Zeitkonstante. Als Schaltung wird $\tau = R * C$ durch eine Kondensatorentladekurve realisiert.

„Die Koeffizienten in den Formeln richten sich nach dem Verlauf der Drehmomentkennlinie. In der Realität wird man die Kurve stückweise konstruieren und durch Geraden annähern, deren Steilheit wie bei einer linearen Rampe zu bemessen ist.“ [Schörlin 1996]

c. <u>S-Kurve (kubische Parabel)</u>

Der Einsatz einer kubischen Parabel, oder S-Kurve, in der Schrittmotorsteuerung ist die beste Lösung.

Im Vergleich zu den linearen und exponentiellen Rampen entstehen am Beginn und Ende der Rampen keine Sprünge. Damit entstehen weniger Schrittverluste, durch die mechanischen Eigenschaften des Schrittmotors, wie bei anderen Rampenformen.

Die kubische Parabel passt sich am besten an die Motordrehmomentlinie an.

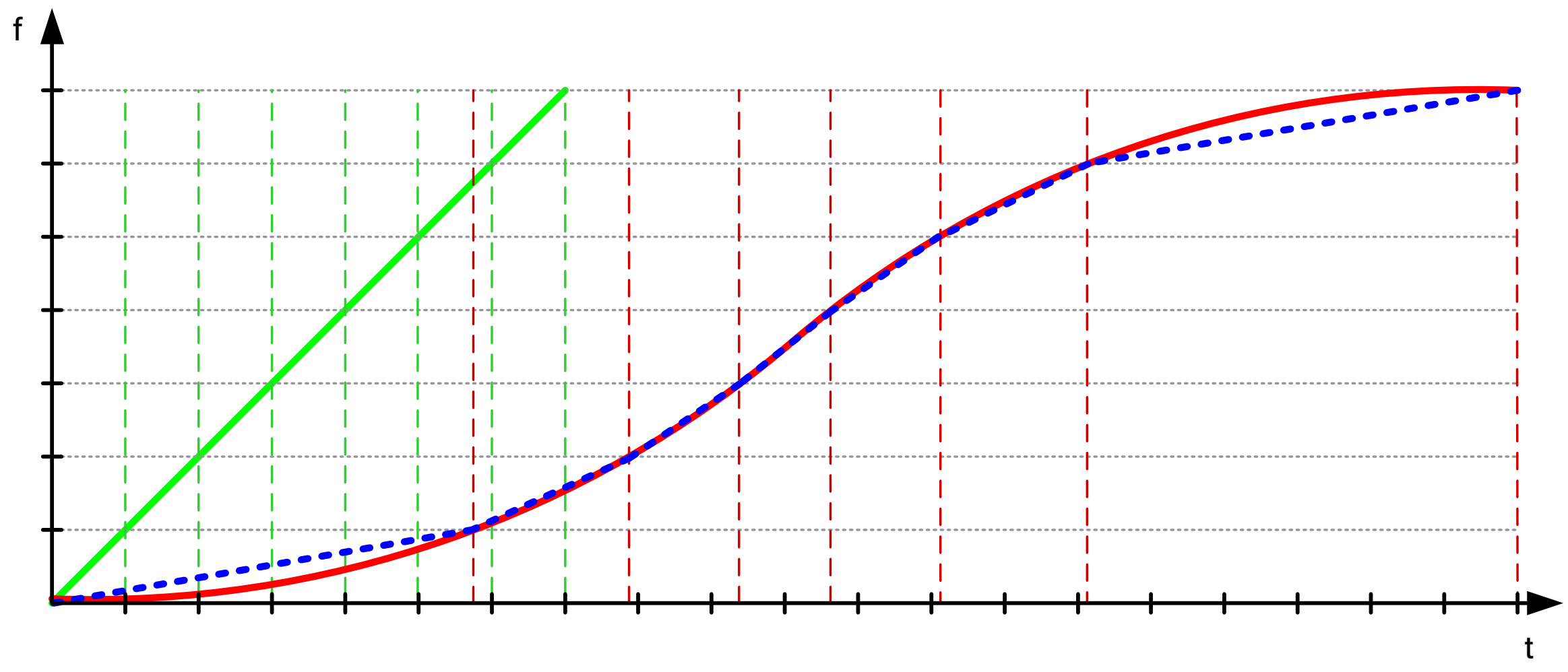

Abbildung 64 - S-Kurven-Modellierung (kubische Parabel)

Eine kubische Parabel (in Abbildung 64 rot) kann am einfachsten modelliert werden, in dem eine lineare Rampe (in Abbildung 64 grün) logarithmisch gestreckt, bzw. gestaucht wird.

In der Realität wird die S-Kurve zwischen den Stützpunkten als lineare Funktionen (in Abbildung 64 blau) errechnet. Die Teilstücke unterscheiden sich somit ausschließlich durch ihren Anstieg. Sie werden wie im Abschnitt 6.6.4.a beschrieben berechnet.

### 6.6.5 Kommunikationsprozedur

Die Kommunikation kann über die RS232-Schnittstelle (serielle Schnittstelle) oder den I²C-Bus erfolgen.

Beide Schnittstellenabfrageprozeduren werden beim Eintreffen eines Befehls jeweils per Interrupt aktiviert. Der entsprechende Datenframe wird verarbeitet und der enthaltene Befehl an die Befehlsverarbeitungsprozedur übergeben.

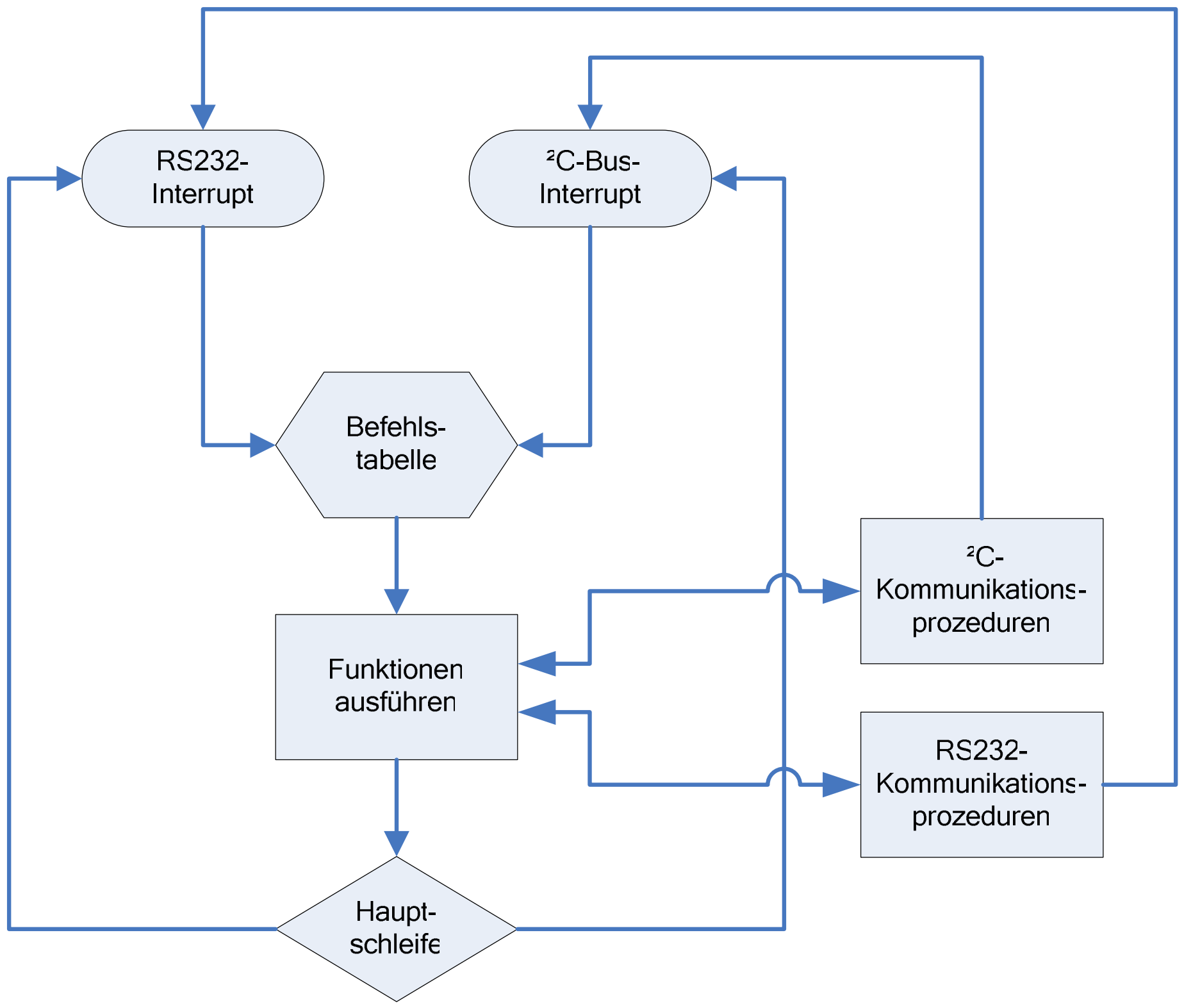

Abbildung 65 - Kommunikationsprozedur und Interruptes

Die softwaretechnische Umsetzung der Kommunikationsprozeduren ist ausführlich in den jeweiligen Quellcodes dokumentiert.

# 7 Abschlussbetrachtung

Das in dieser Diplomarbeit vorgestellte Umgebungserfassungssystem stellt den Prototypen eines funktionsfähigen Grundsystems dar. Durch die bereits in der Konzeptionsphase eingeplanten Systemressourcen (im Microcontroller, 256kByte externer I²C-EEPROM, zwei Schrittmotoranschlüsse und –steuerungen,…) kann das System erweitert, verändert oder auf ein beliebigen Roboter angepasst werden. Die Skalierbarkeit des Systems war bereits in der Planungsphase ein wichtiges Kriterium, da es sich um ein Modul für experimentelle Roboter handelt.

Da das System in einem relativ kurzen Zeitraum konzipiert und gebaut wurde, bestehen noch Optimierungsmöglichkeiten sowohl in der Programmierung des Moduls, als auch beim Aufbau des Scannerkopfes.

Die wesentlichen Funktionen wurden jedoch im Modul integriert, wodurch eine Grundlage für eine problemlose Erweiterung der Programmierung geschaffen wurde.

Auf dem beiliegendem Datenträger finden sich ebenfalls Beispiele für die Modifikationsmöglichkeiten des Sensorkopfes und der Mechanischen Komponenten.

Um die Nachvollziehbarkeit des Projektes zu erhöhen und Zusatzinformationen zu den einzelnen Modulelementen geben zu können, finden sich auf der CD sowohl erweiterte, als auch zusätzliche Kapitel zu dieser Diplomarbeit. Alle Details in die Druckversion zu integrieren, hätte den Rahmen der Arbeit überschritten. Sie vermittelt jedoch einen grundlegenden Überblick.

# Literatur

## Sonstige Quellen

# Verwendete Formelzeichen

Dieses Verzeichnis enthält die in der Arbeit verwendeten Symbole und Formelzeichen.

| Mathematische Größen | |
|---|---|
| $P_m(P_{mx}/P_{my})$ | Messpunkt (Standpunkt des Scanners) |
| $P_o(P_{ox}/P_{oy})$ | Objektpunkt (Koordinaten des Objektes relativ zum Scanner) |
| $l_r$ | Rasterbreite (Aufteilungsraster zur Umgebungsspeicherung) |
| $l_m$ | Gemessene Entfernung zum Objekt |
| $l_{\Delta x}$ | Gerasterter x-Abstand zwischen dem Scanner und dem Objekt |
| $l_{\Delta y}$ | Gerasterter y-Abstand zwischen dem Scanner und dem Objekt |
| $\Delta\varepsilon$ | Abtastsegmentgröße (in °) |
| $\Delta\varphi$ | Messsegmentgröße (in °) |

Tabelle 10 - Mathematische Größen

| Physikalische Größen | |
|---|---|
| $-B_{\max}$ | Maximale negative Induktivität |
| $+B_{\max}$ | Maximale positive Induktivität |
| $\alpha$ | Schrittwinkel |
| $2p$ | Polzahl |
| $m$ | Strangzahl (Phasenzahl) |
| $S$ | Schrittzahl (Umdrehungsteil) |
| $n$ | Drehzahl (Umdrehungsfrequenz) |
| $f_s$ | Schrittfrequenz |
| $n_s$ | Start-Stop-Drehzahl |
| $n_0$ | angestrebte Enddrehzahl |
| $n_{aus}$ | Drehzahl des ausgehenden Zahnrades |
| $n_{ein}$ | Drehzahl des eingehenden Zahnrades |
| $z_{aus}$ | Zähnezahl des ausgehenden Zahnrades |
| $z_{ein}$ | Zähnezahl des eingehenden Zahnrades |
| $\omega_{aus}$ | Drehwinkel des ausgehenden Zahnrades |
| $\omega_{ein}$ | Drehwinkel des eingehenden Zahnrades |
| i | Unter-/Übersetzungsverhältnis |

| Physikalische Größen | |
|---|---|
| $t_{Messdauer}$ | Zeit, die für den Vermessungsvorgang benötigt wird |
| $t_{Anfahrtszeit_Startposition}$ | Zeit, um den Sensorkopf in Anfangsposition zu drehen |
| $t_{Anfahrtzeit_Nullposition}$ | Zeit, um den Sensorkopf in Ruheposition zu drehen |

Tabelle 11 - Physikalische Größen

| Variablen und Arrays | |
|---|---|
| L | Layer |
| Lmax | Anzahl der Layer |
| n | Neuronennummer |
| nmax | Anzahl der Neuronen im größten Layer |
| c | Verbindungsnummer |
| wmax | Anzahl der Gewichte im größten Layer |
| O[] | Speichermatrix der Neuronenausgaben |
| W[] | Gewichtsmatrix des künstlichen neuronalen Netzes |

Tabelle 12 - Variablen und Arrays

# Glossar

**Abtastung**

Feststellung der Augenblickswerte (Momentanwerte) bei einem analogen Signal in regelmäßigen zeitlichen Abständen. Im Englischen wird die Abtastung Sampling genannt.

**Abtastwert**

Der Abtastwert, auch Sample genannt, ist der gemessene Momentanwert eines analogen Signals.

**ACK (positive acknowledgement)**

Das ACK-Steuerzeichen wird bei der zeichenorientierten, digitalen Übertragung eingesetzt, um den fehlerfreien Empfang des Datenblockes zu bestätigen. Tritt ein Fehler auf, wird das NAK-Steuerzeichen (negativ acknowledgement) gesendet und die Übertragung wiederholt.

**Adresse**

Als Adresse wird eine Bit- oder Zeichenfolge bezeichnet, die Geräte, Schnittstellen, Nutzer oder Anwendungen kennzeichnet und eindeutig identifiziert.

**Aktor**

Ein Aktor ist eine „technische Funktionseinheit, die das Ergebnis einer elektronischen Verarbeitung in [eine] nicht-elektrische Größe umsetzt." [Freyer] Die Spannungswerte oder Signale werden dabei beispielsweise in eine Bewegung eines Motors umgesetzt.

**Analog-Digital-Wandler (ADW)**

Der Analog-Digital-Wandler (ADW) wird auch als Analog-Digital-Umsetzter (ADU) oder im Englischen als analog-digital-converter (ADC) bezeichnet. Der ADC ist eine technische Funktionseinheit, die ein analoges Signal in ein digitales umsetzt.

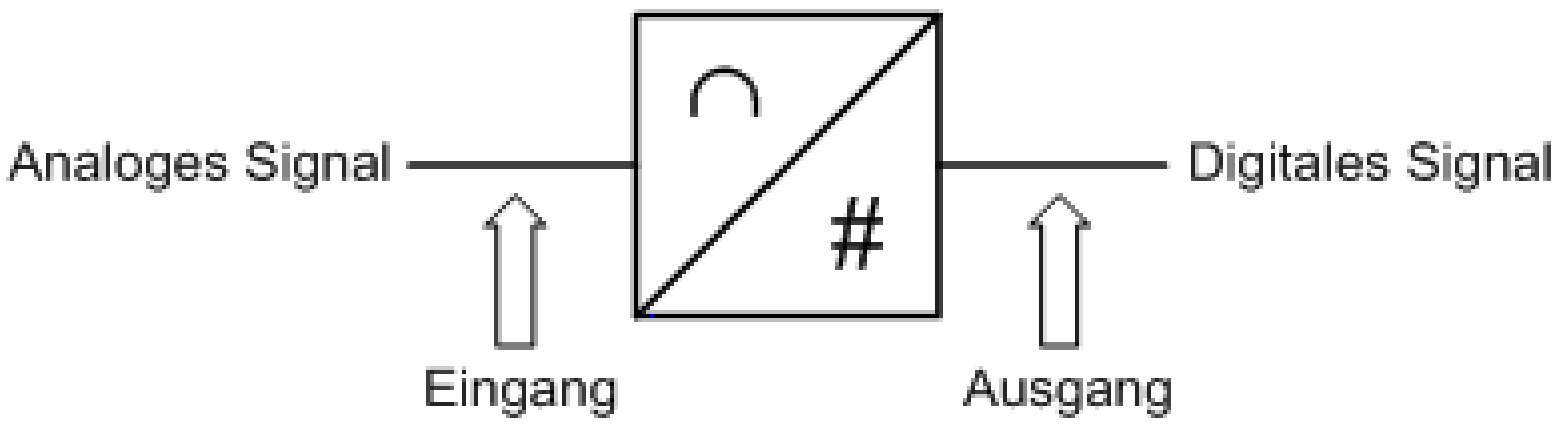

Abbildung 66 - Analog-Digital-Umsetzer

**Baud (Bd)**

Baud ist die Einheit für die Geschwindigkeit der Signalschritte bei digitalen Übertragungen.

**Befehl**

Ein Befehl wird zum Auslösen definierter Aktivitäten eines Programms genutzt. Dadurch werden Unterprogramme aufgerufen, welche Rechenvorgänge, Datenspeicherung, Ladevorgänge oder Hardwareaktivitäten auslösen.

**bidirektional**

Bei bidirektionalen Datenübertragungen ist der Informationstransfer in beide Richtung (A nach B und B nach A) möglich.

**Bus**

Als Bus wird eine Anzahl n von parallelen Signalleitungen bezeichnet, die zur gleichzeitigen Übertragung von Bitmustern genutzt werden.

**CLK (clock)**

Das clock-Signal ist der Zeittakt bei einer Übertragung.

**Daten**

Daten sind Folgen binär kodierter Zeichen, welche in digitalen Systemen verarbeitet, gespeichert und ausgegeben werden können.

**Datenformat**

Als Datenformat wird die Struktur eines Rahmens bei asynchroner Datenübertragung bezeichnet. Er setzt sich dabei aus Rahmendauer, sowie Art und Folge der individuellen Einheit zusammen. Dazu zählen beispielsweise Start-, Daten-, Paritäts- und Stopbits.

**EEPROM**

EEPROM steht für electrically erasable programmable read-only memory. Es ist auch die Bezeichnung $E^2$PROM üblich. Es handelt sich dabei um einen elektrisch löschbaren und reprogrammierbaren Nur-Lese-Speicher.

**Frame**

Ein Frame ist ein Datenrahmen. Es sind „Bitfolgen konstanter Länge, deren Beginn jeweils durch einen Kopf (Header) gekennzeichnet [… sind]. „ [Freyer]

**Fuzzy-Logik**

„Fuzzy-Logik ist eine Methode unsicheres Wissen zu verarbeiten (unscharfe Logik) […], d.h. wenn eine Variable gleichzeitig mehr als einen Wert annehmen müsste." [Ludwig 1991]

**GPS (global positioning system)**

Das GPS ist ein weltweites, satellitengestütztes System für präzise Ortung und Navigation. „Es wird mit 21 umlaufenden Betriebssatelliten gearbeitet, die sich auf 6 Bahnen in ca. 20.000km Höhe bewegen. Der genutzte Frequenzbereich liegt bei 1,5…1,6GHz." [Freyer] Mit Hilfe von mindestens 4 Satelliten kann die exakte Position des Empfängers berechnet werden.

**Host**

Der englische Begriff Host bedeutet Gastgeber oder Hausherr. Es ist eine technische Funktionseinheit, die in einem System die steuernde und/oder überwachende Funktion über die anderen Einheiten und Komponenten wahrnimmt.

**I²C**

I²C (für Inter-Integrated Circuit/Interner Integrierter Schaltungsbus, gesprochen I-Quadrat-C bzw. I-squared-C) ist ein von Philips Semiconductor entwickelter serieller Bus für Computersysteme. Er wird benutzt, um Geräte mit geringer Übertragungsgeschwindigkeit an ein Embedded System oder eine Hauptplatine anzuschließen.

**IC (integrated circuit)**

Integrierte Schaltungen sind auf einem Chip aus Halbleitermaterial eingeprägte mikroskopisch kleine Bauelemente. Mit ICs können analoge und digitale Funktionen realisiert werden. Die Leistungsfähigkeit einer integrierten Schaltung wird als Zahl der integrierten Transistorfunktionen angegeben und durch einen Integrationsgrad gekennzeichnet.

**Master**

Master bedeutet Herr oder Meister. Es ist ein Zentralgerät, welches die Überwachung oder Steuerung der dem System angeschlossenen Geräte (Slaves) übernimmt.

**MIPS (million of instructions per second)**

MIPS gib an, wie viel Millionen Befehle (Instruktionen) ein Prozessor oder Microcontroller pro Sekunde verarbeiten kann.

**Momentanwert**

Für den Momentanwert existiert auch die Bezeichnung Augenblickswert. Es ist der Wert einer physikalischen Größe (z.B. Spannung) zu einem definierten Zeitpunkt.

**NAK (negativ acknowledgement)**

Als NAK wird ein Steuerzeichen in der digitalen, zeichenorientierten Übertragung bezeichnet, welches bei einem fehlerhaft übertragenen Datenblock gesendet wird. Dies verursacht ein erneutes senden des Datenblockes.

**Neuronales Netz**

Neuronale Netze beziehen sich auf die Strukturen des Gehirns von Tieren und Menschen: Neuronen sind in der Art eines Netzes miteinander verknüpft. Biologische Neuronen reagieren auf elektrische oder chemische Reize. Neuronen haben üblicherweise mehrere Eingangsverbindungen sowie eine Ausgangsverbindung. Wenn die Summe der Eingangsreize einen gewissen Schwellenwert überschreitet, "feuert" das Neuron. Das bedeutet, dass ein Aktionspotential an seinem Axonhügel ausgelöst und entlang seines Axons bewegt wird.

**Puls-Weiten-Modulation (PWM)**

Puls-Weiten-Modulation nennt man die Modulation eines Rechtecksignals in seinem Tastverhältnis. Die Periodendauer des Signals bleibt hierbei konstant. PWM ist im Deutschen auch unter PBM (Puls Breiten Modulation) und Pulsdauermodulation bekannt.

**RS232C**

Die RS232C-Schnittstelle ist eine standardisierte, bidirektionale, serielle Schnittstelle mit einer Datenübertragungsrate von bis zu 19,2kbit/s auf unsymmetrischen Leitungen mit bis zu 15m Länge.

**zeichenorientierte Übertragung**

Damit wird eine Übertragungsprozedur bezeichnet, die durch den Austausch zeichenstrukturierter Rahmen gekennzeichnet ist.

# Technischer Anhang und Übersichtstabellen

## IR-Sensoren-Preisübersicht

In der folgenden Tabelle sind die preisgünstigsten Lieferanten gelistet, die bei der Recherche gefunden wurden. Die Tabelle befindet sich auf dem Stand vom Oktober 2005.

| Sensor | Lieferant | Preis |
|---|---|---|
| GP2D12 | www.dcd-robotik.de | 12,80€ |
| GP2D120 | www.rsonline.de | 17,00€ |
| GP2D15 | www.mercato.com | 16.30€ |
| GP2D150 | www.mercato.com | 17,90€ |
| GP2Y0A02YK | www.rsonline.de | 19,00€ |
| GP2Y0D02YK | www.mercato.com | 19,00€ |

Tabelle 13 - IR-Sensoren-Preisübersicht

## Ansteuertabellen für Schrittmotore

| Unipolare Ansteuerung (Vollschritt)[8] | | | | | | |
|---|---|---|---|---|---|---|
| | Phase Spule 1a | Phase Spule 1b | Common[9] Spule 1 | Phase Spule 2a | Phase Spule 2b | Common[10] Spule 2 |
| Schritt 1 | + | 0 | - | 0 | + | - |
| Schritt 2 | + | 0 | - | + | 0 | - |
| Schritt 3 | 0 | + | - | + | 0 | - |
| Schritt 4 | 0 | + | - | 0 | + | - |

Tabelle 14 - Ansteuerungssequenz eines unipolaren Schrittmotors (Vollschrittbetrieb)

| Bipolare Ansteuerung (Vollschritt) | | | | | | |
|---|---|---|---|---|---|---|
| | Phase Spule 1a | Phase Spule 1b | Common Spule 1 | Phase Spule 2a | Phase Spule 2b | Common Spule 2 |
| Schritt 1 | + | - | 0 | + | - | 0 |
| Schritt 2 | + | - | 0 | - | + | 0 |
| Schritt 3 | - | + | 0 | - | + | 0 |
| Schritt 4 | - | + | 0 | + | - | 0 |

Tabelle 15 - Ansteuerungssequenz eines bipolaren Schrittmotors (Vollschrittbetrieb)

[8] - steht für Minus, + steht für Plus, 0 steht für nicht belegt

[9] Mittelabgriff der Motorspule 1

[10] Mittelabgriff der Motorspule 2

| Bipolare Ansteuerung (Halbschritt) | | | | | | |
|---|---|---|---|---|---|---|
| | Phase Spule 1a | Phase Spule 1b | Common Spule 1 | Phase Spule 2a | Phase Spule 2b | Common Spule 2 |
| Schritt 1 | + | - | 0 | + | - | 0 |
| Schritt 2 | + | - | 0 | 0 | 0 | 0 |
| Schritt 3 | + | - | 0 | - | + | 0 |
| Schritt 4 | 0 | 0 | 0 | - | + | 0 |
| Schritt 5 | - | + | 0 | - | + | 0 |
| Schritt 6 | - | + | 0 | 0 | 0 | 0 |
| Schritt 7 | - | + | 0 | + | - | 0 |
| Schritt 8 | 0 | 0 | 0 | + | - | 0 |

Tabelle 16 - Ansteuerungssequenz eines bipolaren Schrittmotors (Halbschrittbetrieb)

# CD-Inhaltsverzeichnis

Auf dem Datenträger befinden sich die in Text beschriebenen:

- Diplomarbeit als HTML- und PDF-Version
- Messdaten und Diagramme
- Datenblätter im PDF-Format
- Schaltpläne und Platinenlayouts
- Tools und verwendete Programme
- Quellkode des Microcontrollerprogrammes
- Bedienungsanleitung des Moduls

Legen Sie den Datenträger in ein CD-Laufwerk ein. Sollte die Indexseite nicht per Auto-Play starten, rufen Sie die „index.html" auf.

# Schaltpläne

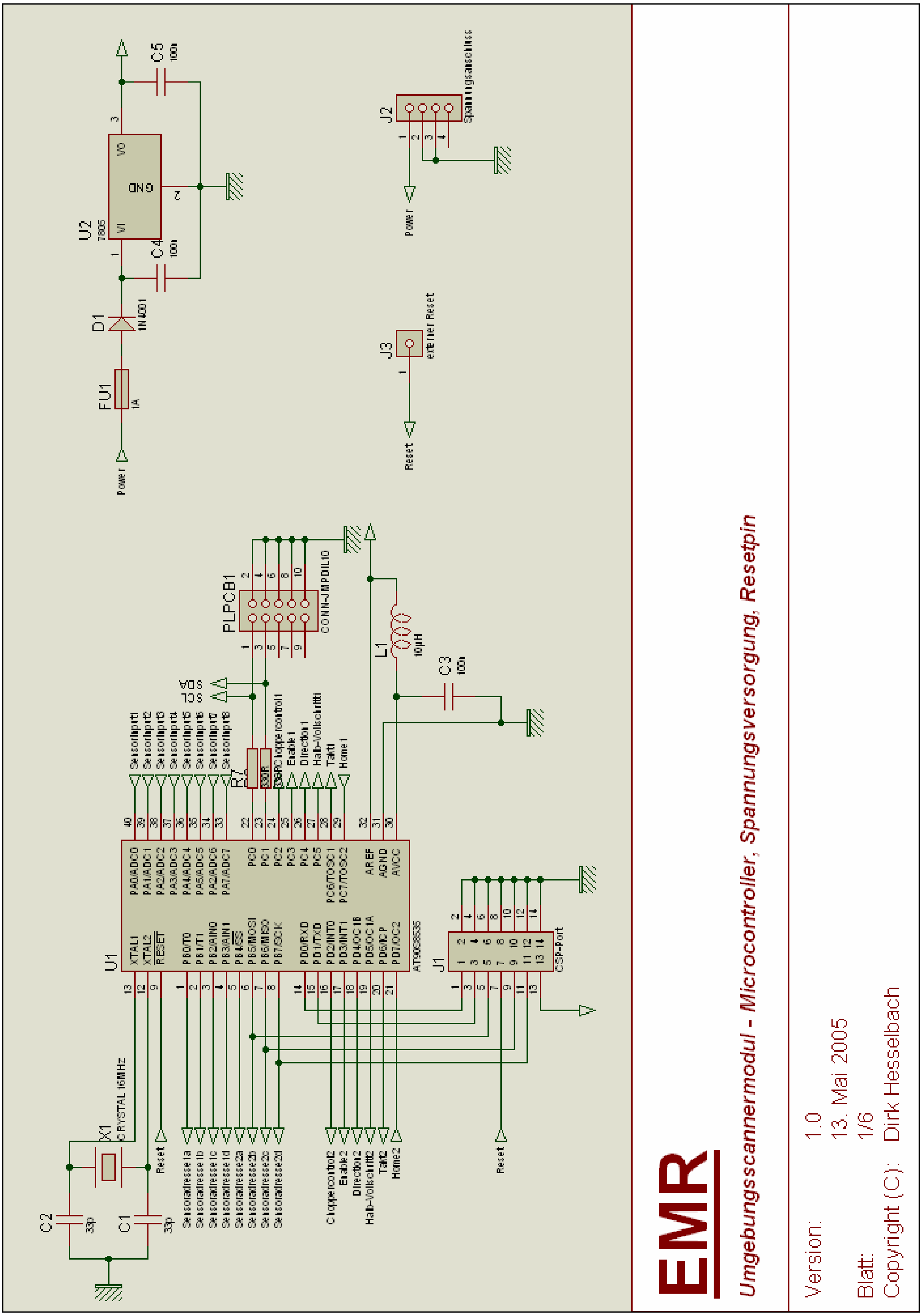

Abbildung 67 - Umgebungsscanner Schaltplan 1/6

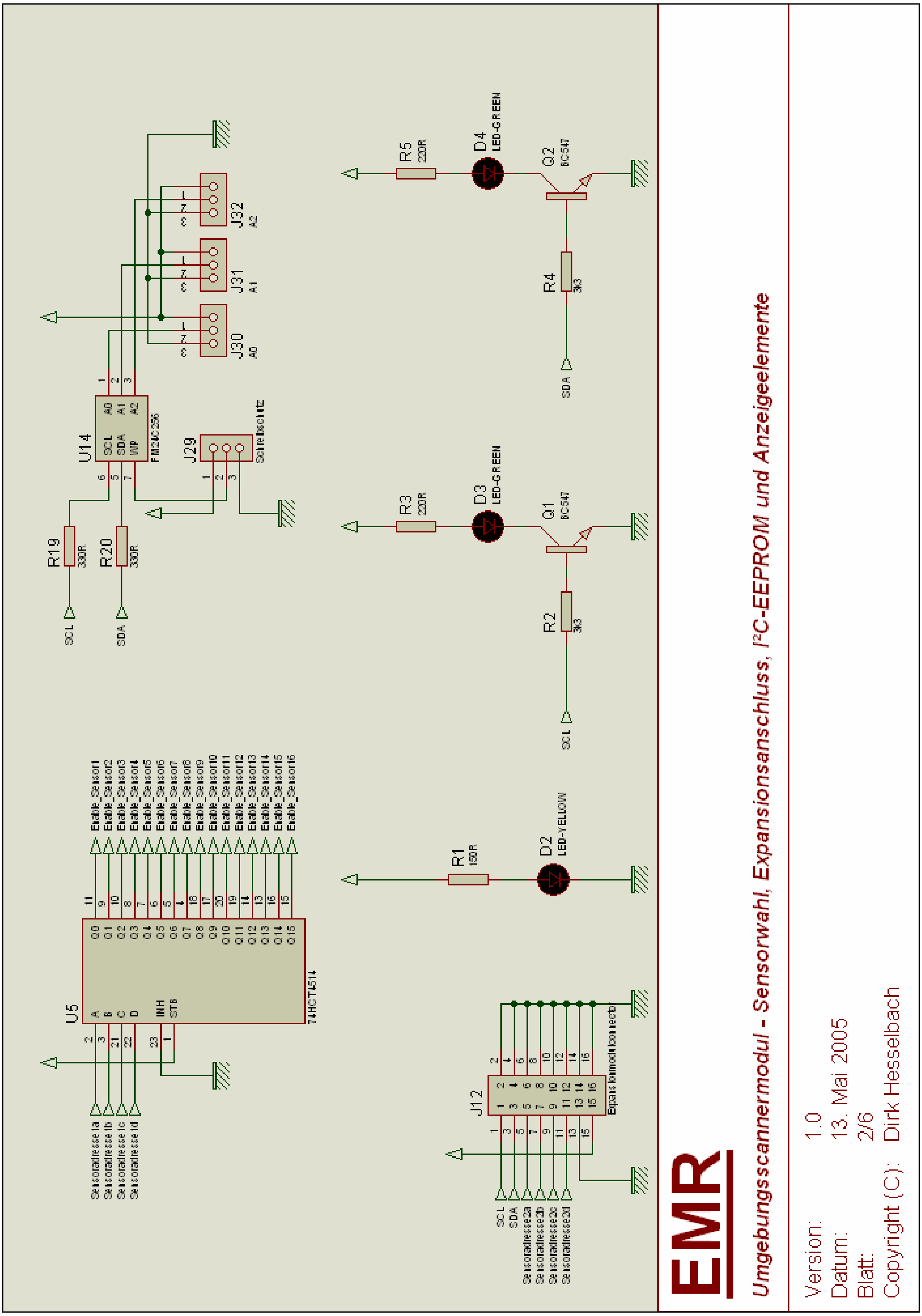

Abbildung 68 Umgebungsscanner Schaltplan 2/6

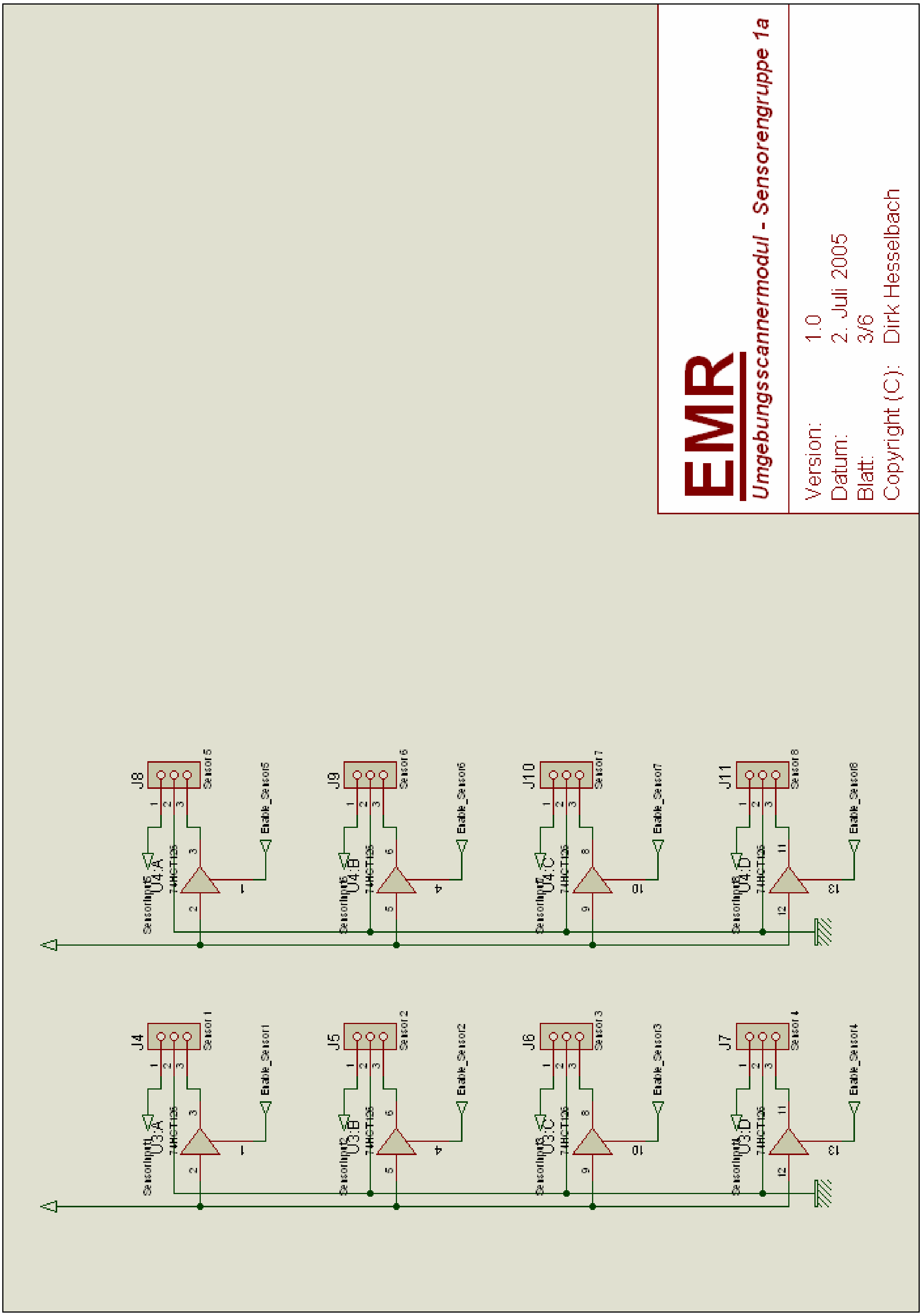

Abbildung 69 Umgebungsscanner Schaltplan 3/6

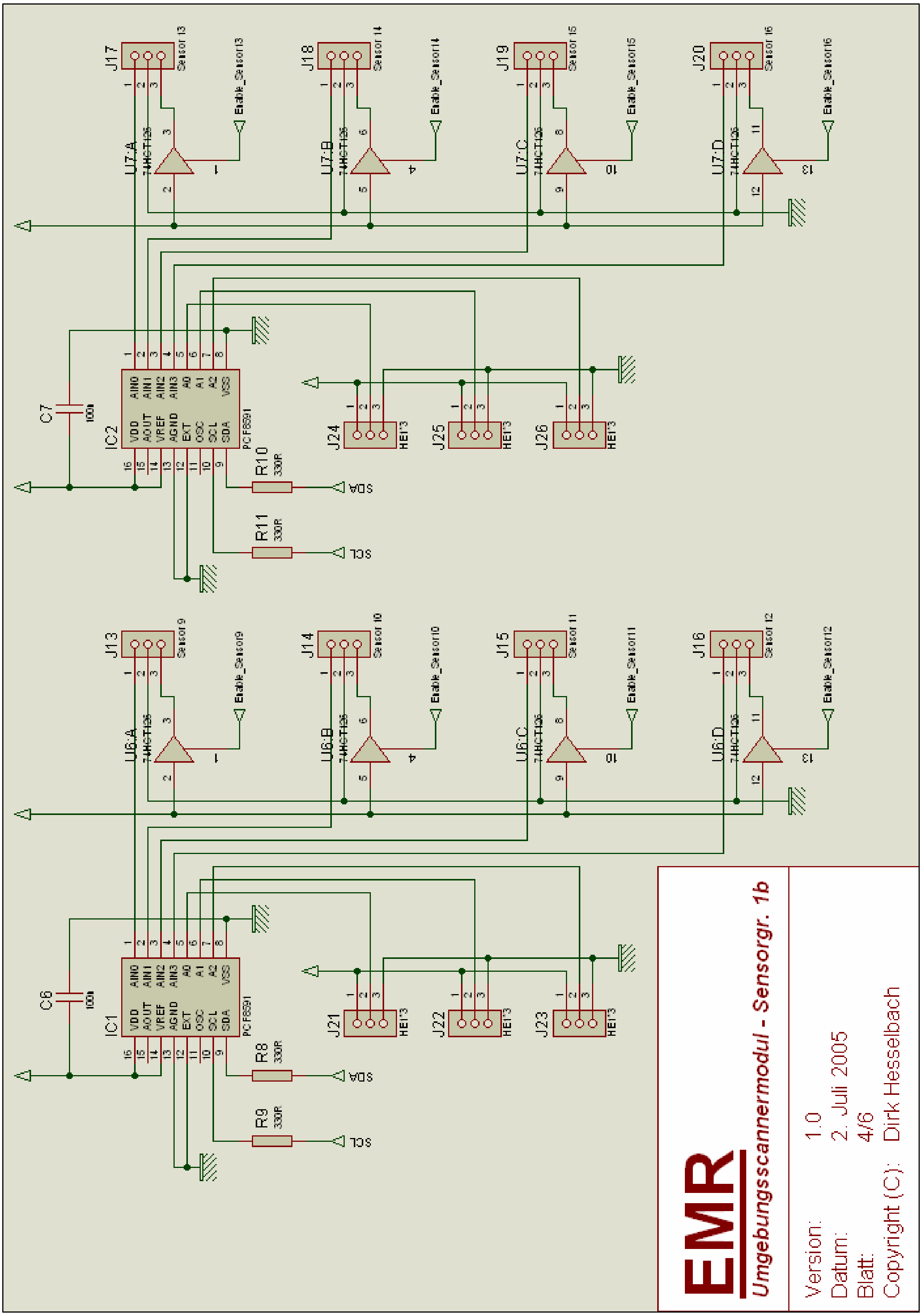

Abbildung 70 Umgebungsscanner Schaltplan 4/6

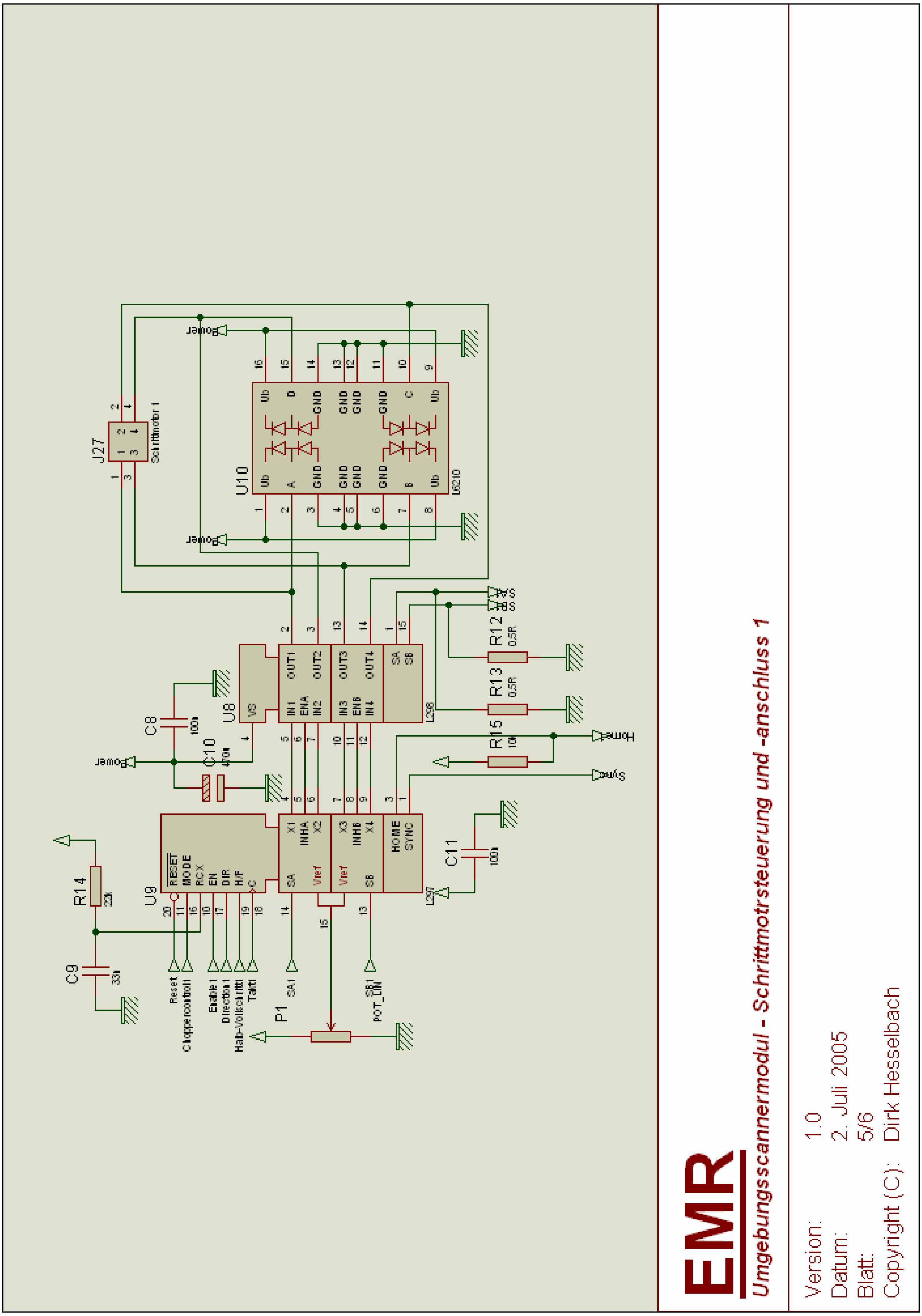

Abbildung 71 Umgebungsscanner Schaltplan 5/6

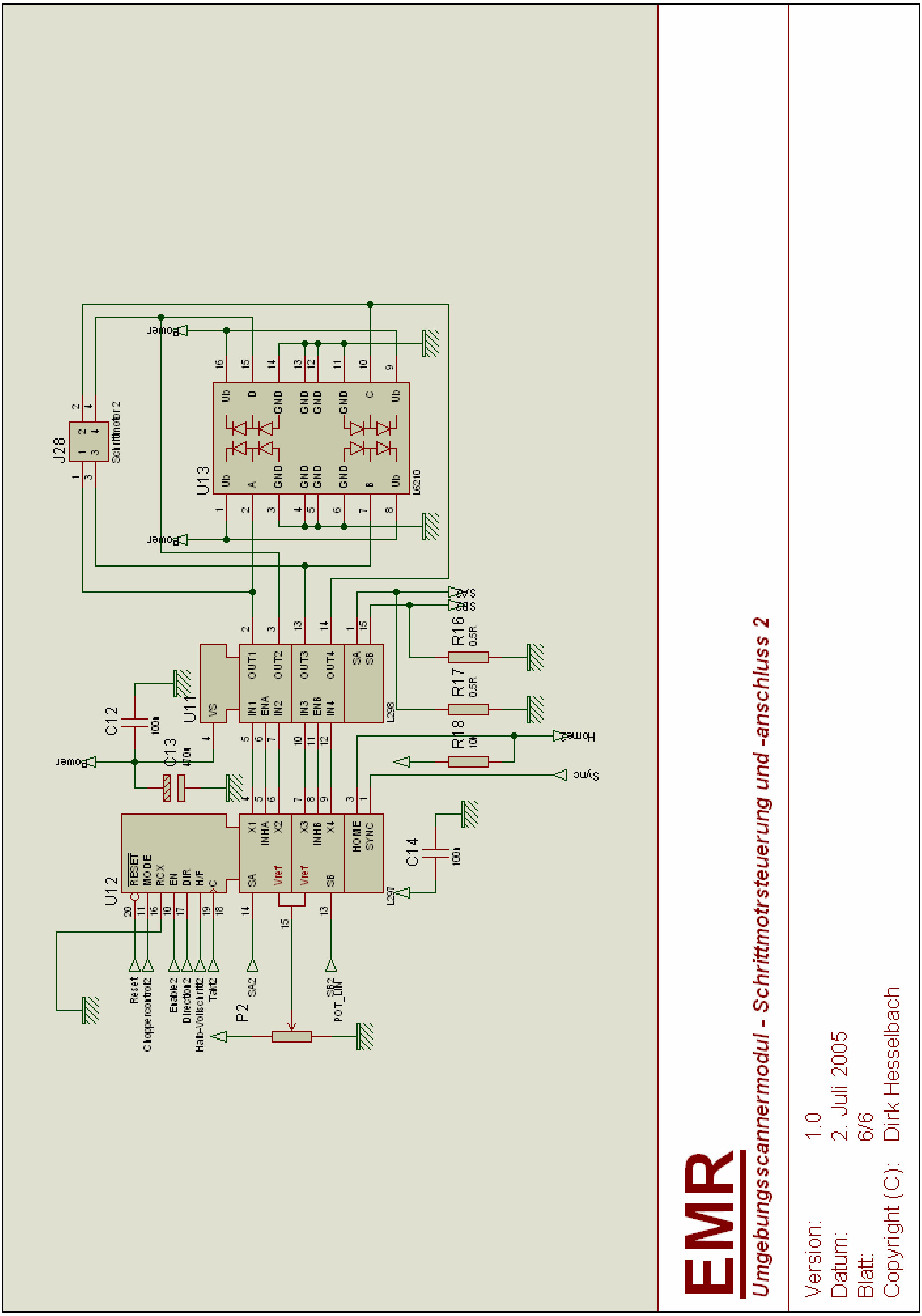

Abbildung 72 Umgebungsscanner Schaltplan 6/6

# Platinenlayouts

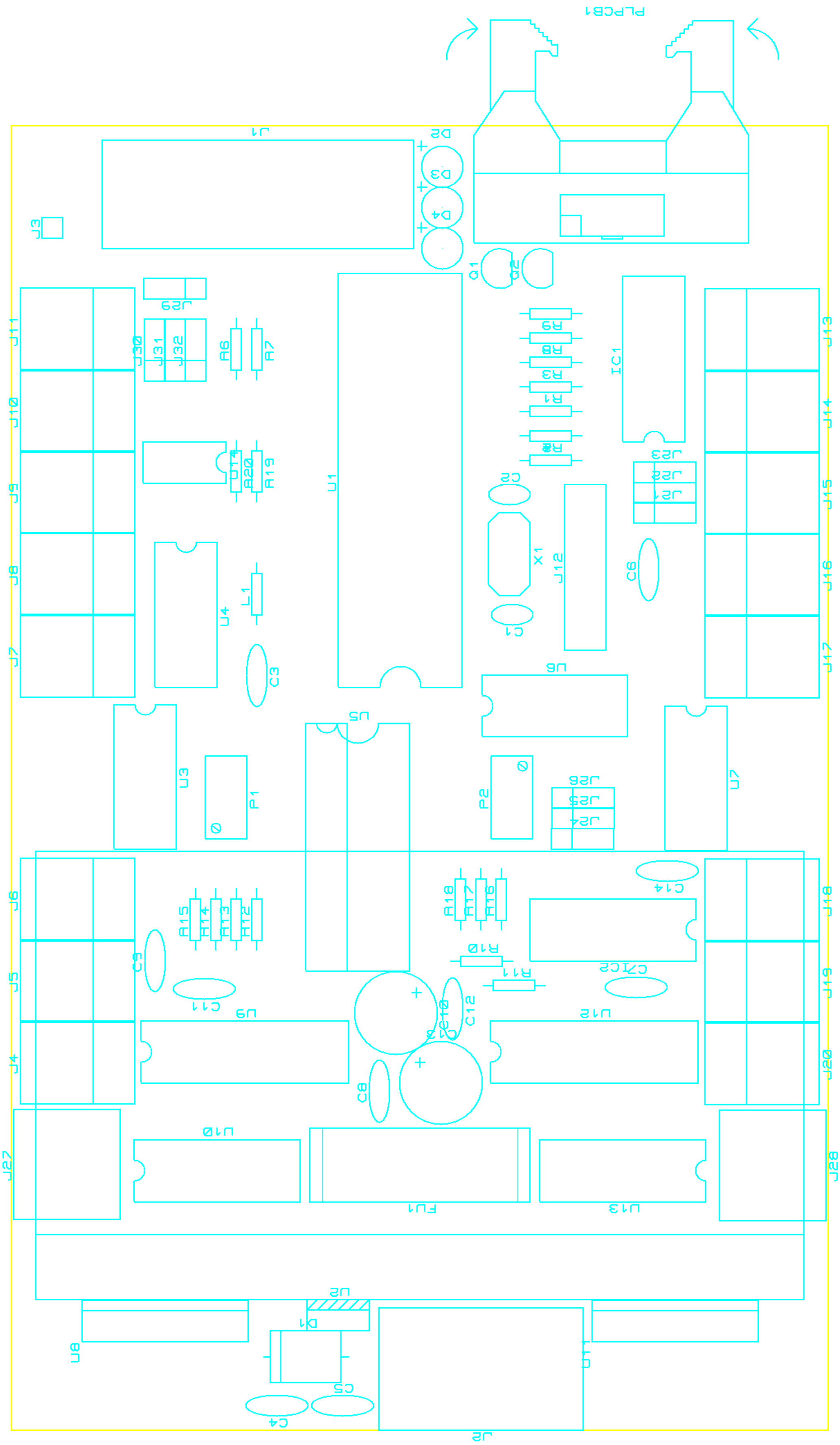

Abbildung 73 - Umgebungsscanner - Bestückungsdruck der Hauptplatine

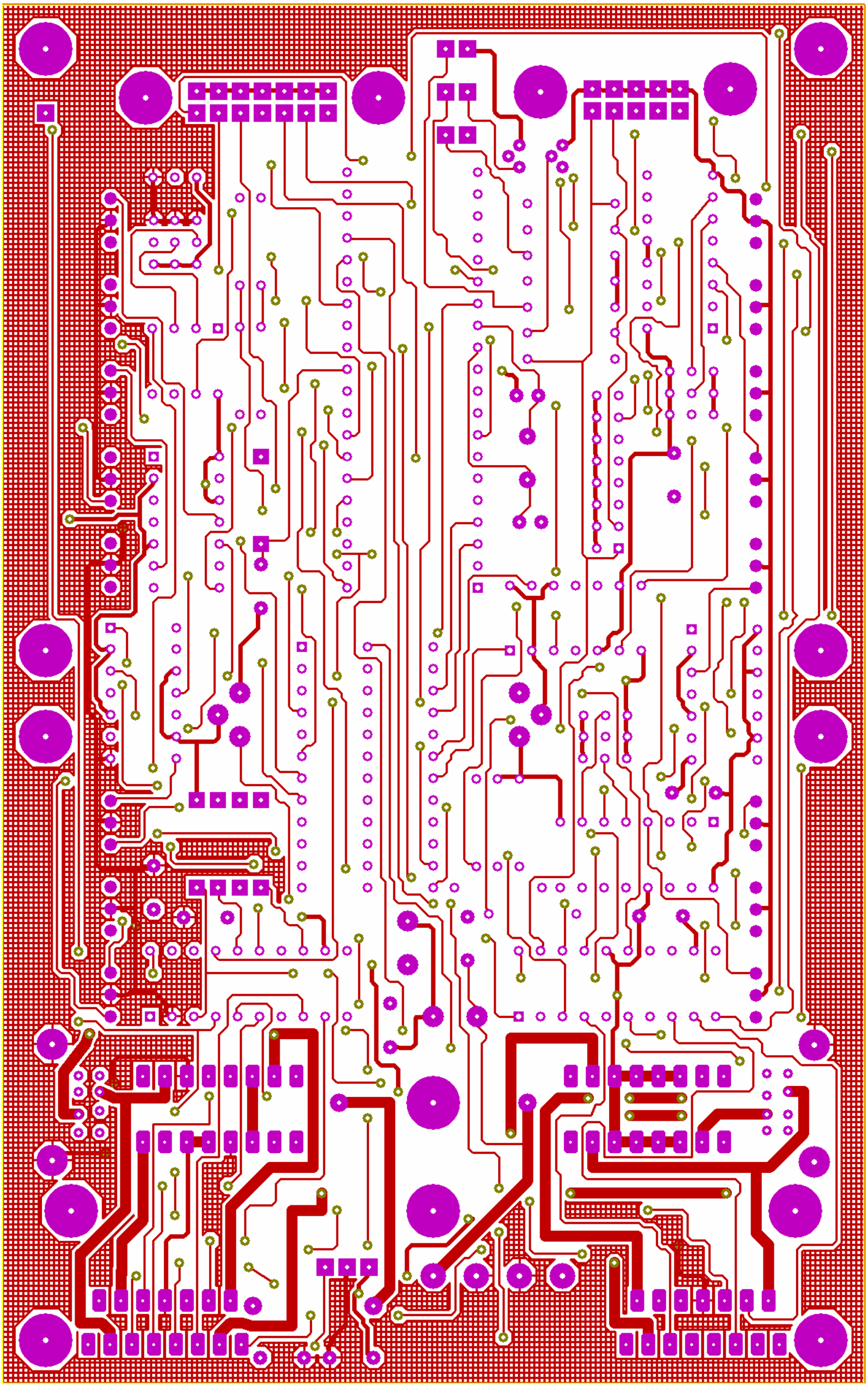

Abbildung 74 – Umgebungsscanner - Top-Layout der Hauptplatine

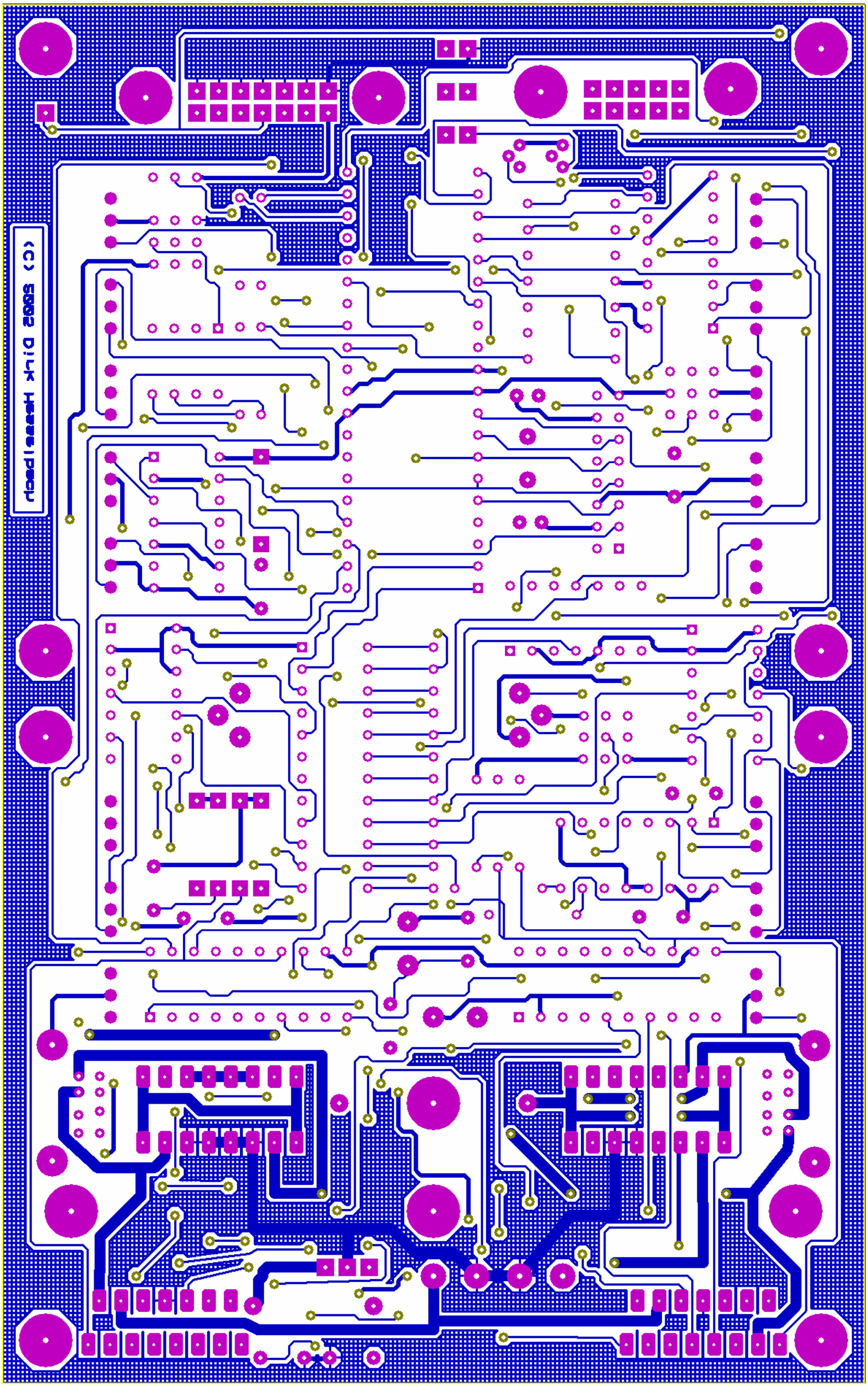

Abbildung 75 – Umgebungsscanner - Bottom-Layout der Hauptplatine

# Befehlstabelle (Auszug)

Die vollständige Befehlstabelle befindet sich auf dem beiliegendem Datenträger.

| Befehl | | Beschreibung |
|---|---|---|
| Hex | Char | |
| 61 31 | a 1 | ADC-Wert von Sensoranschluss 1 abfragen |
| 61 32 | a 2 | ADC-Wert von Sensoranschluss 2 abfragen |
| 61 33 | a 3 | ADC-Wert von Sensoranschluss 3 abfragen |
| 61 34 | a 4 | ADC-Wert von Sensoranschluss 4 abfragen |
| 61 35 | a 5 | ADC-Wert von Sensoranschluss 5 abfragen |
| 61 36 | a 6 | ADC-Wert von Sensoranschluss 6 abfragen |
| 61 37 | a 7 | ADC-Wert von Sensoranschluss 7 abfragen |
| 61 38 | a 8 | ADC-Wert von Sensoranschluss 8 abfragen |
| 61 39 | a 9 | ADC-Wert von Sensoranschluss 9 abfragen |
| 61 3A | a : | ADC-Wert von Sensoranschluss 10 abfragen |
| 61 3B | a ; | ADC-Wert von Sensoranschluss 11 abfragen |
| 61 3C | a < | ADC-Wert von Sensoranschluss 12 abfragen |
| 61 3D | a = | ADC-Wert von Sensoranschluss 13 abfragen |
| 61 3E | a > | ADC-Wert von Sensoranschluss 14 abfragen |
| 61 3F | a ? | ADC-Wert von Sensoranschluss 15 abfragen |
| 61 40 | a @ | ADC-Wert von Sensoranschluss 16 abfragen |
| 61 41 | a A | ADC-Wert von Sensoranschluss 17 abfragen (Erweiterung) |
| 61 42 | a B | ADC-Wert von Sensoranschluss 18 abfragen (Erweiterung) |
| 61 43 | a C | ADC-Wert von Sensoranschluss 19 abfragen (Erweiterung) |
| 61 44 | a D | ADC-Wert von Sensoranschluss 20 abfragen (Erweiterung) |
| 61 45 | a E | ADC-Wert von Sensoranschluss 21 abfragen (Erweiterung) |
| 61 46 | a F | ADC-Wert von Sensoranschluss 22 abfragen (Erweiterung) |
| 61 47 | a G | ADC-Wert von Sensoranschluss 23 abfragen (Erweiterung) |
| 61 48 | a H | ADC-Wert von Sensoranschluss 24 abfragen (Erweiterung) |
| 61 49 | a I | ADC-Wert von Sensoranschluss 25 abfragen (Erweiterung) |
| 61 4A | a J | ADC-Wert von Sensoranschluss 26 abfragen (Erweiterung) |
| 61 4B | a K | ADC-Wert von Sensoranschluss 27 abfragen (Erweiterung) |
| 61 4C | a L | ADC-Wert von Sensoranschluss 28 abfragen (Erweiterung) |
| 61 4D | a M | ADC-Wert von Sensoranschluss 29 abfragen (Erweiterung) |
| 61 4E | a N | ADC-Wert von Sensoranschluss 30 abfragen (Erweiterung) |
| 61 4F | a O | ADC-Wert von Sensoranschluss 31 abfragen (Erweiterung) |

| Befehl | | Beschreibung |
|---|---|---|
| **Hex** | **Char** | |
| 61 50 | a P | ADC-Wert von Sensoranschluss 32 abfragen (Erweiterung) |
| 62 31 | b 1 | Entfernung von IR-Sensor 1 berechnen |
| 62 32 | b 2 | Entfernung von IR-Sensor 2 berechnen |
| 62 33 | b 3 | Entfernung von IR-Sensor 3 berechnen |
| 62 34 | b 4 | Entfernung von IR-Sensor 4 berechnen |
| 62 35 | b 5 | Entfernung von IR-Sensor 5 berechnen |
| 62 36 | b 6 | Entfernung von IR-Sensor 6 berechnen |
| 62 37 | b 7 | Entfernung von IR-Sensor 7 berechnen |
| 62 38 | b 8 | Entfernung von IR-Sensor 8 berechnen |
| 62 39 | b 9 | Entfernung von IR-Sensor 9 berechnen |
| 62 3A | b : | Entfernung von IR-Sensor 10 berechnen |
| 62 3B | b ; | Entfernung von IR-Sensor 11 berechnen |
| 62 3C | b < | Entfernung von IR-Sensor 12 berechnen |
| 62 3D | b = | Entfernung von IR-Sensor 13 berechnen |
| 62 3E | b > | Entfernung von IR-Sensor 14 berechnen |
| 62 3F | b ? | Entfernung von IR-Sensor 15 berechnen |
| 62 40 | b @ | Entfernung von IR-Sensor 16 berechnen |
| 62 41 | b A | Entfernung von IR-Sensor 17 berechnen (Erweiterung) |
| 62 42 | b B | Entfernung von IR-Sensor 18 berechnen (Erweiterung) |
| 62 43 | b C | Entfernung von IR-Sensor 19 berechnen (Erweiterung) |
| 62 44 | b D | Entfernung von IR-Sensor 20 berechnen (Erweiterung) |
| 62 45 | b E | Entfernung von IR-Sensor 21 berechnen (Erweiterung) |
| 62 46 | b F | Entfernung von IR-Sensor 22 berechnen (Erweiterung) |
| 62 47 | b G | Entfernung von IR-Sensor 23 berechnen (Erweiterung) |
| 62 48 | b H | Entfernung von IR-Sensor 24 berechnen (Erweiterung) |
| 62 49 | b I | Entfernung von IR-Sensor 25 berechnen (Erweiterung) |
| 62 4A | b J | Entfernung von IR-Sensor 26 berechnen (Erweiterung) |
| 62 4B | b K | Entfernung von IR-Sensor 27 berechnen (Erweiterung) |
| 62 4C | b L | Entfernung von IR-Sensor 28 berechnen (Erweiterung) |
| 62 4D | b M | Entfernung von IR-Sensor 29 berechnen (Erweiterung) |
| 62 4E | b N | Entfernung von IR-Sensor 30 berechnen (Erweiterung) |
| 62 4F | b O | Entfernung von IR-Sensor 31 berechnen (Erweiterung) |
| 62 50 | b P | Entfernung von IR-Sensor 32 berechnen (Erweiterung) |
| 65 | e | lokale Umgebung erfassen (Umgebungsscan) |

| Befehl | | Beschreibung |
|---|---|---|
| Hex | Char | |
| 6D 61 | m a | Schrittmotor 1 abschalten |
| 6D 65 | m e | Schrittmotor 1 einschalten |
| 6D 68 | m h | Schrittmotor 1 auf Halbschrittmodus einstellen |
| 6D 6C | m l | Drehrichtung von Schrittmotor 1 links |
| 6D 72 | m r | Drehrichtung von Schrittmotor 1 rechts |
| 6D 73 | m s | Schrittmotor 1 führt Schritt aus |
| 6D 76 | m v | Schrittmotor 1 auf Vollschrittmodus einstellen |
| 6E 61 | n a | Schrittmotor 2 abschalten |
| 6E 65 | n e | Schrittmotor 2 einschalten |
| 6E 68 | n h | Schrittmotor 2 auf Halbschrittmodus einstellen |
| 6E 6C | n l | Drehrichtung von Schrittmotor 2 links |
| 6E 72 | n r | Drehrichtung von Schrittmotor 2 rechts |
| 6E 73 | n s | Schrittmotor 2 führt Schritt aus |
| 6E 76 | n v | Schrittmotor 2 auf Vollschrittmodus einstellen |

Tabelle 17 - Befehlsübersicht für das Umgebungserfassungssystem